\documentclass[twoside,11pt]{article}

\usepackage{amsmath}
\usepackage{float}
\usepackage{booktabs}
\usepackage{arydshln}
\usepackage[]{algorithm}
\usepackage{algpseudocode}
\usepackage{xcolor}
\usepackage[normalem]{ulem} 
\usepackage{subfig}
\usepackage{mathtools}
\mathtoolsset{showonlyrefs}
\usepackage[preprint]{jmlr2e}

\newcommand{\N}{\mathcal{N}}

\usepackage{lastpage}
\jmlrheading{26}{2025}{1-\pageref{LastPage}}{8/22; Revised
7/25}{10/25}{22-0973}{Teemu H\"ark\"onen, Sara Wade, Kody Law, and Lassi Roininen}

\ShortHeadings{Mixtures of Gaussian Process Experts with SMC$^2$}{H\"ark\"onen, Wade, Law, and Roininen}
\firstpageno{1}

\begin{document}

\title{Mixtures of Gaussian Process Experts with SMC$^2$}

\author{\name Teemu H\"ark\"onen \email teemu.h.harkonen@aalto.fi \\
       \addr 
       Department of Electrical Engineering and Automation \\
       Aalto University \\
       Espoo, FI-02150, Finland \\
       School of Engineering Sciences\\
       LUT University\\
       Lappeenranta, Yliopistonkatu 34, FI-53850, Finland
       \AND
       \name Sara Wade \email sara.wade@ed.ac.uk \\
       \addr School of Mathematics\\
       University of Edinburgh\\
       Edinburgh, EH9 3FD, United Kingdom
       \AND
       \name Kody Law \email kody.law@manchester.ac.uk \\
       \addr Department of Mathematics\\
       University of Manchester\\
       Manchester, M13 9PL, United Kingdom
       \AND
       \name Lassi Roininen \email lassi.roininen@lut.fi \\
       \addr School of Engineering Sciences\\
       LUT University\\
       Lappeenranta, Yliopistonkatu 34, FI-53850, Finland}

\editor{Anthony Lee}

\maketitle

\begin{abstract}
Gaussian processes are a key component of many flexible statistical and machine learning models. However, they
 exhibit cubic computational complexity and high memory constraints due to the need of inverting and storing a full covariance matrix. To circumvent this, mixtures of Gaussian process experts have been considered where data points are assigned to independent experts, reducing the complexity by allowing inference based on smaller, local covariance matrices. Moreover, mixtures of Gaussian process experts substantially enrich the model's flexibility, allowing for behaviors such as non-stationarity, heteroscedasticity, and discontinuities. In this work, we construct a novel inference approach based on nested sequential Monte Carlo samplers to simultaneously infer both the gating network and Gaussian process expert parameters. This greatly improves inference  compared to importance sampling, particularly in settings when a stationary Gaussian process is inappropriate,  while still being thoroughly parallelizable. %
\end{abstract}

\begin{keywords}
  sequential Monte Carlo, SMC$^2$, Gaussian processes, mixture of experts, classification 
\end{keywords}
\section{Introduction}
\label{sec:introduction}
Gaussian processes (GP) are a versatile tool for statistical modelling of data where the exact form of the mapping between input and output variables is unknown and have found a plethora of applications.
Recent applications of GPs include prediction of battery life times \citep{Richardson:2017}, geometry optimization in theoretical chemistry \citep{Denzel:2018}, spectral peak structure prediction \citep{Wakabayashi:2018}, tokamak plasma modelling \citep{Ho:2019}, medical diagnosis using proteomic data \citep{yu2022non}, pile foundation design \citep{Momeni:2020}, and computational biology \citep{Otto:2024}, just to name a few.

However, straight-forward application of GPs suffers from cubic computational complexity and memory constraints due to the need of inverting and storing a full covariance matrix of size $N \times N$ for $N$ data points.
Moreover, many applications present departures from the standard GP model, such as non-stationary, heteroscedasticity, and non-normality. 
Methods such as non-stationary covariance functions or deep Gaussian processes are a possible solutions for modelling such behaviour \citep{Paciorek:2003, Porcu:2007, Damianou:2013, Heinonen:2016}.
However, they do not provide classification or clustering for individual data points or the input space.
To overcome the above limitations while providing classification, mixtures of GP experts have been considered \citep[e.g.][]{tresp2001mixtures,Rasmussen:2002,meeds2005alternative}.

Mixtures of GP experts belong to the general class of mixtures of experts (MoEs) \citep[see][for reviews]{masoudnia2014,gormley2019}. 
The nomenclature has it origins in machine learning \citep{Jordan:1994}, but they appear in many different contexts such as in statistics, where they are often called dependent mixtures \citep{MacEachern:1999, quintana2022dependent, 10.1214/24-STS966} and econometrics where they are referred to as smooth mixtures \citep{Li:2011, Norets:2010, villani2012generalized}.
MoEs probabilistically partition the input space into regions and specify a local expert within each region.
The gating network maps the experts to local regions of the input space, determining the partition of the data into independent sets. Local experts are fit independently to the data sets, where each expert is a conditional model for the output $y$ given the input $x$, which is specifically a GP model in the case of mixtures of GP experts.
This allows for both improved scalability, as each GP expert fit only requires inversion and storing of the local covariance matrices, and flexibility, as GP experts can capture different behaviors, e.g. smoothness and variability, within each region.

Inference for mixtures of GP experts is often performed via Markov chain Monte Carlo (MCMC), namely, Gibbs sampling \citep{meeds2005alternative, Rasmussen:2002, Gadd:2020}, by augmenting the model with a set of latent allocation variables denoting the labels of the expert associated to each data point.
This provides asymptotically exact samples from the posterior, however it often comes with a high computational cost due to slow mixing and long convergence times.
For faster approximate inference, \cite{yuan2008variational} develop a variational inference scheme, and \cite{tresp2001mixtures} employs an expectation-maximization algorithm for maximum a posteriori (MAP) inference.
A combination of variational inference, MAP, and sparse GPs \citep{quinonero2005unifying} is developed in \cite{nguyen2014fast} to improve scalability to big data (e.g. training size of $10^5$), and this is further expanded upon in \cite{etienam2020ultra} for fast one-pass approximate MAP inference.  

Recently, \citet{Zhang:2019} proposed an embarrassingly parallel approach for mixtures of GP experts based on importance sampling.
Importantly, this allows for uncertainty quantification in the parameters and quantities, as opposed to MAP inference, and avoids the independence or distributional assumptions of variational inference.
Moreover, in the proposal distribution, random partitions of the data points are generated, allowing for easily distributed GP likelihood computation. 
However, it is well known that naive importance sampling requires the number of importance samples to be exponential in the the Kullback-Liebler (KL) divergence of the posterior from the proposal distribution \citep{chatterjee2018sample}.
We discuss how this number can be quite large for mixtures of GP experts due to the massive number of possible partitions, leading to poor inference, particularly when a stationary GP model is inappropriate. 

To address this, we propose to improve upon the approach of \citet{Zhang:2019} by using nested sequential Monte Carlo (SMC) samplers, as introduced by \citet{chopinsmc2}, as a way of inferring the model parameters from the posterior.
We discuss and demonstrate how this provides a more accurate and robust approach, especially for complex target distributions resulting from complex behaviors of the data, such as non-stationarity, discontinuities, or heterogeneity.
Furthermore, for complex target distributions our method can be much more computationally efficient, while still being thoroughly parallelizable.
It is well-known that SMC methods reduce importance sampling complexity from exponential to polynomial \citep{Beskos:2014}.
Some parallelizability must be sacrificed due to weight normalization, but all other stages of the algorithm can be executed independently and in parallel.
There exist versions of SMC that can combine independent asynchronous SMCs in lieu of a single monolithic SMC as well, provided that a minimum ensemble size is used for each one \citep{Jasra:2020, liang2025scalable}.
If more parallel capacity is available, then these methods could be leveraged in place of any of the inner parallel SMCs and/or the outer SMC in our method.
Additionally, it is worth noting that several studies have previously proposed using SMC in various different ways to sample from standard mixture models, such as \cite{Bouchard:2017} for split-merge moves, amongst other approaches \citep{MacEachern:1999, Fearnhead:2004, Fearnhead:2007, Mansinghka:2007, Caron:2009, Carvalho:2010, Ulker:2010}.
SMC methods have also been proposed for GP inference, see for example \cite{ Gramacy:2011, Svensson:2015}.

In summary, we formulate mixtures of GP experts such that nested SMC samplers can be used.
The nested SMC samplers, or SMC$^2$, allow for separation of the model inference into two separate parts, the sampling of the gating network parameters and GP parameters.
We highlight that this allows for incorporation of uncertainty in the GP parameters, in contrast to the MAP estimates employed in \cite{Zhang:2019}.
The use of SMC$^2$ greatly extends the set of feasible target posterior distributions from the  previously proposed embarrassingly parallel importance sampling of mixtures of GP experts.
This is achieved by the improved computational efficiency, allowing improved inference of the mixture of GP experts parameters.
Alternatively, one can obtain inference of similar quality with fewer samples due to the improved KL divergence provided by the SMC$^2$.

The rest of the paper is organized as follows. We formulate the statistical model in Section \ref{sec:model}, which is followed by a review of importance sampling in Section \ref{sec:importanceSampling}.
We present the proposed method by first introducing the inner SMC sampler in Section \ref{sec:SMC} and concluding with the nested SMC$^2$ sampler in Section \ref{sec:SMC2}. 
In Section \ref{sec:predictiveDistribution}, we provide our construction of the predictive distribution, and results are presented for two illustrative one-dimensional data sets along with one-dimensional motorcycle helmet acceleration, 3D NASA Langley glide-back booster simulation, and 4D Colorado precipitation data sets in Section \ref{sec:numericalExamples}.
We conclude with final remarks and a discussion in Section \ref{sec:discussion}.
\section{Model}
\label{sec:model}
In a supervised setting, MoEs provide a flexible framework for modelling the conditional relationship between the outputs $y$ given the inputs $x$, and two general approaches exist for defining MoEs, namely, generative and discriminative.
Generative MoEs model $(y,x)$ jointly, while discriminative MoEs model the conditional of interest directly. In this work, we focus on the discriminative approach, which has the advantage of avoiding modelling of and placing any distributional assumptions on the input variables. 

We assume that the data consists of a sample of size $N$ input variables $X = \{x_1, \dots, x_N \}$, with $x_i = ( x_{i,1}, \dots, x_{i,D})$, and output variables $Y = \{y_1, \dots, y_N \}$. 
For simplicity, we focus on the case when $x_i  \in \mathbb{R}^{ D}$ and  $y_i \in \mathbb{R}$, although extensions for other types of inputs and outputs (e.g. binary, categorical, counts) can also be constructed.
MoEs assume 
\begin{equation}
    y_i \mid x_i, \Psi, \Theta \sim \sum_{ k = 1}^K p_k(x_i \mid \Psi) p(y_i \mid x_i, \Theta_k),
\end{equation}
where $( p_1(x_i \mid \Psi), \dots, p_K(x_i \mid \Psi)) =: {\boldsymbol{p}}(x_i \mid \Psi) $ is the gating network which takes values in the $K-1$-dimensional simplex and the expert $p(y \mid x_i, \Theta_k)$ is a conditional model, with local relevance determined by $p_k( x_i \mid \Psi)$ at the input location $x_i$.
Here, $\Psi := ( \psi_1, \ldots, \psi_K)$ is a shorthand for the gating network parameters and $\Theta = ( \Theta_1, \ldots, \Theta_K) $ contains the expert parameters.
MoEs can be augmented with a set of latent allocation variables $c_i \in \lbrace 1, \ldots, K \rbrace$ and equivalently represented by
\begin{equation}
\begin{split}
     y_i \mid x_i, c_i= k, \Theta_k &\sim  p(y_i \mid x_i,\Theta_k),\\
     c_i \mid x_i, \Psi & \sim \text{Categorical}\left( {\boldsymbol{p}(x_i \mid \Psi)} \right),
 \end{split}
\end{equation}
where $c_i$ determines the partition of the data points into the $K$ groups.
We use $C = (c_1, \dots, c_N)$ to denote the vector of labels associated with each data point.
Various proposals exist for different choices of the experts and gating networks, ranging from linear to nonlinear approaches, such as trees \citep{Gramacy:2008}, neural networks \citep{etienam2020ultra}, and GPs \citep{tresp2001mixtures}.
In the following, we consider GP experts and define the weights through normalized kernels. 
 
\subsection{Experts}
For the experts, we consider the nonparametric regression model:
\begin{equation}
    y_i \mid x_i, c_i=k, f_k, \sigma^2_{k,\epsilon} \sim \N(f_k(x_i), \sigma^2_{k,\epsilon}),
\end{equation}
where $f_k(\cdot)$ is the unknown regression function mapping the inputs to the outputs and $\sigma^2_{k,\epsilon}$ is the noise variance for the $k$th expert.
We place a nonparametric GP prior \citep{williams2006gaussian} on the regression function:
\begin{equation}
    f_k \sim \text{GP}\left( m_k, \Sigma( \cdot, \cdot; \theta_k) \right),
\end{equation}
with $m_k$ and $\Sigma( \cdot,\cdot; \theta_k)$ denoting, respectively, a constant mean and the covariance function depending on the kernel hyperparameters $\theta_k$.
In the regression setting, the functions can be analytically marginalized, resulting in the likelihood:
\begin{equation}
    Y_k \mid \left( X_k, m_k, \theta_k \right) \sim \N\left( m_k, \Sigma( X_k, X_k; \theta_k) + \sigma^2_{k,\epsilon}I \right),
    \label{eq:gpLikelihood}
\end{equation}
where $(X_k, Y_k) = \{ (x_i, y_i) : c_i = k \}$ are the data points with category $c_i = k$ with $\Sigma( X_k, X_k; \theta_k)$ denoting the covariance function evaluated at the inputs $\{ x_i : c_i = k \}$ and $I$ denotes the identity matrix. 
Extensions for multivariate or other types of outputs can be constructed through a generalized multivariate Gaussian process framework \citep{Gadd:2020}.

\begin{figure}
    \centering
    \includegraphics[width=0.9\textwidth]{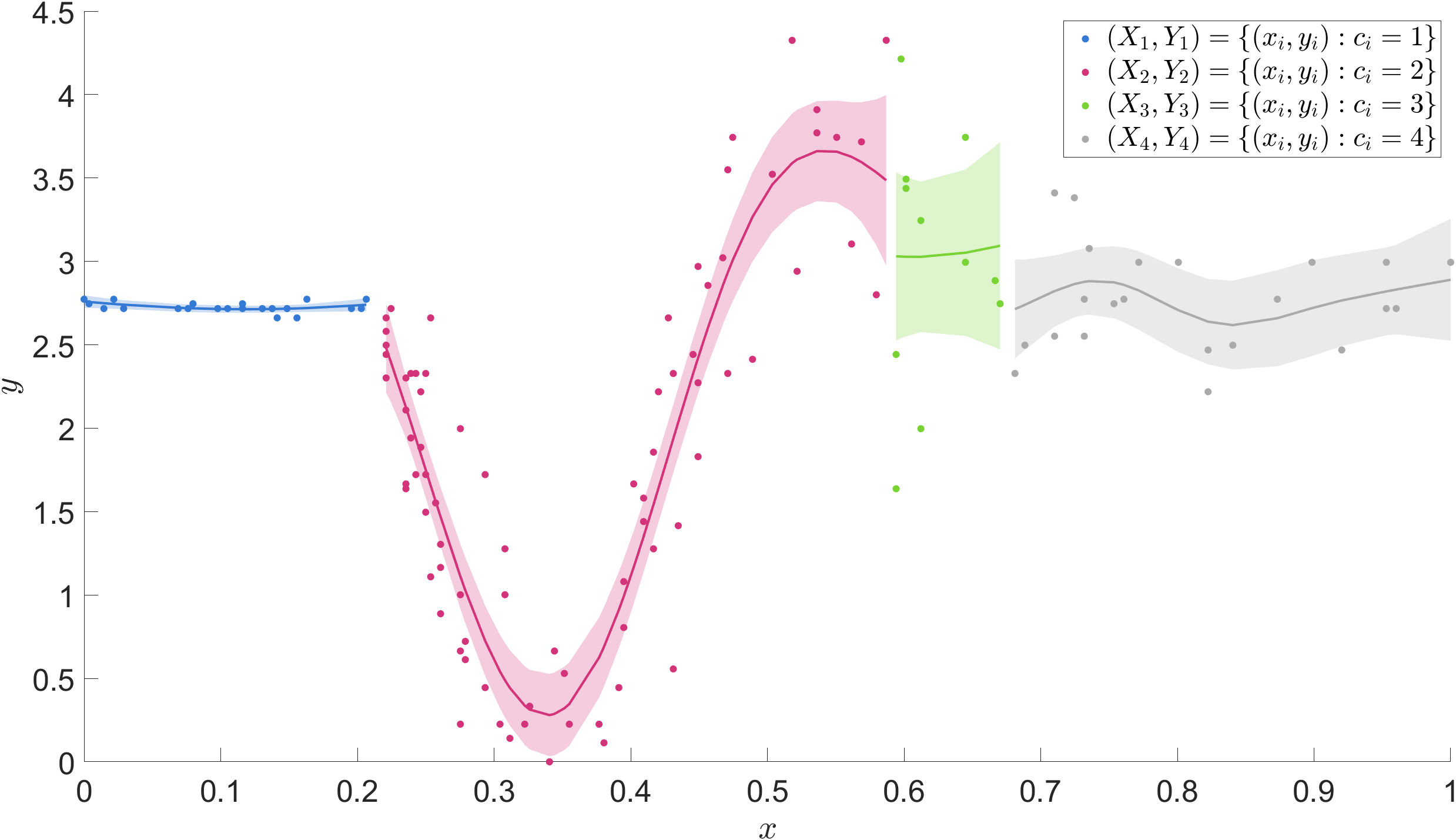}
    \caption{An example of partitioned data with four clusters. The clustered data points are shown in blue, red, green, and gray for each cluster with their respective GP mean and 95\% pointwise credible intervals in the corresponding color.}
    \label{im:examplePartition}
\end{figure}

To simplify notation, we incorporate the noise variance within the covariance function and use an anisotropic squared exponential covariance function for each of the GPs, such that the $ij$th element of the covariance matrix is given by
\begin{equation}
    \left[ \Sigma( X_k, X_k; \theta_k) \right]_{i,j} = \sigma_{k,\varepsilon}^2\delta( x_{i} - x_{j}) + \sigma_{k,f}^2 \prod_{d = 1}^D \exp\left( -\frac{ (x_{i,d} - x_{j,d})^2 }{ l_{k,d}^2 } \right),
\end{equation}
where $x_{i}, x_{j} \in X_k$, $\sigma_{k,f}^2$ is the signal variance, and $l_d$ is the length scale along the $d$th dimension, with $\delta(x)$ denoting the Kronecker delta function and  $\theta_k = (\sigma_{k,\varepsilon}, \sigma_{k,f}, l_{k,1}, \ldots, l_{k,D})$.
Furthermore, we use the notation $\Theta_k = (m_k, \theta_k)$ for the k$th$ GP parameters and $\Theta = (\Theta_1, \ldots, \Theta_K)$ for all the GP parameters.
An example partition and its corresponding GP fits are illustrated in Figure \ref{im:examplePartition}.
\subsection{Gating networks}
We define the gating network through the kernels $\left\{\mathcal{K}_1( x_i \mid \psi_1), \dots, \mathcal{K}_K( x_i \mid \psi_K) \right\}$; specifically,
\begin{equation}
    p_k(x_i) = \frac{ \mathcal{K}_k(x_i \mid \psi_k)}{ \sum_{ k' = 1}^K \mathcal{K}_{k'}(x_i \mid \psi_{k'})}. 
    \label{eq:gatingnetwork}
\end{equation}
To ensure the gating network $\boldsymbol{p}(x_i) = ( p_1(x_i), \dots, p_K(x_i))^T$ takes values in the $K-1$-dimensional simplex, we assume $0<\mathcal{K}_k( x \mid \psi)<\infty$ for all $x \in \mathbb{R}^D$.
Examples of kernels used in literature include the multinomial logistic model with $\mathcal{K}_k( x \mid \psi_k) = \exp(\psi_k^Tx)$ \citep{Li:2011} and the factorized kernel $\mathcal{K}_k( x \mid \psi_k) = v_k \prod_{d=1}^D \mathcal{K}_{k,d}( x_d \mid \psi_{k,d})$ \citep{antoniano2014bayesian}.
In the latter, mixed types of inputs are allowed by specifying each $\mathcal{K}_{k,d}( x_d \mid \psi_{k,d})$ to correspond to the kernel of standard distributions, depending on the nature of the $d$th input.
We emphasize that this does not correspond to any model assumptions on the input, which in fact may be fixed.

In the following, we use weighted $D$-dimensional normal distributions with diagonal covariance matrices for the kernels, individually denoted by $\mathcal{K}_k$:
\begin{equation}
    \mathcal{K}_k( x_i \mid \psi_k) := \mathcal{K}_k( x_i \mid v_k, \mu_k, \Sigma_k) = v_k \mathcal{N}( x_i \mid \mu_k, \Sigma_k), \label{eq:kernel}
\end{equation}
where $v_k$ denotes the weight of the component with $\mu_k \in \mathbb{R}^{ D}$ and $\Sigma_k \in \mathbb{R}^{D \times D}$ being the mean and a diagonal covariance matrix, respectively.  
As such, we define $\psi_k := ( v_k, \mu_k, \Sigma_k)^T = ( v_k, \mu_{k,1}, \ldots, \mu_{k,D}, \sigma^2_{k,1}, \ldots, \sigma^2_{k,D})^T$ where $(\sigma^2_{k,1}, \dots, \sigma^2_{k,D} )$ are the diagonal elements of the $k$th covariance matrix.
Hard cluster assignments using the above kernel result in quadratic boundaries.
More flexible kernel specifications could be used for nonlinear boundaries such as mixtures of normals or the feature space method, where partitioning is done in a higher-dimensional space, producing non-linear partitions in the original input space \citep{Filippone:2008}.
More generally, alternative specifications of the kernel may be employed.

For example, \cite{Zhang:2019} also construct the gating network through normalized kernels and consider full covariance matrices $\Sigma_k$, employing the normal-inverse Wishart distribution as a prior for the means and covariances of the gating network.
Instead the factorized form of the kernel based on a diagonal covariance matrix in Eq.~\eqref{eq:kernel} results in reduced computational complexity relative to the dimension $D$ of the inputs, from $\mathcal{O}(D^3)$ in the case of the full covariance matrix to $\mathcal{O}(D)$ in the factorized form.
Furthermore, the factorized form allows for straightforward extensions to include multiple input types, as in \cite{antoniano2014bayesian}.
Lastly, the inverse-Wishart prior can suffer from poor parametrization \citep{consonni2001conditionally}, with only a single parameter to control variability.
We instead follow \cite{Gelman:2006} in using independent half-normal distributions for the variance parameters.
Additionally, we use a normal distribution for the mean $\mu_k$.

For the weights, we employ a Dirichlet prior:
\begin{equation}
    (v_1, \dots, v_K) \sim \text{Dir}( \alpha / K, \dots, \alpha / K),
    \label{eq:dirichletWeights}
\end{equation}
where $\alpha$ is the Dirichlet concentration parameter.
The concentration parameter $\alpha$ can be used to enforce sparsity \citep{Rousseau:2011}, as in sparse hierarchical MoEs.
This helps to avoid overfitting, which has been noted to be prevalent in dense mixtures \citep{Titsias:2002, Mossavat:2011, Iikubo:2018, Ozan:2021}.
Smaller concentration parameter values promote sparsity, and in the extreme case when $\alpha \rightarrow 0$, prior mass is concentrated on the vertices of the simplex and all weight is placed on a single component.
Thus, $K$ can be considered an upper bound on the number of components, and through appropriate selection of $\alpha$, the sparsity-promoting prior allows the data to determine the number of components required. 
In standard mixtures, \cite{Rousseau:2011} show that asymptotically the extra components will be emptied, provided that $\alpha/K < \varrho/2$, where $\varrho$ is the number of parameters in a single mixture component.
Alternatively, the Jeffreys prior for the multinomial distribution, with $\alpha = K/2$ in our case, can be used as a simple non-informative choice \citep{Grazian:2018}.
The Dirichlet-distributed weights can also be constructed through independent Gamma random variables, that is $\nu_1, \ldots, \nu_K$ are independent with $\nu_k \sim \text{Gamma}(\alpha/K, 1)$ and defining each $v_k = \nu_k / \sum_{k'=1}^K \nu_{k'}$.
Notice that the normalization term $\sum_{k=1}^K \nu_{k}$ cancels out in our definition of the gating network, so we can rewrite Eq.~\eqref{eq:kernel} as: 
\begin{equation}
    \mathcal{K}_k( x_i \mid \psi_k) := \mathcal{K}_k( x_i \mid v_k, \mu_k, \Sigma_k) = \nu_k \mathcal{N}( x_i \mid \mu_k, \Sigma_k).
\end{equation}

\subsection{Mixture of GP experts}\label{sec:mod}
\begin{figure}
\begin{center}
    \includegraphics[width=0.3\textwidth]{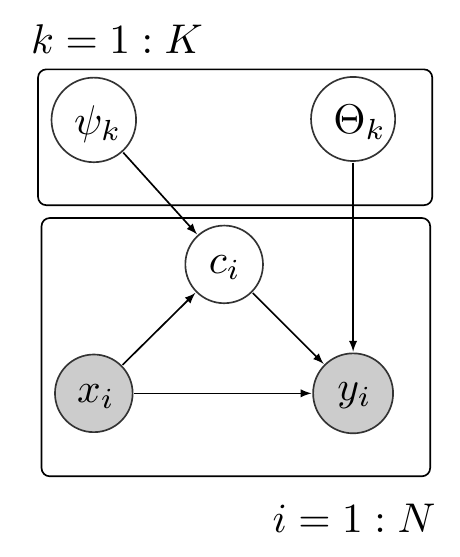}
	\caption{Plate diagram showing relations between the data and model parameters for MoE.}
	\label{im:gp_moe}
\end{center}
\end{figure}
\begin{table}
\centering
\begin{tabular}{c|c}
\toprule
Parameter & Prior distribution \\
\midrule
$\mu_{k,d}$ & $\mathcal{N} \left( G_{k,d}, \left( \frac{\Delta_G}{ \sqrt[\leftroot{0} \uproot{0} D] {K} + 1} \right)^2 \right)$ \\
$\sigma_{k,d}$ & $\mathcal{N}_{\frac{1}{2}}\left(0, \left( \frac{\Delta_\sigma}{ \sqrt[\leftroot{0} \uproot{0} D] {K} + 1}\right)^2 \right)$ \\
$m_k$ & $\mathcal{U} \left( 0, \max Y \right)$ \\
$\sigma_{k,\varepsilon}$ & $\mathcal{N}_{\frac{1}{2}} \left(0, 0.25^2 \right)$ \\
$\sigma_{k,f}$ & $\mathcal{N}_{\frac{1}{2}} \left(0, 0.25^2 \right)$ \\
$l_{k,1}$ & $\mathcal{N}_{\frac{1}{2}} \left(0, 0.125^2 \right)$ \\
\bottomrule
\end{tabular}
\caption{Prior distributions for $\Psi$ and $\Theta$ for our higher-dimensional data sets. To encourage components that are well-separated along the input space, the prior means for the gating networks parameters $\mu_k$ are constructed using a linearly-spaced grid, so that spacing along each dimension is $1/\sqrt[\leftroot{0} \uproot{0} D] {K}$, with $G_{k}= (G_{k,1},\ldots, G_{k,D})$ denoting the $k$th grid point. For $D = 1$, we use $\Delta_G = 0.25$ and $\Delta_\sigma = 0.25$ with $\Delta_G = 0.05$ and $\Delta_\sigma = 0.01$ for the higher-dimensional examples.}
\label{table:priorParametersND}
\end{table}
The above results in our statistical model being 
\begin{equation}
\begin{split}
 Y_k \mid \left( X_k, m_k, \theta_k \right) &\sim \text{GP}\left( m_k, \Sigma( X_k, X_k; \theta_k) \right),\\
    c_i \mid x_i, \Psi & \sim \text{Categorical}({\boldsymbol{p}(x_i)}), \\
    \psi_k &\sim \pi_{0}(\psi_k), \\
    \Theta_k &\sim \pi_0(\Theta_k),
\end{split}
\end{equation}
where $\pi_0(\Theta_k)$ denotes the prior distribution for the GP parameters, which assumes independence across $k=1,\ldots, K$, and
\begin{equation}
    p_k(x_i) = \frac{ \nu_k \mathcal{N}( x_i \mid \mu_k, \Sigma_k)}{ \sum_{ k' = 1}^K \nu_{k'} \mathcal{N}( x_i \mid \mu_{k'}, \Sigma_{k'})},
\end{equation}
with $\nu_k \sim \text{Gamma}(\alpha/K, 1)$ and $\mu_{k,d}$ and $\sigma^2_{k,d}$ assigned independent normal and half-normal priors, respectively.
A plate diagram of the model is provided in Figure \ref{im:gp_moe}.
The prior distributions used for one-dimensional and higher-dimensional data sets are presented in Table \ref{table:priorParametersND}, respectively.
\section{Importance sampling}
\label{sec:importanceSampling}
\cite{Zhang:2019} propose an embarrassingly parallel algorithm based on importance sampling (IS) to estimate the posterior distribution of $C$ and $\Psi$, using a MAP approximation for the GP parameters $\Theta$ as well as the predictive distribution of the data in parallel. 
For this, an importance distribution $q(C, \Psi \mid X)$ is constructed which is easy to sample from.
The importance distribution is chosen to be the prior distribution such that $q(C, \Psi \mid X) = \pi(C, \Psi \mid X)$. Thus,
\begin{equation}
\begin{split}
    q(C, \Psi \mid X) &= p( C \mid X, \Psi) \pi(\Psi) = \prod_{i=1}^N p(c_i \mid x_i, \Psi) \prod_{k=1}^K \pi_0(\psi_k) \\
    &= \prod_{k = 1}^K \prod_{ i : c_i = k} \frac{ v_k N( x_i \mid \mu_k, \Sigma_k)}{ \sum_{ k' = 1}^K v_{k'} N(x_i \mid \mu_{k'}, \Sigma_{k'} )}  \pi_0(\psi_k).
\end{split}
\label{eq:proposal}
\end{equation}
Using the above, $J$ particles are sampled in parallel from $q(C, \Psi \mid X)$ and the weight $\omega_j$ for the $j$th particle $(C_j, \Psi_j)$ in the IS estimation is given as  
\begin{equation}
\begin{split}
    \omega_j &\propto \frac{ \pi( C_j, \Psi_j \mid X, Y)}{ q( C_j, \Psi_j \mid X)} \propto \frac{ p( Y \mid X, C_j, \Psi_j) \pi( C_j, \Psi_j \mid X)}{\pi( C_j, \Psi_j \mid X)} \\ &= p( Y \mid X, C_j, \Psi_j) \approx  \prod_{k=1}^K \N\left(Y_{k,j} \mid \widehat{m}_{k,j}, \Sigma( X_k, X_k; \widehat{\theta}_{k,j})\right),
\end{split}
\label{eq:weightsImportanceSampling}
\end{equation}
where $\widehat{m}_{k,j}$ and $\widehat{\theta}_{k,j}$ are MAP estimates. 
For a new input $x^*$, the predictive mean and density of the output are computed by averaging according to the respective particle weights and gating network probabilities at the new location $x^*$:
\begin{equation}
\begin{split}
    \mathbb{E}[ y^* \mid x^*, X, Y] &= \sum\limits_{j = 1}^{J}  \omega_j \sum\limits_{k=1}^K  m^*_{k,j} \, p_k( x^* \mid \Psi_{j}),\\
        \pi( y^* \mid x^*, X, Y) &= \sum\limits_{j = 1}^{J}  \omega_j \sum\limits_{k=1}^K \mathcal{N}( y^* \mid m^*_{k,j}, \Sigma^*_{k,j}) \, p_k( x^* \mid \Psi_{j}),
\end{split}
\label{eq:importanceSamplingPredictions}
\end{equation}
where $ m^*_{k,j}$ and $\Sigma^*_{k,j}$ are the predictive mean and covariance of the associated $K$ GPs for each of the $J$ particles.
For more details on the importance sampling approach, see \cite{Zhang:2019}.

As mentioned in the Introduction, the number of particles required for importance sampling is exponential in the KL divergence of the posterior from the proposal distribution \citep{chatterjee2018sample}, that is $J \propto \exp[{D_{\text{KL}}(\pi( C, \Psi \mid X, Y) \, \| \, q( C, \Psi \mid X))}]$. 
As an extreme example, consider a uniform proposal distribution for $C$, with $q(C \mid X, \Psi) = 1/K^N$, obtained in the limiting case when $\alpha \rightarrow \infty$ and $\sigma^2_{k,d} \rightarrow \infty$.
On the other hand, suppose the posterior on $C$ is very concentrated on a single partition $C^*$; due to invariance with respect to a relabelling of the components, the posterior in this case is uniform over the set $\mathcal{C}$ of $C$ that are equivalent to $C^*$ up to relabelling of the components.
Thus, $\pi(C \mid X,Y) = \sum_{C \in \mathcal{C}} 1/|\mathcal{C}| \delta_{C}$, where the size of $\mathcal{C}$ is $|\mathcal{C}| = \binom{K}{t}t!$, with $t$ denoting the number of clusters in $C^*$.
Here, the KL divergence is:
\begin{equation}
    {D_{\text{KL}}(\pi( C, \Psi \mid X, Y) \, \| \, q( C, \Psi \mid X))} = \sum_{C \in \mathcal{C}} 1/|\mathcal{C}| \log \left( \frac{K^N}{|\mathcal{C}|}\right) = \log \left( \frac{K^N}{\binom{K}{t} t!} \right),
\end{equation}
and the number of particles required is $J \propto  K^N/(\binom{K}{t} t!)$. Even for a very small sample size, e.g. $N=10$, $K=5$, and $t=2$, this requires $J \propto 488,281$, and when $N=20$, this increases to $J \propto 4,768,371,582,031$. 

While this is a purposefully constructed edge case, some insight on behaviour for stationary data in comparison to data that present departures, such non-stationary,  discontinuity, or heteroskedasticity, can be postulated.
Stationary data can be modelled well over a large class of partitions, as all data can be effectively modelled via a single GP model, and partitioning into any reasonable number of subgroups will not degrade the fit too much.
Therefore, sampling of the partitions can be done effectively using strategies such as importance sampling, as the ``distance'' between the proposal and posterior is smaller in comparison to non-stationary and discontinuous cases.
Instead, the partitions play a more important part in modelling non-stationary and discontinuous data; the posteriors are concentrated on a smaller number of partitions, thus requiring more samples to have a suitable approximation.

This motivates us to use sequential Monte Carlo (SMC) methods, and specifically SMC$^2$, to improve our estimation of the posterior or retain the same level of quality with the same number of or fewer particles.
SMC methods, also known as particle filtering and smoothing, are a standard tool in statistical signal processing \citep{Sarkka:2013} and provide a robust method for sampling posterior distributions \citep{Chopin:2002, Moral:2006}.
In particular, we use sequential likelihood tempering to estimate the gating network parameters $\Psi$ and the allocation variables $C$, while fitting $K$ independent Gaussian processes to the partitioned data.
This is in contrast to parameter estimation for a single GP using data tempering, such as by \cite{Gramacy:2011} and \cite{Svensson:2015}.
In our case, likelihood tempering seems more appropriate, as the mixture configuration will depend heavily on new data, and dynamically adding data may require significant readjustments of the gating network.
For further discussion on using tempering versus growing subsets of data, see Sections 17.3 and 17.3.4, in particular, of \cite{chopinpapa}
In addition, as SMC methods employ a collection of weighted particles, their application is a natural extension to the methodology presented in \cite{Zhang:2019}.

Owing to the nested structure of the SMC$^2$, we refer to the nested SMC samplers as the inner and outer SMC samplers.
Below, we present the posterior distribution for the discriminative MoE model and the inner SMC sampler for estimating the GP expert parameters for a fixed partition and gating network parameters $C$ and $\Psi$.
This is followed by the presentation of the outer SMC sampler for the gating network parameters and the partition.
Together, these two samplers constitute our full nested SMC$^2$ estimation scheme.
\section{Inner SMC sampler}
\label{sec:SMC}
Using the discriminative MoE model in Section \ref{sec:mod}, the target posterior distribution for $C$, $\Psi$, and $\Theta$ is
\begin{equation}
    \pi( C, \Psi, \Theta \mid X, Y) \propto p(Y \mid X, C, \Theta) p(C \mid X, \Psi) \pi_0(\Psi)\pi_0( \Theta),
    \label{eq:gatingPartitionCovParamPosterior}
\end{equation}
which can be marginalized with respect to the GP parameters $\Theta$ to yield
\begin{equation}
    \pi( C, \Psi \mid X, Y) \propto p(Y \mid X, C) p(C \mid X, \Psi) \pi_0( \Psi ).
    \label{eq:gatingPartitionPosterior}
\end{equation}
In this case, the likelihood $p(Y \mid X, C)$ is given as
\begin{equation}
\begin{split}
    p(Y \mid X, C) &= \int p(Y \mid X, C, \Theta) \pi_0(\Theta) \, \text{d}\Theta = \prod\limits_{k = 1}^K \int p\left( Y_k \mid X_k, \Theta_k \right)\pi_0( \Theta_k) \, \text{d}\Theta_k,
\end{split}
\label{eq:partitionLikelihood}
\end{equation}
with the log-likelihood of the individual GPs being
\begin{equation}
\begin{split}
    \log p( Y_k \mid X_k, \Theta_k) = &-\frac{1}{2} ( Y_k - m_k)^T \Sigma( X_k, X_k; \theta_k)^{-1} ( Y_k - m_k)\\ &-\frac{1}{2} \log \left\vert \Sigma( X_k, X_k; \theta_k) \right\vert - \frac{N_k}{2} \log 2\pi,
\end{split}
\end{equation}
where $\left\vert \Sigma( X_k, X_k; \theta_k) \right\vert$ is the determinant of the covariance matrix and $N_k$ is the number of data points in the $k$th group.

\begin{figure}[!t]
    \centering
    \includegraphics[width = 0.32\textwidth]{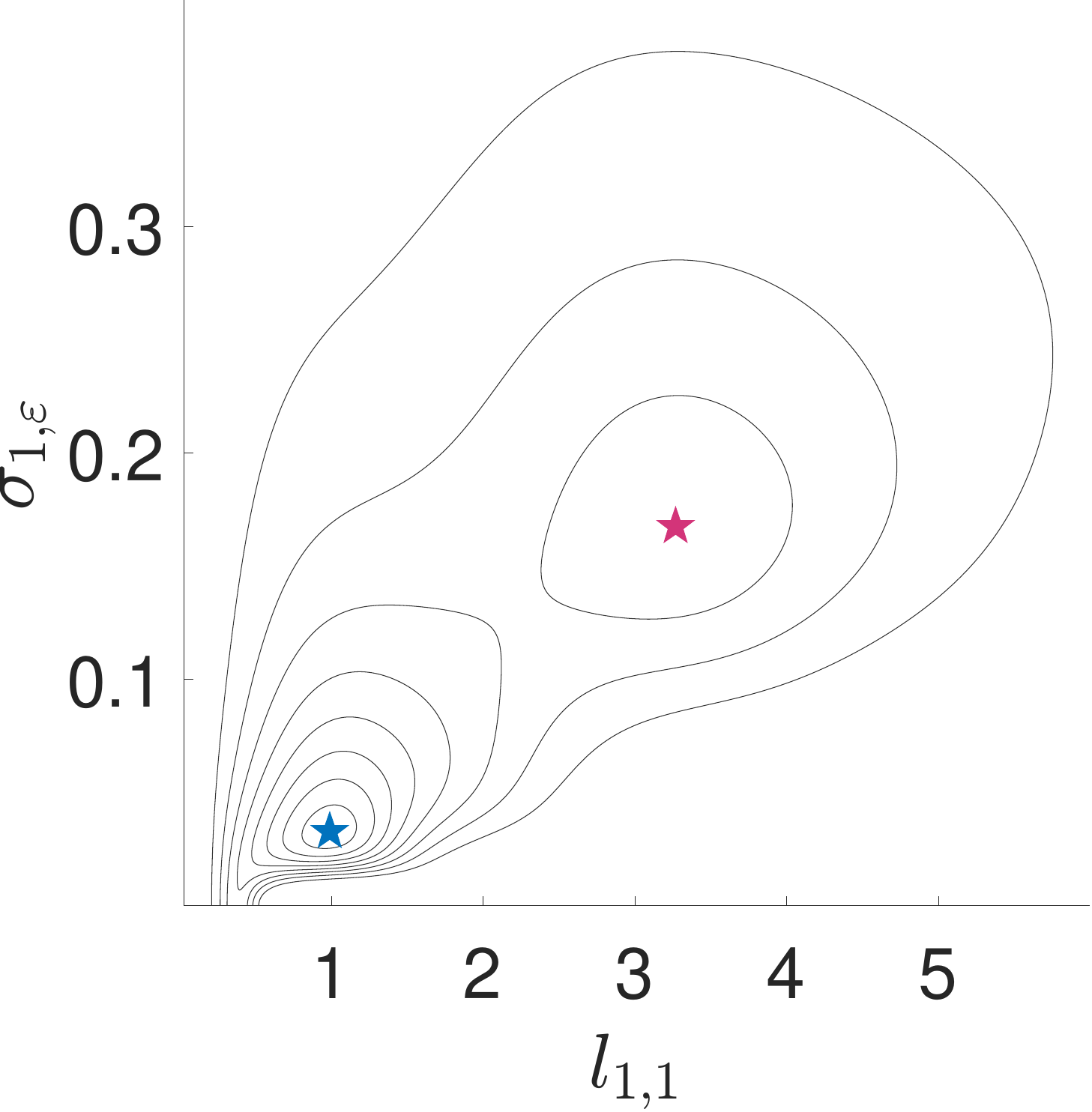}
    \includegraphics[width = 0.32\textwidth]{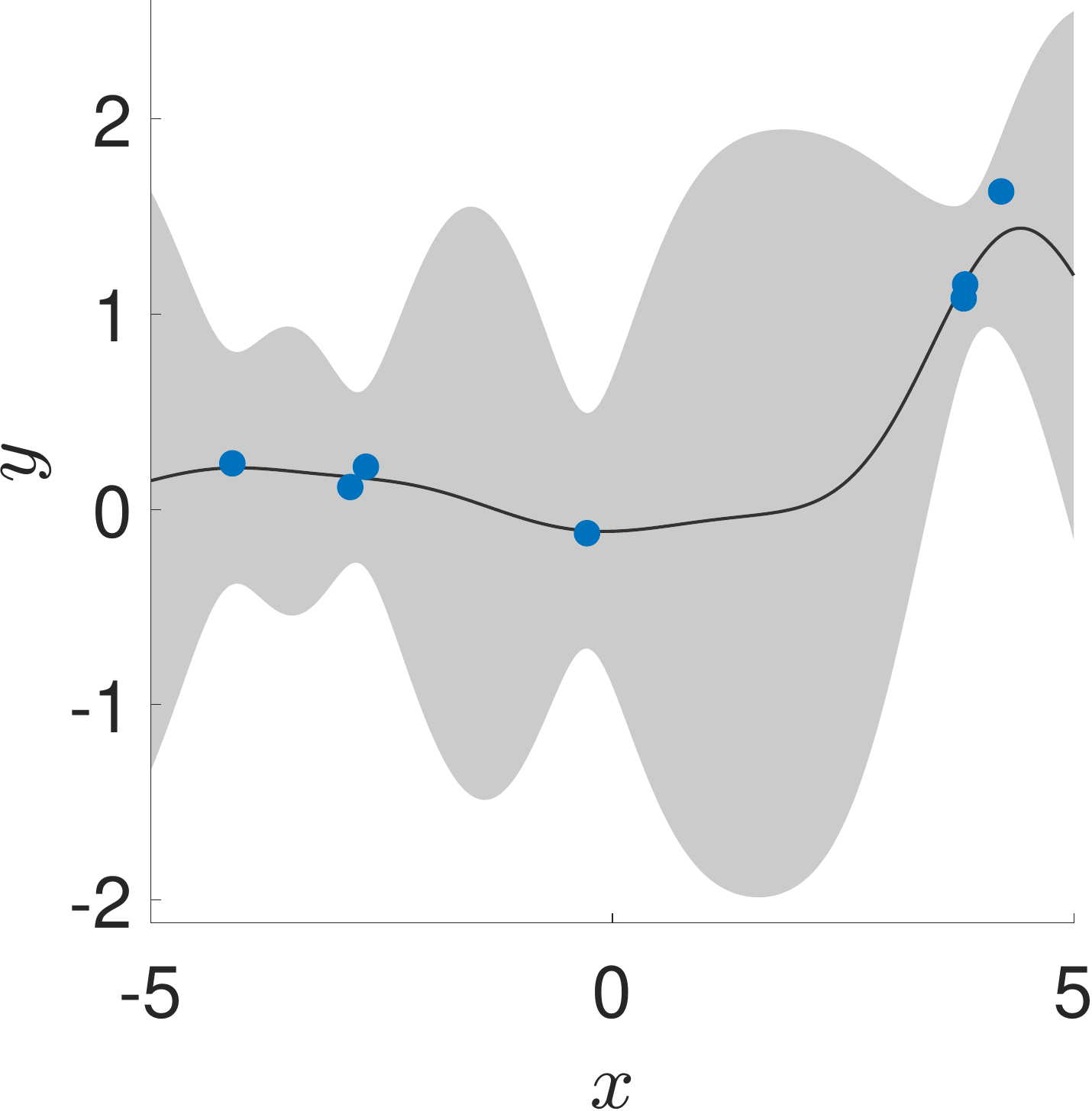}\\
    \includegraphics[width = 0.32\textwidth]{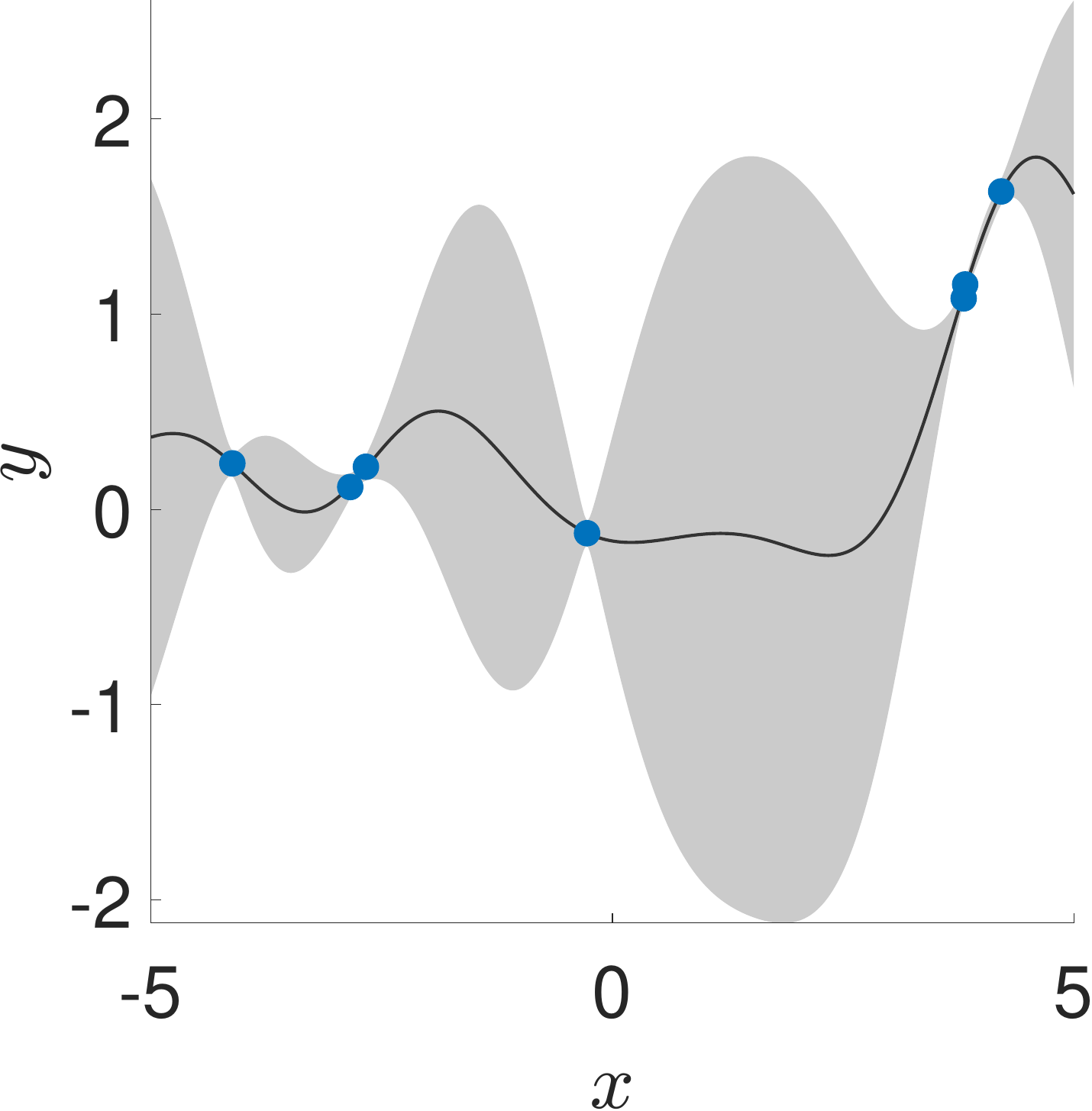}
    \includegraphics[width = 0.32\textwidth]{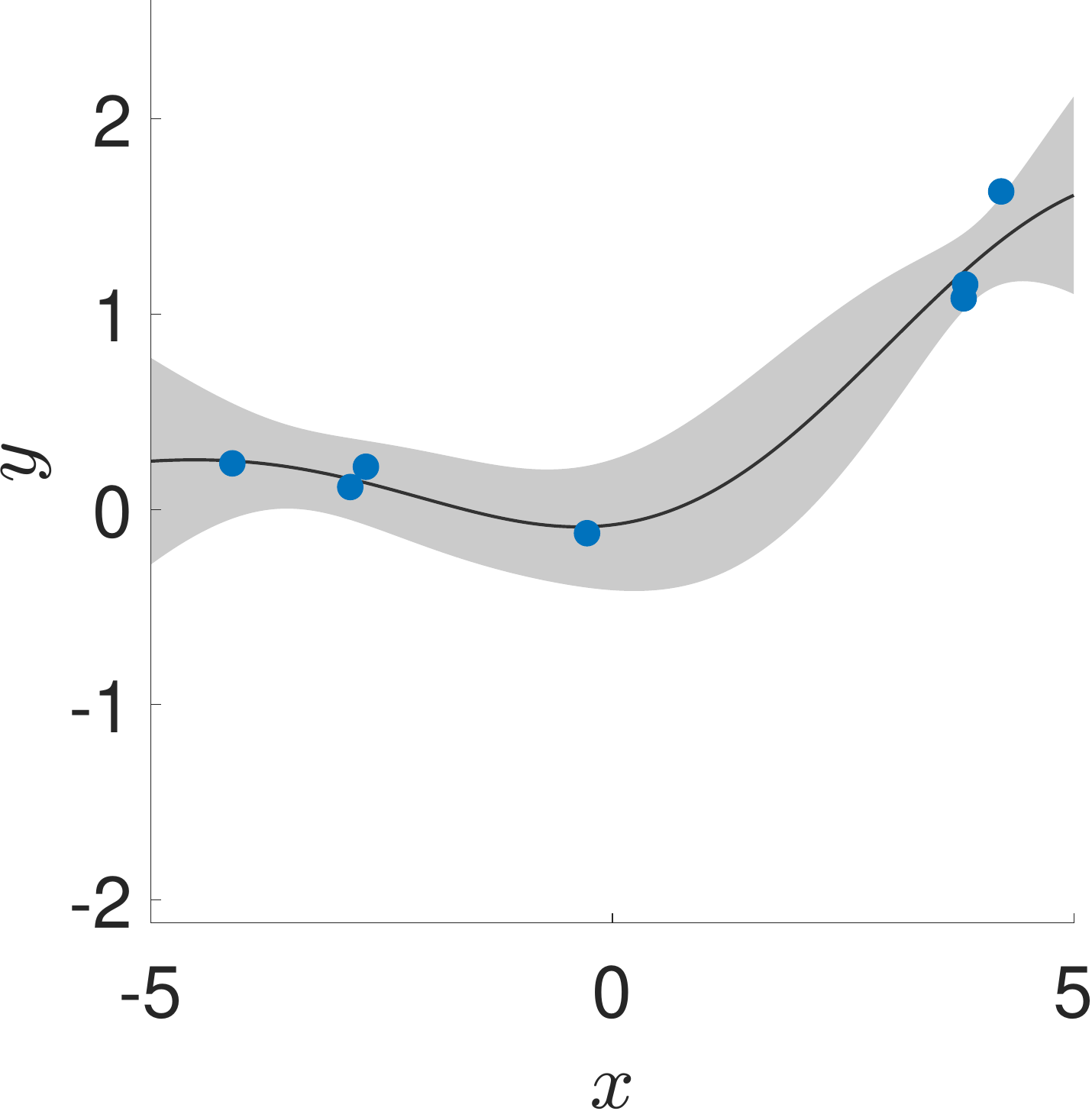}
    \includegraphics[width = 0.32\textwidth]{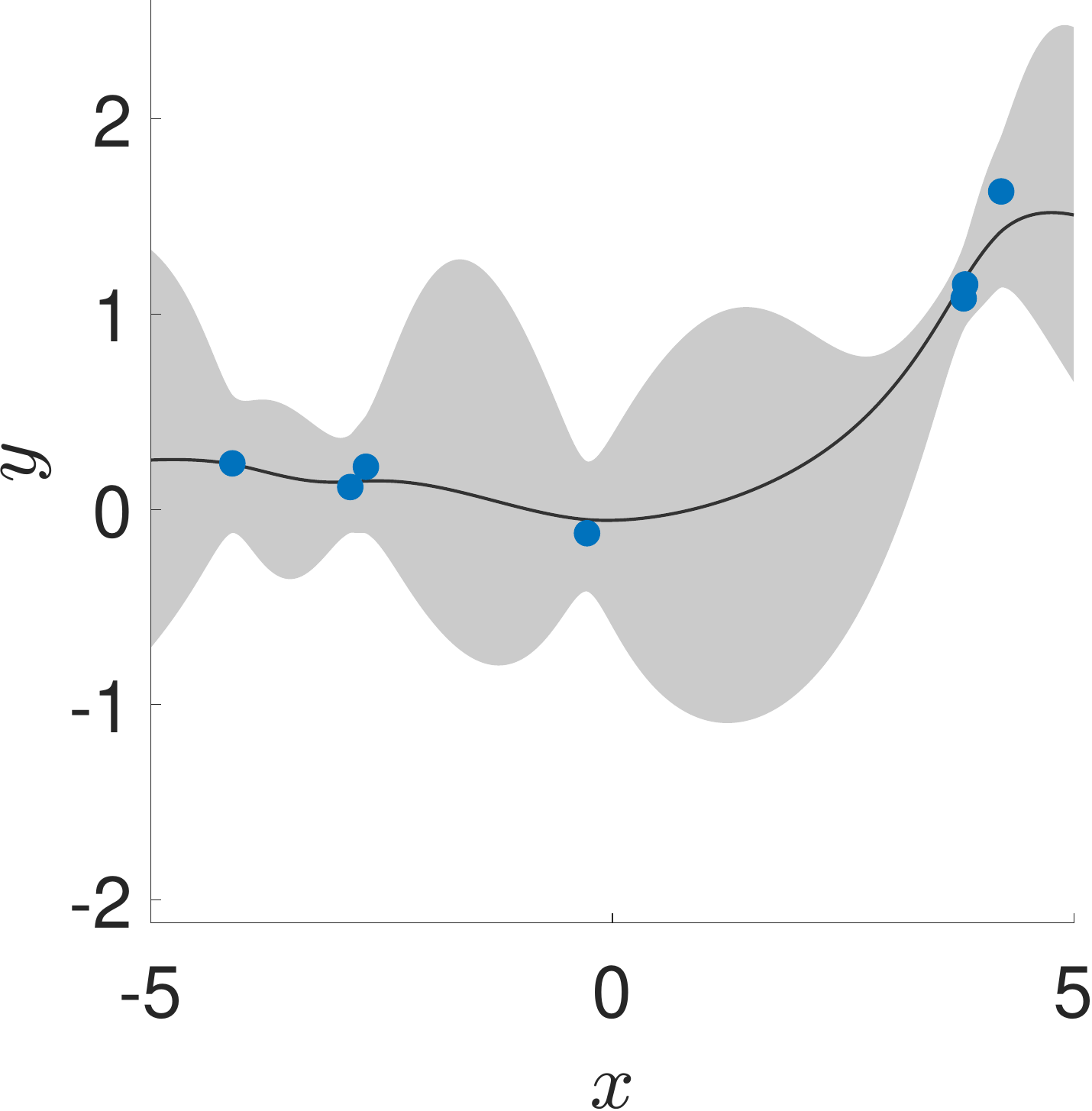}
    \caption{On top left, the posterior distribution $p(\Theta \mid Y,X)$ in Eq.~\eqref{eq:partitionLikelihood} as a function of the noise standard deviation $\sigma_{1,\varepsilon}$ and length scale $l_{1,1}$ for 7 data points from a single zero-mean Gaussian process expert with $\left( \sigma_{1,\varepsilon}, \sigma_{1,f}, l_{1,1} \right) = \left( 0.1^{1/2}, 1, 1 \right)$. The local maxima of the posterior distribution are illustrated with blue and red stars. On top right, the 7 data points in blue together with the true predictive mean and 95\% interval in black and gray. Below, from left to right, the predictive mean and 95\% intervals corresponding to the short length scale local minimum, long length scale local minimum, and marginalization over the posterior distribution $p(\Theta \mid Y, X)$.}
    \label{im:examplePredictivePosterior}
\end{figure}

\begin{figure}[!t]
    \centering
    \includegraphics[width = \linewidth]{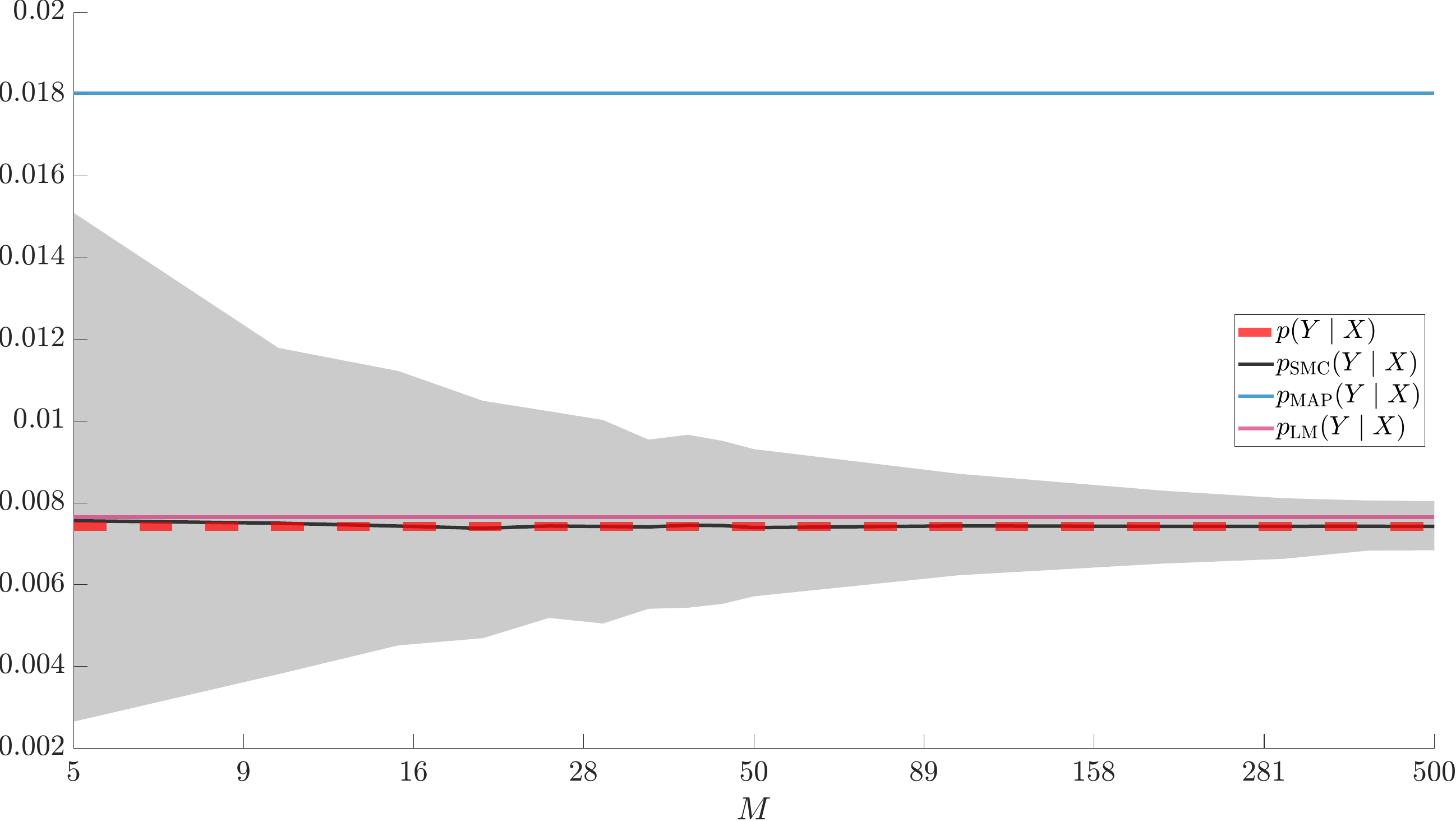}
    \caption{The marginalized likelihood estimates $p_\mathrm{MAP}( Y \mid X)$, $p_\mathrm{LM}( Y \mid X)$, and $p( Y \mid X)$ using the MAP estimate (blue line), long length scale local minimum (red line), and high-accuracy numerical integration (dashed red lines), respectively. The MAP estimate leads to overfitting and high likelihood for the training data.  The mean $p_\mathrm{SMC}( Y \mid X)$ and 95\% confidence intervals given by repeated SMC runs with different particle amounts $M$ shown in black and gray, respectively, match closely with numerical integration.}
    \label{im:exampleMarginalLikelihood}
\end{figure}

Estimating the marginalized likelihood $ p(Y \mid X, C) $ could be approached using MAP estimates, similarly to Eq.~\eqref{eq:weightsImportanceSampling}.
However, the MAP estimates have certain disadvantages \cite[p. 136]{williams2006gaussian}.
First, given multiple modes, optimization is not guaranteed to converge to the MAP estimate.
Second, the uncertainty for the predictions is misspecified.
Lastly, the resulting marginal likelihood estimates are biased.
To clearly illustrate this, consider the simplified case of $ K = 1 $ expert.
We are interested in probing the posterior $p(\Theta \mid Y, X)$, and in particular computing the normalizing constant $p(Y\mid X)$.
The predictive distribution and marginal likelihood estimates are shown in Figures \ref{im:examplePredictivePosterior} and \ref{im:exampleMarginalLikelihood}, respectively.
Figure \ref{im:examplePredictivePosterior} shows the posterior distribution $p(\Theta \mid Y,X)$ in Eq.~\eqref{eq:partitionLikelihood} as a function of the length scale and noise standard deviation parameters for a GP with 7 data points.
The posterior distribution exhibits two local minima: one with smaller length scale and noise standard deviation parameters, and one with larger length scale and noise standard deviation.
The corresponding GP predictive means and 95\% predictive intervals are misspecified (first two plots in the bottom row of Figure \ref{im:examplePredictivePosterior}).
In contrast, the marginalized predictive mean and 95\% predictive intervals (last plot in the bottom row of Figure \ref{im:examplePredictivePosterior}) more closely follow the data-generating predictive mean and predictive intervals, showing the benefit of marginalization over a plug-in MAP approximation.
Figure \ref{im:exampleMarginalLikelihood} shows the biases resulting from using either of the local minima 
as surrogates for marginalizing $p(\Theta\mid Y, X)$ together with the mean and 95\% confidence intervals for the marginal likelihood estimates using SMC with varying particle amounts, $M$.
The SMC samplers are run 500 times for each $M$.
The marginal likelihood estimates provided by the SMC samplers show no bias and consistently converge towards the correct marginalized likelihood estimate.
The marginalized likelihood estimate for $ p(Y \mid X) $ has been computed using high-accuracy numerical integration.
The above points led us to construct the likelihood-tempered inner SMC sampler to consistently estimate $ p(Y \mid X, C) $.

We build a sequence of intermediate tempered conditional posterior distributions for the parameters $C$ and $\Psi$
as follows
\begin{equation}
    \pi_t( C, \Psi \mid X, Y) \propto p_t(Y \mid X, C) p(C \mid X, \Psi) \pi_0( \Psi ),
    \label{eq:gatingPartitionTemperedPosterior}
\end{equation}
with
\begin{equation}
    p_t( Y \mid X, C) := \int p( Y \mid X, C, \Theta)^{\kappa^{(t)}} \pi_0(\Theta) \, \text{d}\Theta = \prod\limits_{k = 1}^K \int p\left( Y_k \mid X_k, \Theta_k \right)^{\kappa^{(t)}}\hspace{-0.1cm}\pi_0( \Theta_k) \, \text{d}\Theta_k,
    \label{eq:temperedMarginalLikelihood}
\end{equation}
and where $t$ stands for the ``time'' step of the SMC or SMC$^2$ sampler and $\kappa^{(t)}$ is a strictly increasing sequence of parameters, $ 0 = \kappa^{(0)} < \dots < \kappa^{(t-1)} < \kappa^{(t)} < \dots < \kappa^{(T)} = 1 $, controlling the degree of tempering.
We discuss the explicit construction of the tempering sequence, which depends on the outer SMC sampler, in the following Section.

\begin{figure}[!t]
\begin{center}
    \includegraphics[width = 0.8\textwidth]{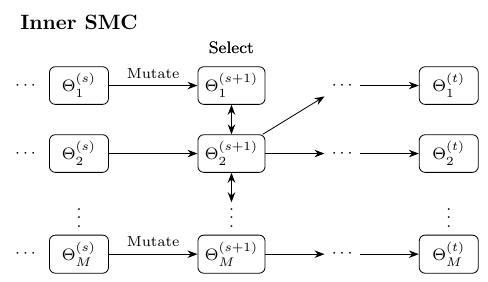}
    \caption{An illustration of the inner SMC sampler ${\Upsilon}( \Theta_{1:M}^{(0:t)}, A^{(t)} \mid C, X, Y)$. 
    The SMC sampler uses $M$ particles $\Theta_{1:M}^{(0:t)}$ to construct an approximation for the posterior distribution of the GP expert parameters at each iteration corresponding to the current level of tempering $\kappa^{(t)}$. The inner SMC sampler provides an unbiased estimator for the marginal likelihood $p_t( Y \mid X, C)$ in Eq.~\eqref{eq:temperedMarginalLikelihood} according to Eq.~\eqref{eq:Z_tM}. The auxiliary parameters $A^{(t)}$ appear in the mutation and selection steps and are not used elsewhere or retained.}
    \label{im:schematicInnerSMC}
\end{center}
\end{figure}

The inner SMC sampler allows us to unbiasedly estimate the intractable tempered marginal likelihood  $p_t( Y \mid X, C)$ by an ensemble $\{\Theta_{m}^{(t)}\}_{m=1}^{M}$ for any outer SMC particle $(C,\Psi)$, as follows.
We have the tempered unnormalized posterior
\begin{equation}\label{eq:pit}
    \pi_t( Y, C, \Psi, \Theta) = \pi( C,\Psi,\Theta \mid X )\prod_{s=1}^t p( Y \mid X, C, \Theta)^{ \kappa^{(s)} - \kappa^{(s - 1)} } .
\end{equation}
Now, for any given $( C, \Psi)$, we can construct a non-negative and unbiased estimator of the joint $\pi_t( Y, C, \Psi)$ as %
\begin{equation}
    \hat \pi_t( Y, C, \Psi) = \widehat{Z}^M_t(Y \mid X, C) \pi( C, \Psi \mid X) \, , \label{eq:est}
\end{equation}
where 
\begin{equation}
    \widehat{Z}^M_t( Y \mid X, C) := \prod_{s = 1}^t \frac{1}{M} \sum_{m=1}^M p( Y \mid X, C, \Theta^{(s-1)}_m)^{ \kappa^{(s)} - \kappa^{(s-1)} } \, ,
    \label{eq:Z_tM}
\end{equation}
is a non-negative and unbiased estimator \citep{Moral:2004, Dai:2022} of $p_t( Y \mid X, C)$ in Eq.~\eqref{eq:temperedMarginalLikelihood} with $\Theta^{(s)}_m$ being the output from an SMC with fixed $C$.
In other words, the initial particles $ \Theta^{(0)}_m$ are i.i.d from the prior distribution $ \Theta^{(0)}_m \sim \pi_0(\Theta) $. 
For subsequent iterations $ t = 1, \dots, T$, the particles are obtained as $ \Theta^{(t)}_m \sim \mathcal{M}_t(\widehat{ \, \Theta}^{(t)}_m,\, \cdot) $, where $\mathcal{M}_t$ is an MCMC kernel, $\widehat{\Theta}^{(t)}_m = \Theta_{I_m^{(t)}}^{(t)}$ are resampled particles, and $I_m^{(t)} \sim \text{Categorical}({\bf \omega}^{(t)})$ are resampling indices for $m = 1,\dots, M$, with ${\bf \omega}^{(t)} = ( \omega^{(t)}_1, \dots, \omega^{(t)}_M) $ and the individual particle weight are given as $ \omega^{(t)}_m \propto p( Y \mid X, C, \Theta^{(t-1)}_m)^{ \kappa^{(t)} - \kappa^{(t-1)} } $.
Here $\mathcal{M}_t$ is an MCMC kernel which keeps
\begin{equation}
    p_t( \Theta \mid C, X, Y) \propto \pi_t( Y, C, \Psi, \Theta)
\end{equation}
invariant.
Let $A^{(t)}$ denote $I_{1:M}^{(1:t)}$ and all the auxiliary variables of the SMC sampler for $\Theta$.
The distribution
\begin{equation}
    \pi_t( C, \Psi, \Theta_{1:M}^{(0:t)}, A^{(t)} \mid X,Y) \propto \widehat{Z}^M_t( Y \mid X, C) \pi( C, \Psi \mid X) {\Upsilon}( \Theta_{1:M}^{(0:t)}, A^{(t)} \mid C, X, Y),
    \label{eq:extendedSpaceTargetDistribution}
\end{equation}
where ${\Upsilon}( \Theta_{1:M}^{(0:t)}, A^{(t)} \mid C, X, Y)$ is the distribution of the SMC sampler given $C$ (up to time $t$), is such that for any $ m = 1, \dots, M $ and any $ t = 0, \dots, T $,
\begin{equation}\label{eq:jointposterior}
    \int \varphi( C, \Psi, \Theta) \, \text{d}\pi_t( C, \Psi, \Theta \mid X, Y) = \int \varphi( C, \Psi, \Theta^{(t)}_m) \, \text{d}\pi_t( C, \Psi, \Theta_{1:M}^{(0:t)}, A^{(t)} \mid X, Y).
\end{equation}
In other words, it has the target of interest as its marginal.
We note that the SMC sampler ${\Upsilon}( \Theta_{1:M}^{(0:t)}, A^{(t)} \mid C, X, Y)$ is independent of $\Psi$, hence its omission.
We present a schematic in Figure \ref{im:schematicInnerSMC} and pseudo-code in Algorithm \ref{alg:smc} for the SMC sampler ${\Upsilon}( \Theta_{1:M}^{(0:t)}, A^{(t)} \mid C, X, Y)$.
In the following Section, we combine the above inner SMC sampler with sampling of the gating network parameters $\Psi$ and partition $C$.  
\begin{algorithm}[!t]
\caption{SMC for fully Bayesian GP estimation.}
\label{alg:smc}
\begin{algorithmic}
\State \textbf{Initialize:}
    \State Draw $\Theta^{(0)}_m \sim \pi_0( \Theta )$.
    \State Set $\omega_m^{(0)} = \frac{1}{M}$ for $m=1,\dots,M$.
    \State Set $\widehat{Z}_{0}^M = 1$.
    \State
\State \textbf{Estimation of $\{(\Theta,\omega)_m\}_{m=1}^M \approx p_t(\Theta \mid X,Y) \propto p(Y \mid X,\Theta)^{ \kappa^{(t)} }\pi_0(\Theta)$ 
and $\pi(Y \mid X)$:}
\vspace{10pt}
\While{$\kappa^{(s)} < \kappa^{(t)}$}
    \State $s = s + 1$.
    \State Define $\overline \omega^{(s)}_m := p(Y \mid X,\Theta_m^{(s-1)})^{\kappa^{(s)} - \kappa^{(s-1)}}$
    \State Set $\widehat{Z}_{s}^M \leftarrow \left(\frac1M\sum_{m=1}^M \overline \omega^{(s)}_m\right)\widehat{Z}_{s-1}^M$, following Eq.~\eqref{eq:Z_tM}.
    \State Select new indices $I_m^{(s)}$ according to $\omega^{(s)}_m \propto \overline \omega^{(s)}_m$
    \State Draw $ \Theta^{(s)}_m \sim \mathcal{M}_s( \Theta^{(s-1)}_{I_m^{(s)}},\, \cdot)$, 
    where $\mathcal{M}_s$ is an MCMC kernel targeting $p_s(\Theta \mid X,Y)$.
    \State Set $\omega_m^{(s)} = \frac{1}{M}$ for $m=1,\dots,M$.
\EndWhile
\end{algorithmic}
\end{algorithm}
\section{SMC\texorpdfstring{$^2$}{TEXT}}
\label{sec:SMC2}
\begin{figure}[!t]
\begin{center}
    \includegraphics[width = .8\textwidth]{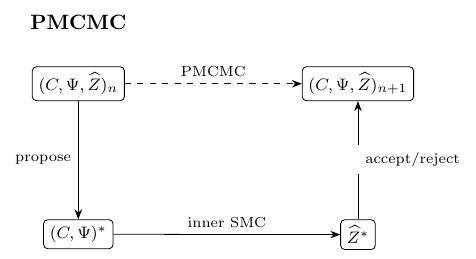}
    \caption{An illustration of one step of the particle Markov chain Monte Carlo (PMCMC) method, for simulating from the posterior gating network and partition $\pi_t( C, \Psi \mid X, Y)$. The inner SMC sampler ${\Upsilon}( \Theta_{1:M}^{(0:t)}, A^{(t)} \mid C^*, X, Y)$ as shown in Figure~\ref{im:schematicInnerSMC} delivers an unbiased estimator $\widehat{Z}^M_t( Y \mid X, C^*)$ in Eq.~\eqref{eq:Z_tM} of $p_t( Y \mid X, C^*)$ according to Eq.~\eqref{eq:temperedMarginalLikelihood}, which is used to decide whether to accept the proposal, following Eq.~\eqref{eq:acceptanceProbabilityPMMH}. One or all of the $M$ inner SMC particles $\Theta_{1:M,n}^{(0:t)}$~\eqref{eq:temperedMarginalLikelihood} associated with $\widehat{Z}^{M,n}_{t}( Y \mid X, C^n)$ can be retained for estimation of the joint $\pi_t( C, \Psi, \Theta \mid X, Y)$.}
    \label{im:pmcmc}
\end{center}
\end{figure}

The inner SMC sampler presented in the previous Section is used as the MCMC proposal for the outer SMC sampler.
Targeting the tempered extended posterior distribution $\pi_t( C, \Psi, \Theta_{1:M}^{(0:t)}, A^{(t)}, \mid X, Y)$ using MCMC is called particle MCMC (PMCMC)  and is a particular instance of the pseudo-marginal method \citep{Andrieu:2009}.
When the MCMC method used is Metropolis-Hastings, it is called particle marginal Metropolis-Hastings (PMMH) \citep{Doucet:2010}.
In particular, letting $n$ denote the index of the MCMC, the proposal consists of sampling $( C, \Psi)^* \sim \mathcal{Q}( ( C, \Psi)_n, \cdot)$ and then $( \Theta_{1:M}^{(0:t)})^* \sim {\Upsilon}( \Theta_{1:M}^{(0:t)}, A^{(t)} \mid C^*, \Psi^*)$.
The proposal $( C, \Psi, \Theta_{1:M}^{(0:t)})^*$ is then accepted with probability 
\begin{equation}
    \frac{\widehat{Z}^M_t( Y \mid X, C^*) \pi( ( C, \Psi)^* \mid X) \mathcal{Q}( ( C, \Psi)^*, ( C, \Psi)_n)} {\widehat{Z}^M_t( Y \mid X, C_n) \pi( ( C, \Psi)_n \mid X) \mathcal{Q}( ( C, \Psi)_n, ( C, \Psi)^*)}.
    \label{eq:acceptanceProbabilityPMMH}
\end{equation}
We denote the resulting MCMC kernel by $\mathcal{H}_t$ and note that by design $\pi_t \mathcal{H}_t = \pi_t$.
We target $\pi_t( C^{(t)}, \Psi^{(t)}, \Theta_{1:M}^{(0:t)}, A^{(t)} \mid X,Y)$ for $ t = 1, \dots, T$ using an SMC sampler \citep{Moral:2006}, where the MCMC mutation at time $t$ is given by the PMMH kernel $\mathcal{H}_t$ (indices $t$ are introduced on the variables for convenience and to avoid confusion).
This method is called SMC$^2$ owing to the nested SMCs \citep{chopinsmc2}.
We present pseudo-code for the PMCMC proposal in Algorithm \ref{alg:pmcmc}.
An illustration of the PMCMC proposal flow is illustrated in Figure \ref{im:pmcmc}. 
\begin{algorithm}[!t]
\caption{PMCMC step for $p_t( C, \Psi, \Theta \mid X, Y) \propto p(Y \mid X, C, \Theta)^{ \kappa^{(t) }} p( C \mid X, \Psi) \pi_0( \Psi, \Theta)$}
\label{alg:pmcmc}
\begin{algorithmic}
    \State Begin with $(C,\Psi,\Theta^{(0:t)}_{1:M}) \sim \pi_t( C, \Psi, \Theta_{1:M}^{(0:t)}, A^{(t)} \mid X,Y)$ and $\widehat{Z}_t^M( Y \mid X, C)$.
    \State Propose $( C, \Psi)^* \sim \mathcal{Q}( ( C,\Psi), ~\cdot~)$. 
    \State Propose $\Theta^{*,(0:t)}_{1:M} \approx p_t( \Theta \mid X, Y, C^*) \mapsto \widehat{Z}^M_t( Y \mid X, C^*)$, using SMC Algorithm \ref{alg:smc}.
    \State Accept $ \left(C,\Psi, \Theta^{(0:t)}_{1:M} \right)^*$ (and store $\widehat{Z}^M_t( Y \mid X, C^*)$) with probability
    $$
    1 \wedge \frac{\widehat{Z}^M_t( Y \mid X, C^*) \pi( ( C, \Psi)^* \mid X) \mathcal{Q}( ( C, \Psi)^*, ( C, \Psi))} {\widehat{Z}^M_t( Y \mid X, C) \pi( ( C, \Psi) \mid X) \mathcal{Q}( ( C, \Psi), ( C, \Psi)^*)} \, .
    $$
\end{algorithmic}
\end{algorithm}

In particular, we approximate Eq.~\eqref{eq:extendedSpaceTargetDistribution} using the particles $\{ C_j^{(t)}, \Psi_j^{(t)}, ( \Theta_{1:M}^{(0:t)})_j\}_{j = 1}^J$ with weights $\{ w_j^{(t)}\}_{j = 1}^J$.
Note that we do not need the auxiliary variables $A^{(t)}_j$, as they will always be integrated out.
In fact, for estimates we also only need one $\Theta_{j,m}^{(t)}$ for each $j$, sampled uniformly from $\left(\Theta_{j,1}^{(t)},\dots \Theta_{j,M}^{(t)} \right)$, however in principle one can average over the ensemble, which is a Rao-Blackwellisation of the former.
Furthermore, if we do not need to estimate functions of $\Theta$, then we only need the normalizing constant estimates $\widehat{Z}^M_t( Y \mid X, C_j^{(t)})$ along with the $( C, \Psi)_j^{(t)}$ particles.
For $J$ particles, estimating the tempered posterior in Eq.~\eqref{eq:gatingPartitionTemperedPosterior} using the model on the extended state space given in Eq.~\eqref{eq:extendedSpaceTargetDistribution}, the particle weights are recursively related according to
\begin{equation}
   w_j^{(t)} \propto \frac{ \widehat{Z}^M_t( Y \mid X, C_j)} { \widehat{Z}^M_{t - 1}( Y \mid X, C_j)} w_j^{(t - 1)} = \sum\limits_{m = 1}^M p_t( Y \mid X, C_j, \Theta_m^{(t - 1)})^{ \kappa^{(t)} - \kappa^{(t-1)}} w_j^{(t - 1)},
    \label{eq:weight_update}
\end{equation}
where the initial particle weights are set as $w_j^{(0)} = \frac{1}{J}$.
However, the recursive update to the particle weights degrades the sample distribution. This degradation can be quantified using the effective sample size (ESS):
\begin{equation}
    J_{\text{ESS}}^{(t)} = \frac{1}{\sum\limits_{j = 1}^J \left( w_j^{(t)} \right)^2}.
    \label{eq:ess}
\end{equation}
We determine $\kappa^{(t)}$ adaptively so that the relative reduction in the ESS is approximately a predefined learning rate, $\eta$.
This adaptive scheduling allows one to control the sample degeneracy at each iteration $t$ by keeping the change between the subsequent target posterior distributions $\pi_{t-1}( C, \Psi, \Theta_{1:M}^{(0:t-1)}, A^{(t-1)} \mid X,Y)$ and $\pi_t( C, \Psi, \Theta_{1:M}^{(0:t)}, A^{(t)} \mid X,Y)$ approximately constant.
Overall, smaller learning rates allow for more aggressive exploration of the target distribution with fewer iterations $T$ and higher learning rates lead to more iterations $T$, albeit with more similar subsequent target posterior distributions, resulting in easier inference.
In \cite{syed2024optimised}, this is made precise and it is shown that there is a sufficient
density of tempering beyond which the algorithm behaves well, and below which it breaks catastrophically.
We use $\eta = 0.9$, which delivers good performance for our problem.
More specifically, the next value in the tempering sequence, $\kappa^{(t)}$, is determined by solving a minimization problem:
\begin{equation}
    \kappa^{(t)} = \underset{\kappa_t}{\arg \min}\left\vert \frac{ \sum\limits_{j = 1}^J \left( \sum\limits_{m = 1}^M p_t( Y \mid X, C_j, \Theta_m^{(t - 1)})^{ \kappa_t - \kappa^{(t-1)}} w_j^{(t - 1)} \right)^2 }{J_{\text{ESS}}^{(t - 1)}} - \eta \right\vert,
\end{equation}
where the numerator is the ESS of the current time step according to $\kappa_t$ and with the constraint $\kappa^{(t)} > \kappa^{(t- 1)}$.
In particular, we resample the particles after each iteration $t$ of the algorithm for both the inner and outer SMC samplers.
One can save a bit of variance by using weighted particles but this was not found to be significant for our numerical examples.

Resampling introduces duplicates of the particles into the sample.
The MCMC mutation with $\mathcal{H}_t$ counteracts this.
In particular, we propose new $\Psi^*_{1:J}$ particles by separately proposing the gating network means and standard deviations from the weights.
We construct the random walk proposal for the means and standard deviations as 
\begin{equation}
     (\mu_{1:K}, \Sigma_{1:K})^*_{j} = (\mu_{1:K}, \Sigma_{1:K})_{j} + \xi_{\mu, \Sigma,j}
\end{equation}
where $\xi_{\mu, \Sigma,j}\sim \mathcal{N}( 0, \Sigma_{\mu, \Sigma} )$ with $\Sigma_{\mu, \Sigma} \in \mathbb{R}^{ N_{\mu,\Sigma} \times N_{\mu,\Sigma} }$ and $ N_{\mu,\Sigma} = 2 DK$, being the weighted empirical covariance matrix of the $J$  particles, $ (\mu_{1:K}, \Sigma_{1:K})_{1:J}$.
We enforce positivity of the variance parameters by rejecting negative proposals due to our choice of prior distributions.
The gating network weights, as in \cite{Lukens:2020}, are proposed as
\begin{equation}
    \log \nu_{1:K,j}^* = \log \nu_{1:K, j} + \xi_{\nu,j}
\end{equation}
where $\xi_{\nu,j}\sim \mathcal{N}( 0, \Sigma_{\nu} )$ with $\Sigma_{\nu} \in \mathbb{R}^{ N_{\nu} \times N_{v} }$ and $ N_{v} = K$, being the weighted empirical covariance matrix of the $J$ particles of the weights.

The proposal for the partition $ C^*$ given the proposed gating network parameters $\Psi^*$ follows the prior, that is $ c^*_i \sim \text{Categorical}(\boldsymbol{p}(x_i \mid \Psi^*))$ independently for $i=1, \ldots, n$. 
This leads to convenient cancellations in the acceptance probability.
In cases when the clusters are not well separated by the inputs, we may need better proposals for the partitions.
Possible alternatives are single-site Gibbs sampling \citep{Rasmussen:2002} or split-merge moves which update multiple allocation variables simultaneously, first appearing in \cite{Jain2004} and more recently in \cite{Bouchard:2017}.

For the inner SMC, we use similar MCMC updates as for the outer SMC.
The proposals are constructed independently for each of the $K$ experts.
We propose new $ \Theta_{k,j,m}^*$ particles again with a random walk proposal given as
\begin{equation}
    \Theta_{k,j,m}^* = \Theta_{k,j,m} + \xi_{\Theta_{k,j},m}
\end{equation}
where $ \xi_{\Theta_{k,j},m} \sim \mathcal{N}( 0, \Sigma_{ {\Theta_{k,j}} })$ with $ \Sigma_{ {\Theta_{k,j}} } \in \mathbb{R}^{ N_\theta \times N_\theta }$ and $N_\theta = D + 3$, being the weighted empirical covariance matrix of the $M$ inner  particles, $\left( \Theta_{k,j,1}, \dots, \Theta_{k,j,M} \right)$, corresponding to the GP parameters of the $k$th expert.
Positivity of the variance parameters is again enforced via our choice of prior distributions.

\begin{figure}[!t]
\begin{center}
    \includegraphics[width = 0.75\textwidth]{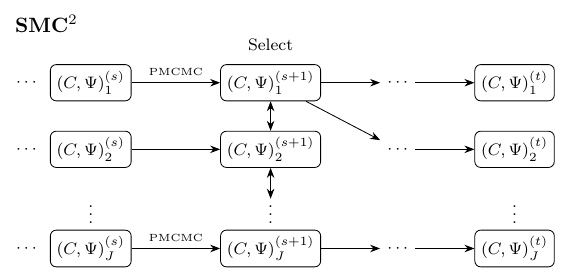}
    \caption{An illustration of the SMC$^2$ sampler for mixtures of Gaussian process experts. The outer SMC sampler uses the $J$ particles for the partitions $C^{(t)}_j$ and gating network parameters $\Psi^{(t)}_j$. Each outer particle is associated with an inner SMC sampler ${\Upsilon}( \Theta_{1:M}^{(0:t)}, A^{(t)} \mid C, X, Y)$ which provides an unbiased estimate for the marginal likelihood $p_t( Y \mid X, C_j)$ in Eq.~\eqref{eq:temperedMarginalLikelihood} according to Eq.~\eqref{eq:Z_tM}. The unbiased marginal likelihood estimate can then be used as a PMMH proposal for the MCMC acceptance ratio in Eq.~\eqref{eq:acceptanceProbabilityPMMH}, as shown in Fig.\ref{im:pmcmc}.}
    \label{im:schematicSMC2}
\end{center}
\end{figure}

We use a criterion to automatically adapt the number of MCMC steps taken to mutate the particles suggested by \cite{chopinsmc2:mcmcmSteps}.
Note that the exact form presented below is defined only in the book's complementary software.
Particularly, we perform MCMC steps until a relative decrease in a distance metric is below a threshold $\delta$
\begin{equation}
    \dfrac{ \left\vert d^{(n)} - d^{(n-1)} \right\vert }{ d^{(n - 1)} } < \delta,
\end{equation}
where
\begin{equation}
    d^{(n)} = \frac{1}{N_\rho} \sum\limits_{i = 1}^{N_\rho} \left\Vert \rho_i^{(n)} - \rho_i^{(0)} \right\Vert_2,
\end{equation}
and $\|\cdot \|_2$ denotes Euclidean norm.
Here, $\rho$ denotes the set of parameters being updated in the MCMC, i.e. $\rho =(\Psi_j, C_j)$ in the outer SMC and $\rho = (\Theta_{k,j,m})$ for the inner SMC.
For the $n$th MCMC step,  $\rho_i^{(n)}$ denotes the $i$th of $N_\rho$ particles for the inner SMCs, while for the outer SMC, $\rho_i^{(n)}$ denotes the $i$th of $N_\rho$ unbiased estimates of the likelihood in Eq.~\eqref{eq:est}.
For the outer SMC, we monitor the change in the likelihood in Eq.~\eqref{eq:est} as it is invariant to label-switching.
We present a schematic for the full SMC$^2$ sampler in Figure \ref{im:schematicSMC2}.

Before ending this section, it is relevant to reflect on the complexity of this method.
Just as with naive importance sampling, the cost is ultimately dominated by the Gaussian process likelihoods with complexities proportional to $\mathcal{O}(K N^3/K^3)$, assuming $N / K$ data points for each of the $K$ experts, which can be distributed across the $K$ experts.
The likelihood needs to be estimated for each of $M$ inner and $J$ outer $\Theta_{j,m}$ particles at each time step $t = 0, \dots, T$.
For the mutation PMCMC step, this must be done $t$ times at step $t$, that is when the clustering $C$ changes we must initialize the particles $\Theta_{j,m} \sim \pi_0(\Theta)$ from the prior and rerun the inner SMC sampler.
Therefore, the total number of likelihood estimations is $ M J \sum_{t=0}^T t = \mathcal{O}( T^2 M J)$, and the total cost is $\mathcal{O}(T^2 M J N^3/K^2)$.
This seems expensive, particularly if one makes the natural choice $ T, M, J \propto N$, see for example \cite{chopinsmc2, Beskos:2014, Chopin:2020}, but the point is that, in comparison to naive importance sampling, $N^4$ is much less than the number samples required in importance sampling, e.g. $N^4 \ll K^N$ in the example of Section \ref{sec:importanceSampling}.
Moreover, many evaluations can be performed in parallel at either the level of the inner particles, the $K$ independent GP experts, or by running parallel inner SMC samplers.

In terms of memory, keeping track of each inner particle and their ancestor particles requires $2 + D$ GP kernel hyperparameters for each of of the $K$ experts at each time step $t = 0, \dots, T$, resulting in a total of $(2K + DK) M T$ numbers to be kept in memory.
This is repeated for each of the $J$ outer particles which, along with the inner particles, require $1 + 2D$ gating network parameters for each of the $K$ experts and $N$ labels for the partitions $C_j$ again at each time step $t = 0, \dots, T$.
This results in a total memory use of $\mathcal{O}( (2K + DK) M T + (K + 2DK + N)JT )$.
However, we do not use the intermediate results in our predictive inference. 
Thus, we are able to leave out $T$ and simplify the memory complexity to $\mathcal{O}( (2K + DK)M + (K + 2DK + N)J)$ and keep track of particles only at the current iteration.
\begin{figure}
    \centering
    \begin{minipage}{0.49\textwidth}
    \subfloat[][]{ \includegraphics[width = \textwidth]{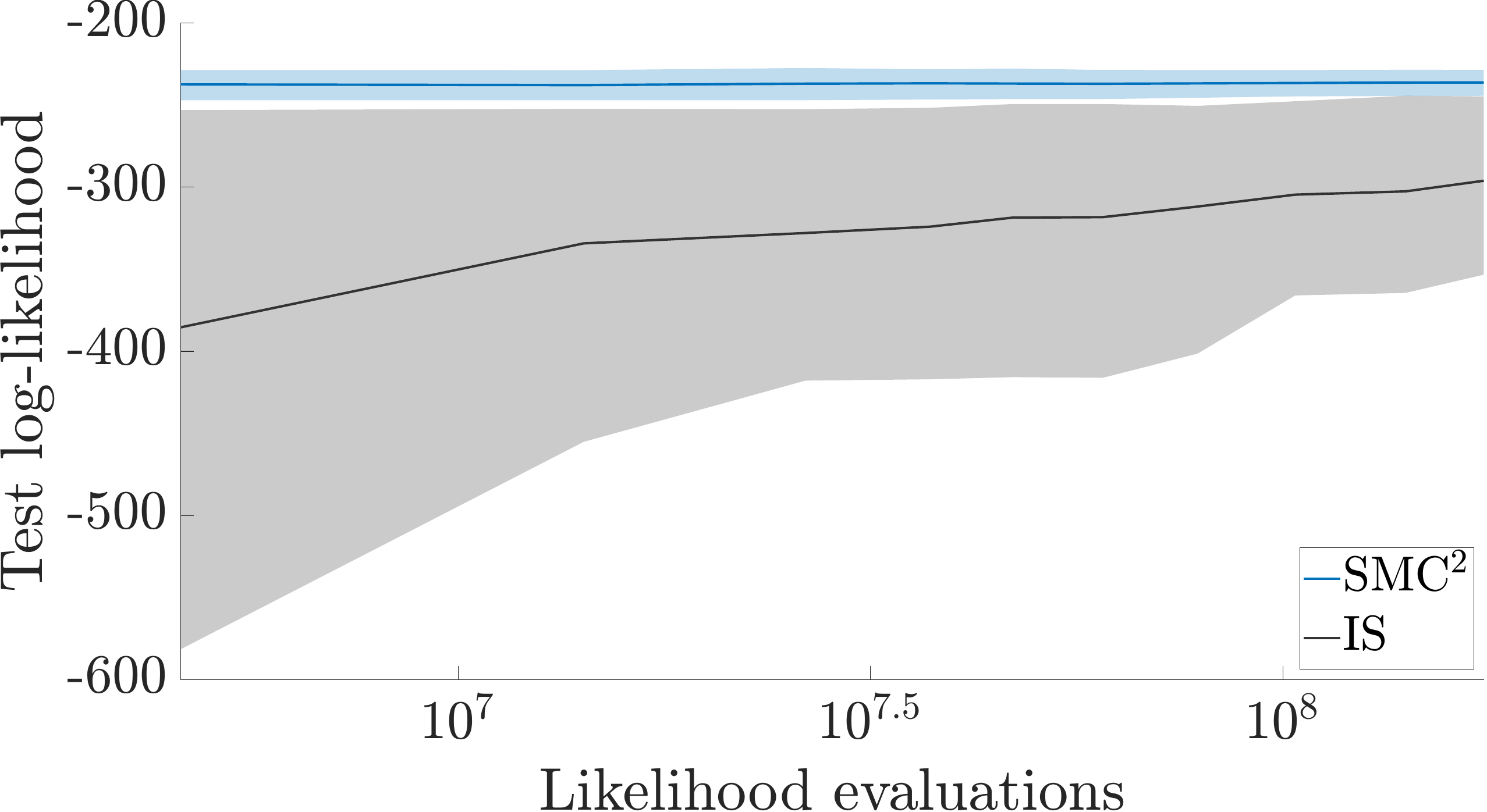} }
    \end{minipage}
    \begin{minipage}{0.49\textwidth}
    \subfloat[][]{ \includegraphics[width = \textwidth]{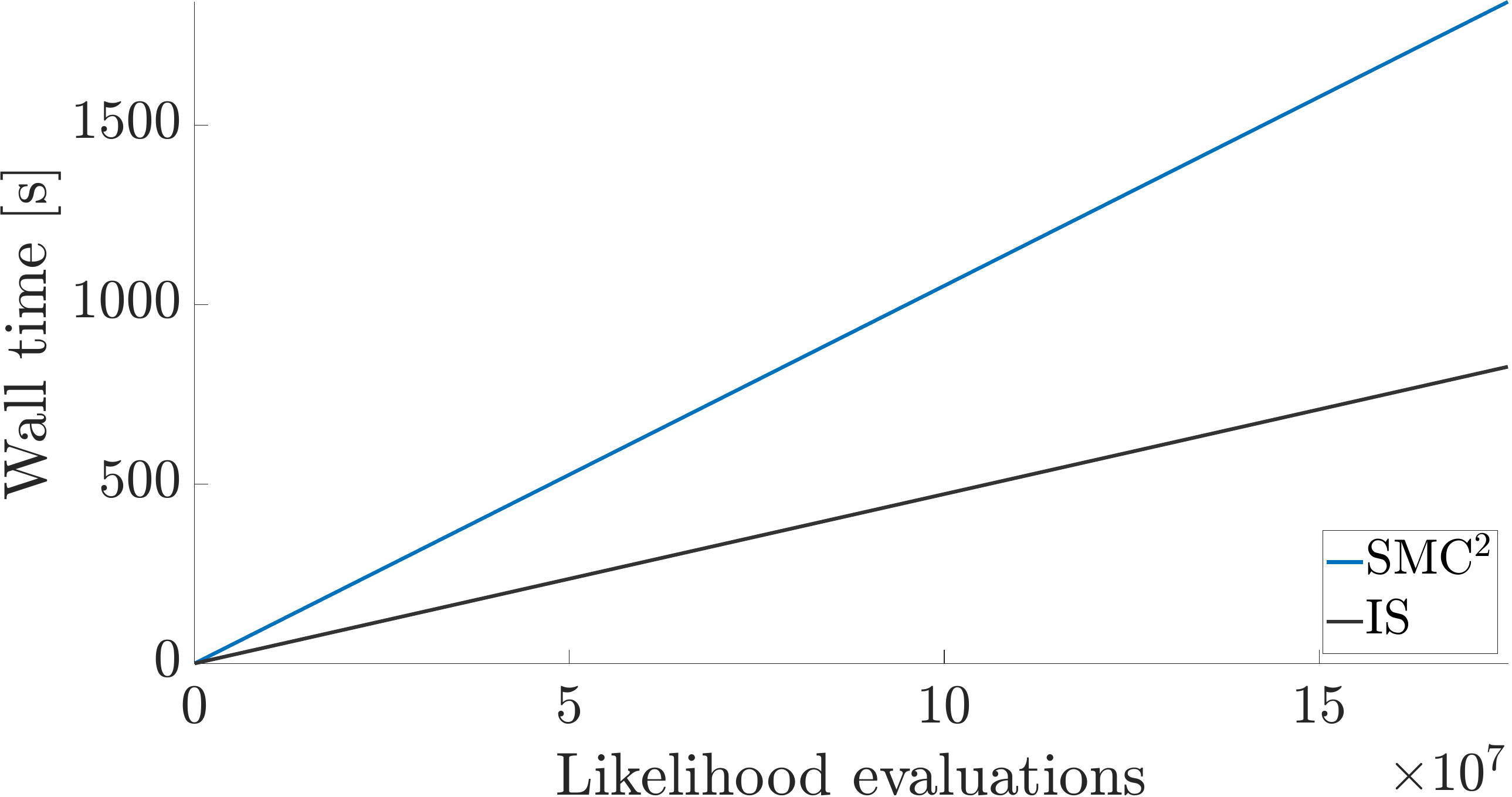} }
    \end{minipage}
    \caption{Efficiency of the methods, measured in terms of the test log-likelihood score (a) and wall-clock time (b), as a function of the total number of likelihood evaluations.}
    \label{im:testLogLikelihood_comparison_IS_SMC2}
\end{figure}

The efficiency of the methods in terms of number of likelihood evaluations is demonstrated in Figure \ref{im:testLogLikelihood_comparison_IS_SMC2}. 
From the left panel it is clear that SMC$^2$ converges quickly and with a small spread, whereas IS has a large spread and converges more slowly.
IS is embarrassingly parallel, while there one expects a negligible hardware-dependent communication overhead for SMC$^2$ in between likelihood evaluations.
We consider likelihood evaluations to be a fair hardware-agnostic proxy for wall-clock time, given that communication arises at higher order.
The right panel shows median wall-clock time of the two methods as a function of the likelihood evaluations on a fixed computer architecture with 64 physical central processing unit cores of an AMD Ryzen Threadripper 3990X.
Note that this architecture limits the parallelism of both methods equally.
Here, SMC$^2$ is slower by a factor of $2.2$.
For further comparisons and discussion on run time for IS versus alternative approaches, such as robust Bayesian Committee Machine (RBCM) \citep{Deisenroth:2015, Liu:2018} and treedGP \citep{Gramacy:2007, Gramacy:2008, Gramacy:2010}, see the original study by \cite{Zhang:2019}.
\section{Predictive distribution}
\label{sec:predictiveDistribution}
\begin{algorithm}[!t]
\caption{SMC$^2$ sampled mixture of Gaussian process experts.}
\label{alg:smc2_moe}
\begin{algorithmic}
\State \textbf{Initialize:}
    \State \hspace{\algorithmicindent} Set $t = 0$ and $\kappa^{(t)} = 0$.
    \State \hspace{\algorithmicindent} Draw $U^{(0)}_j \sim p(C,\Psi, \Theta_1,\dots \Theta_M \mid X)$,
    \State \hspace{\algorithmicindent} Set $\widehat{Z}^M_{0}(C_j) = 1$.
    \State \hspace{\algorithmicindent} Set initial weights $\boldsymbol{w}^{(0)} = ( w_1^{(0)}, \dots, w_J^{(0)})$, where $w_j^{(0)} = \frac{1}{J}$.
    \State
\State \textbf{Estimation of $\pi( C, \Psi, \Theta \mid X, Y)$:}
\vspace{10pt}
\While{$\kappa^{(t)} < 1$}
    \State $t = t + 1$.
    \State Determine $\kappa^{(t)}$.
    \State Resample $U^{(t-1)}_j$ given particles weights $\boldsymbol{w}^{(t)}$, where $w_j^{(t)}$ follow Eq.~\eqref{eq:weight_update}.
    \State Set $w_j^{(t)} = \frac{1}{J}$.
    \State Run one step of the inner SMC sampler (Algorithm \ref{alg:smc}) for $U^{(t-1)}_j$ to obtain $\widehat{U}^{(t)}_j$. 
    \State Update $\widehat{Z}^M_{ t}(C_j)$ according to Eq.~\eqref{eq:Z_tM}.
    \State Draw $U^{(t)}_j \sim \mathcal{K}_t( \widehat{U}^{(t) }_j,\, \cdot)$ -- PMCMC kernel (Algorithm \ref{alg:pmcmc}) targeting $\pi_t$ in Eq. \eqref{eq:extendedSpaceTargetDistribution}.
\EndWhile
\State
\State \textbf{Prediction:}
\State \hspace{\algorithmicindent} Compute the predictive distributions according to Eq.~\eqref{eq:predictionPosterior}.
\end{algorithmic}
\end{algorithm}

We construct the predictive distribution by computing the individual GP predictive distributions for each expert given the partition $C_j$ and GP parameters $\Theta_{j,m}$, for every particle. These are then weighted by the particle weights and the gating network with parameters $\Psi_j$. 
Specifically, for each of the $K$ GPs, the predictive mean and variance for a given input $x^* \in \mathbb{R}^{D}$ is given by 
\begin{equation}
    \mathbb{E}_{y^*_{k,j,m}} := \mathbb{E}\left[y^* \mid x^*, X, Y, C_j, \Psi_j, m_{k,j,m}, \theta_{k,j,m} \right] = \Sigma_* \Sigma_X^{-1} ( Y_{k,j} - m_{k,j,m}) + m_{k,j,m},
    \label{eq:predictiveMean}
\end{equation}
and
\begin{equation}
    \text{Var}_{y^*_{k,j,m}} := \text{Var}\left[y^* \mid x^*, X, Y, C_j, \Psi_j, m_{k,j,m}, \theta_{k,j,m} \right] = \Sigma_{**} - \Sigma_* \Sigma_{X}^{-1} \Sigma_*^T,
    \label{eq:predictiveVariance}
\end{equation}
where $\Sigma_{**} = \Sigma( x^*, {x^*}; \theta_{k,j,m})$, $\Sigma_* = \Sigma( x^*, X_{k,j}; \theta_{k,j,m})$, and $\Sigma_{X} = \Sigma( X_{k,j}, X_{k,j}; \theta_{k,j,m})$ are used for brevity.
The individual normal marginal predictive distributions are combined into a mixture distribution:
\begin{equation}
    \pi( y^* \mid c^* = k, x^*, X, Y, C_j, \Psi_j ) = \sum\limits_{m = 1}^{M} \omega_m \mathcal{N}( y^* \mid \mathbb{E}_{y^*_{k,j,m}}, \text{Var}_{y^*_{k,j, m}}),
\end{equation}
which, in practice, would be evaluated over a grid of $y^{*}$ values for each $x^*$ value where prediction is desired.
These mixture distributions are further compounded across the $K$ experts according to
\begin{equation}
    \pi( y^* \mid x^*, X, Y, C_j, \Psi_j ) =\sum\limits_{k = 1}^{K} \pi( y^*_{j} \mid c^*=k, x^*, X, Y, C_j, \Psi_j ) p_k(x^* \mid \Psi_{j}),
\end{equation}
where the distributions are weighted according to the gating network defined in Eq.~\eqref{eq:gatingnetwork}.
Lastly, the distributions obtained for each $(C_j, \Psi_j)$ particle are added together according to their particle weights as
\begin{equation}
    \pi(y^* \mid x^*, X, Y) = \sum\limits_{j=1}^J w_j \pi( y^* \mid x^*, X, Y, C_j, \Psi_j).
    \label{eq:predictionPosterior}
\end{equation}
Similarly, the predictive mean is computed as
\begin{equation}
    \mathbb{E}[y^* \mid x^*, X, Y] = \sum\limits_{j=1}^J w_j \sum\limits_{k = 1}^{K}  p_k(x^* \mid \Psi_{j}) \sum\limits_{m = 1}^{M} \omega_m \, \mathbb{E}_{y^*_{k,j,m}}.
\end{equation}
Note that by resampling at each iteration $t$, the weights are $w_j = \frac{1}{J}$ and $\omega_i = \frac{1}{M}$. 
Pseudo-code for the method is shown in Algorithm \ref{alg:smc2_moe}.
\section{Numerical examples}
\label{sec:numericalExamples}
We demonstrate the method using multiple simulated and real-life one-dimensional data sets along with two multi-dimensional data sets.
The simulated one-dimensional sets of data consist of 1) a continuous function with homoskedastic noise and 2) a discontinuous function with heteroskedastic noise.
The first data set is generated using
\begin{equation}
    y_k = \sin(\pi x_k)\cos( (\pi x_k)^3 ) + \varepsilon(x_k)
    \label{eq:stationaryDataModel}
\end{equation}
where $\varepsilon(x_k) \sim \mathcal{N}( 0, 0.15^2)$ meaning that the measurement noise is homoskedastic.  In contrast, the second data set is generated by
\begin{equation}
    y_k = \begin{cases}
    \sin(60x_k) - 2 + \varepsilon_1(x_k), & x_k \leq 0.3 \\
    10 + \varepsilon_2(x_k), & 0.3 < x_k \leq 0.5 \\
    2\cos(4\pi x_k) - 10 + \varepsilon_3(x_k) & 0.5 < x_k \leq 1 
    \end{cases}
    \label{eq:nonstationaryDataModel}
\end{equation}
where $\varepsilon_1(x_k) \sim \mathcal{N}( 0, 0.5^2)$, $\varepsilon_2(x_k) \sim \mathcal{N}( 0, 0.25^2)$, and $\varepsilon_3(x_k) \sim \mathcal{N}( 0, 1^2)$ are used to generate heteroskedastic measurement errors. 
We also consider the motorcycle data set studied in \cite{Silverman:1985}, which shows obvious heteroskedastic noise and non-stationarity, with areas exhibiting different behaviours.
The data measures head acceleration in a simulated motorcycle accident for testing helmets; the measurements are practically constant initially, with a low noise level, which abruptly turns oscillating with high noise and gradually decaying to a possibly constant state while still having clearly higher measurement noise in comparison to the initial constant state.
Lastly, we also study the 3D Langley glide-back booster simulation and 4D Colorado precipitation data sets. 

The Langley glide-back booster is a rocket booster designed by NASA to allow the booster to glide down from the atmosphere instead of requiring crashing it into the ocean.
A more detailed description of the booster and the related data set is provided in  \cite{Gramacy:2008}.
As in \citet{Gramacy:2008}, we model the lift as the output variable and the input variables include the vehicle speed at atmospheric re-entry, angle of attack, and the sideslip angle which are referred to in the data set as Mach, $\alpha$, and $\beta$, respectively.
The data set consists of $3167$ data points of which we use the subset of $N = 1900$ data points.

The Colorado precipitation data set has been collected by the Colorado Climate Center and is available online.
The data set consists of monthly total precipitation amounts measured at various weather stations located in Colorado.
The monthly precipitation is modelled as the output variable with the input variables being the station longitude, latitude, and elevation, together with the measurement month.
We use data measured over October--December 2022 to showcase the distinct phenomenon of increased mean precipitation at higher elevations during winter months \citep{Mahoney:2015}.
Our data set corresponding to the period October--December 2022 consists of $392$ data points.

For all data sets, the inputs and outputs are normalized such that $X \in [0, 1]^D$, $\min Y = 0$, and $\text{Var}[Y] = 1$.
The prior distributions for the gating network parameters $\Psi$ and experts parameters $\Theta$ are summarized in Table \ref{table:priorParametersND}.
Specifically, we use normal and half-normal priors for the gating network means and variances, respectively, along with a uniform prior for the GP constant mean functions and half-normal priors for variance and length scale parameters of the GPs.
To encourage well-separated components along the input space, the prior means for the gating network parameters are constructed using a linearly spaced grid along each dimension. 

\begin{figure}[!t]
    \begin{minipage}{0.32\textwidth}
    ~
    \end{minipage}
    \hspace{\fill} 
    \begin{minipage}{0.32\textwidth}
    \subfloat[][Ground truth. ]{\includegraphics[width=\textwidth]{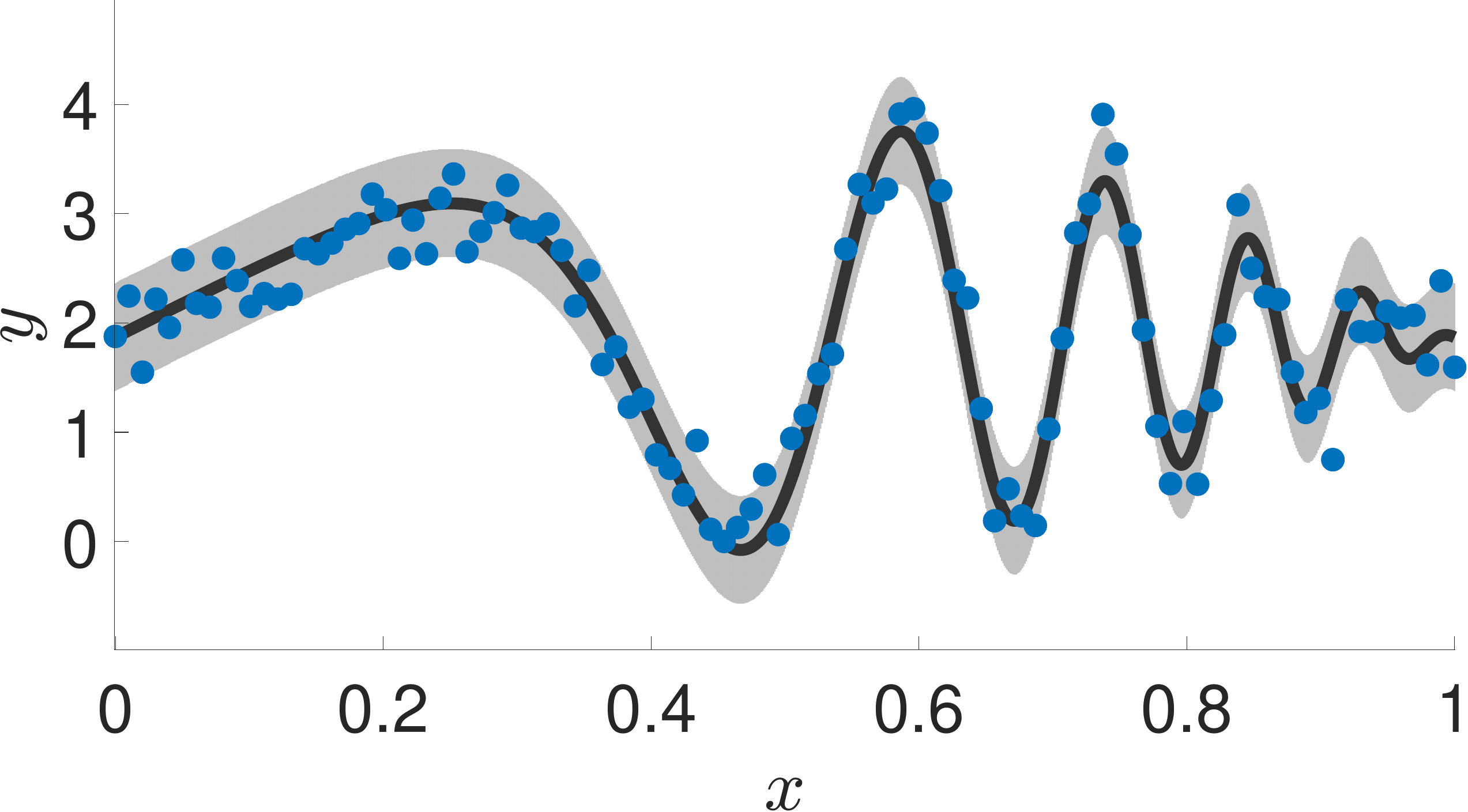}}
    \end{minipage}
    \hspace{\fill} 
    \begin{minipage}{0.32\textwidth}
    ~
    \end{minipage}
    \hspace{\fill} 

    \vspace*{0.5cm} 

    \begin{minipage}{0.32\textwidth}
    \subfloat[][SMC$^2$-MoE, $\alpha = 0.1$. ]{\includegraphics[width=\textwidth]{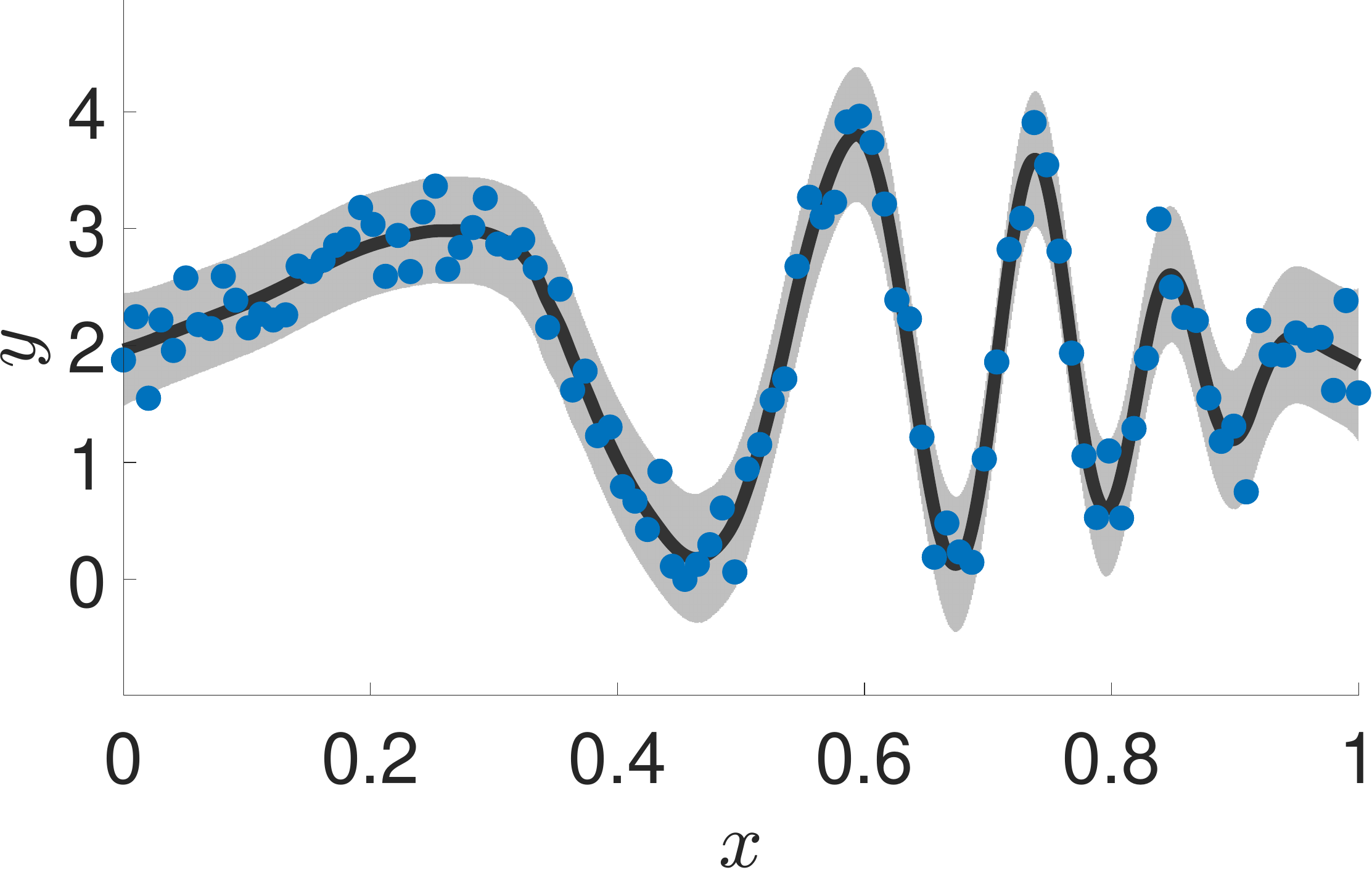}}
    \end{minipage}
    \hspace{\fill} 
    \begin{minipage}{0.32\textwidth}
    \subfloat[][SMC$^2$-MoE, $\alpha = 1$. ]{\includegraphics[width=\textwidth]{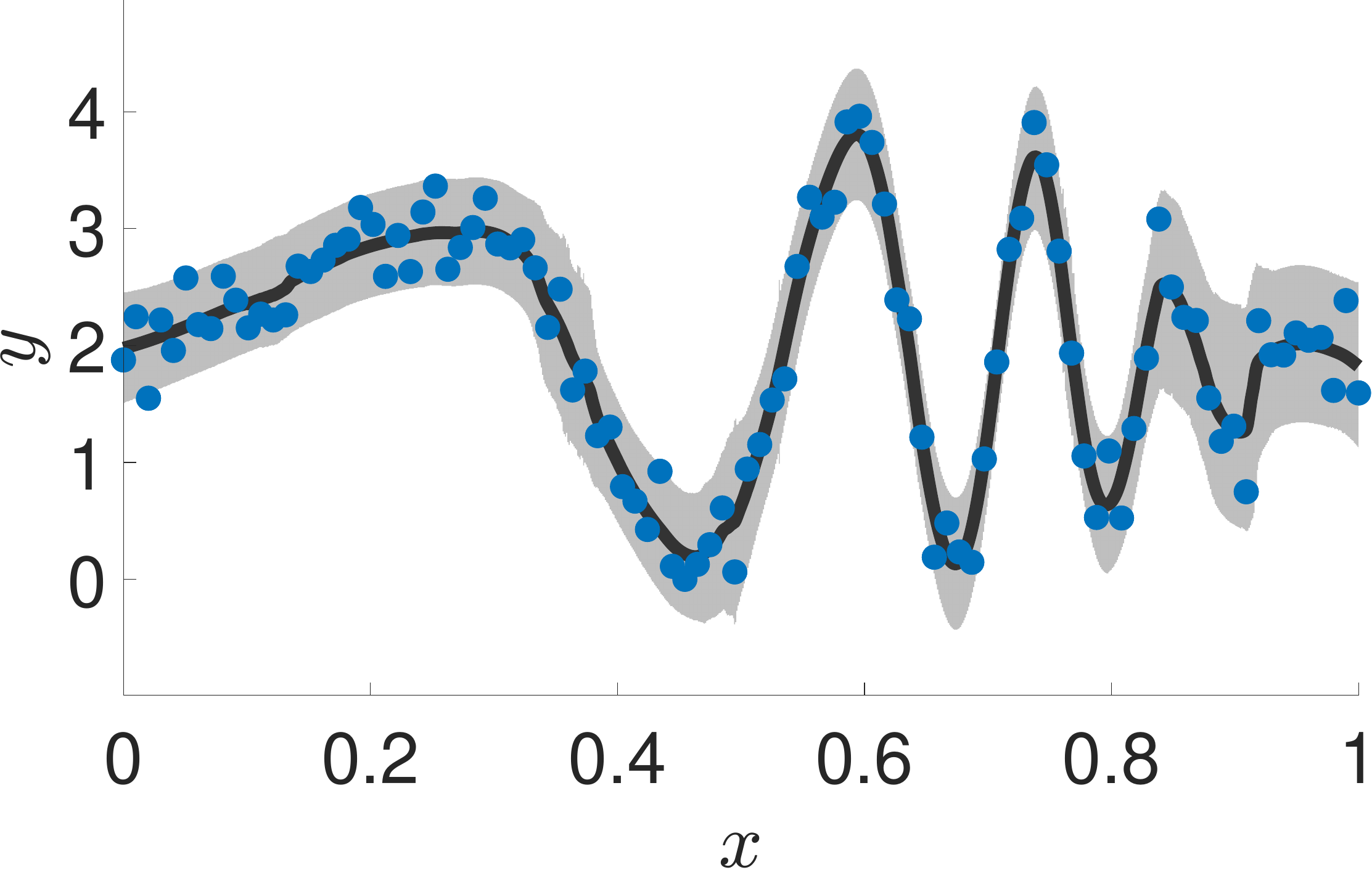}}
    \end{minipage}
    \hspace{\fill} 
    \begin{minipage}{0.32\textwidth}
    \subfloat[][SMC$^2$-MoE, $\alpha = K/2$. ]{\includegraphics[width=\textwidth]{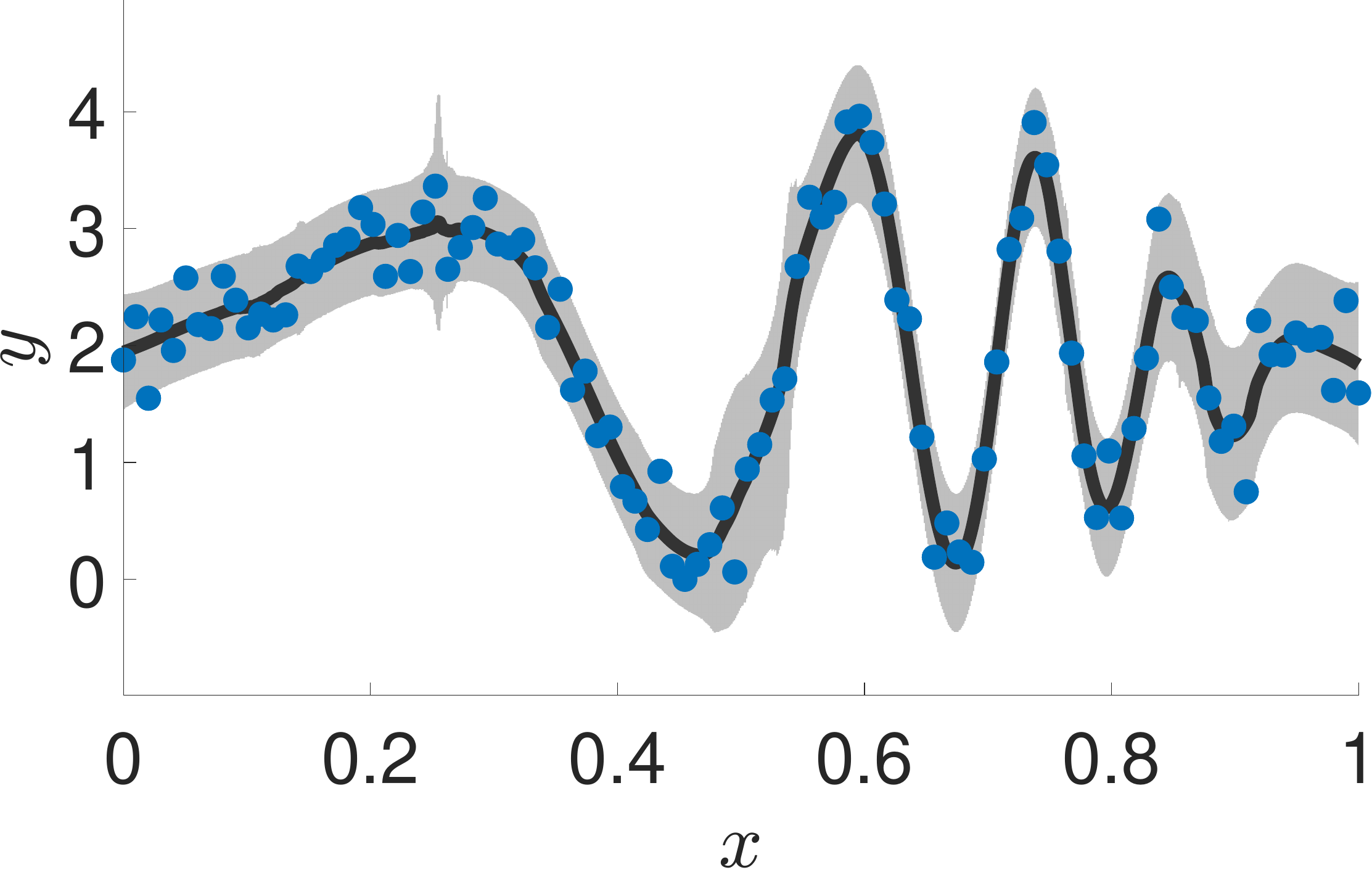}}
    \end{minipage}
    \hspace{\fill} 

    \vspace*{0.5cm} 

    \begin{minipage}{0.32\textwidth}
    \subfloat[][IS-MoE, $\alpha = 0.1$. ]{\includegraphics[width=\textwidth]{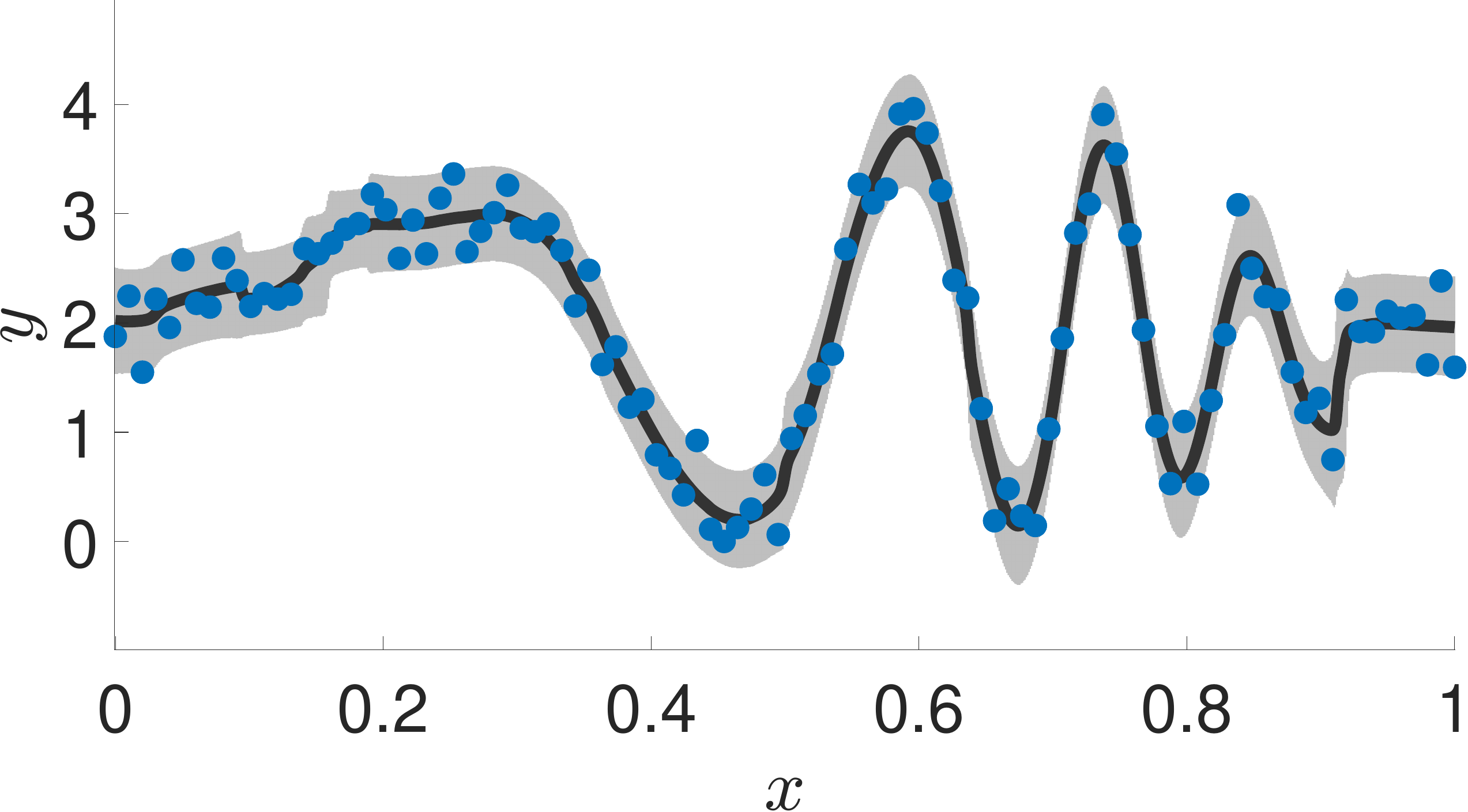}}
    \end{minipage}
    \hspace{\fill} 
    \begin{minipage}{0.32\textwidth}
    \subfloat[][IS-MoE, $\alpha = 1$. ]{\includegraphics[width=\textwidth]{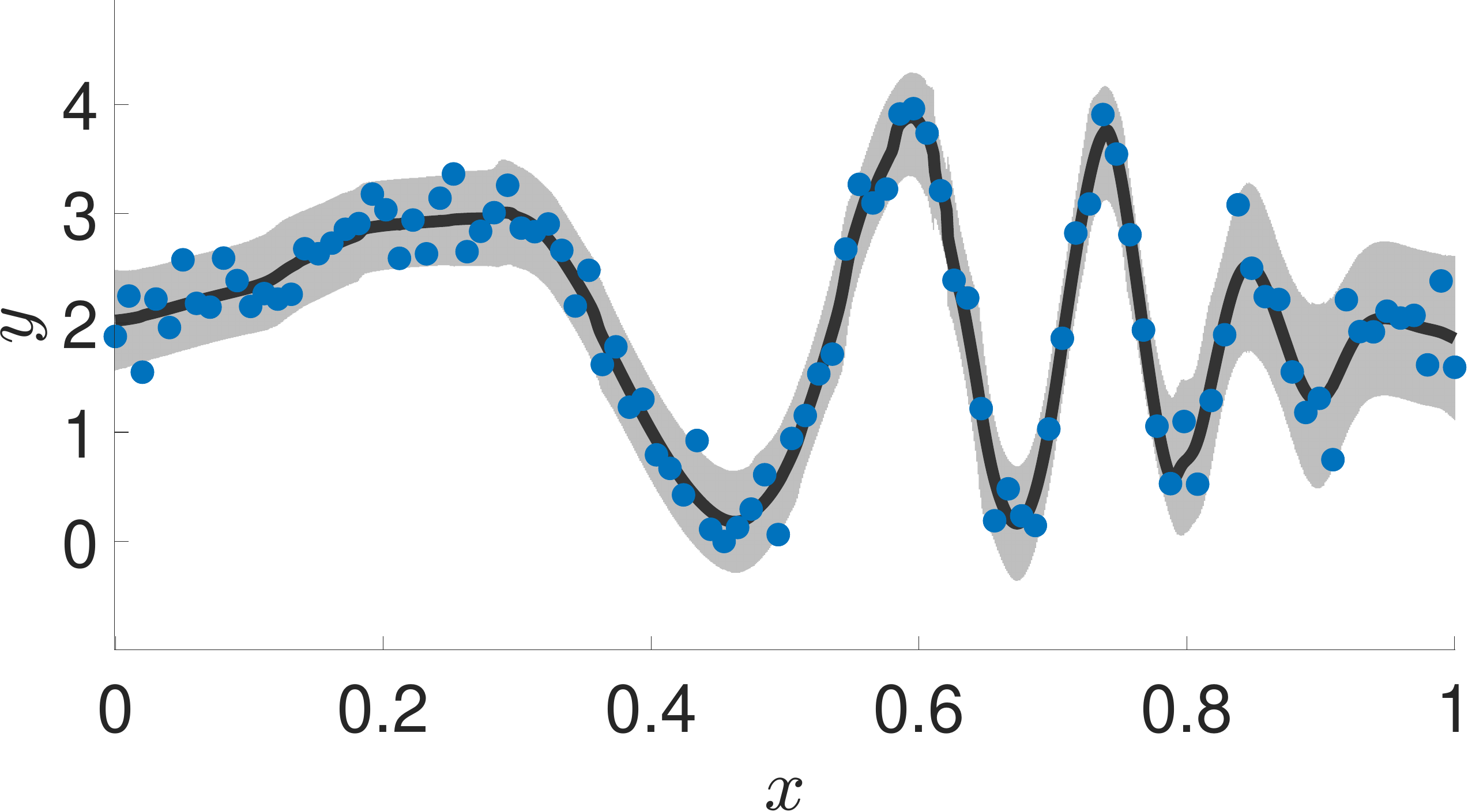}}
    \end{minipage}
    \hspace{\fill} 
    \begin{minipage}{0.32\textwidth}
    \subfloat[][IS-MoE, $\alpha = K/2$. ]{\includegraphics[width=\textwidth]{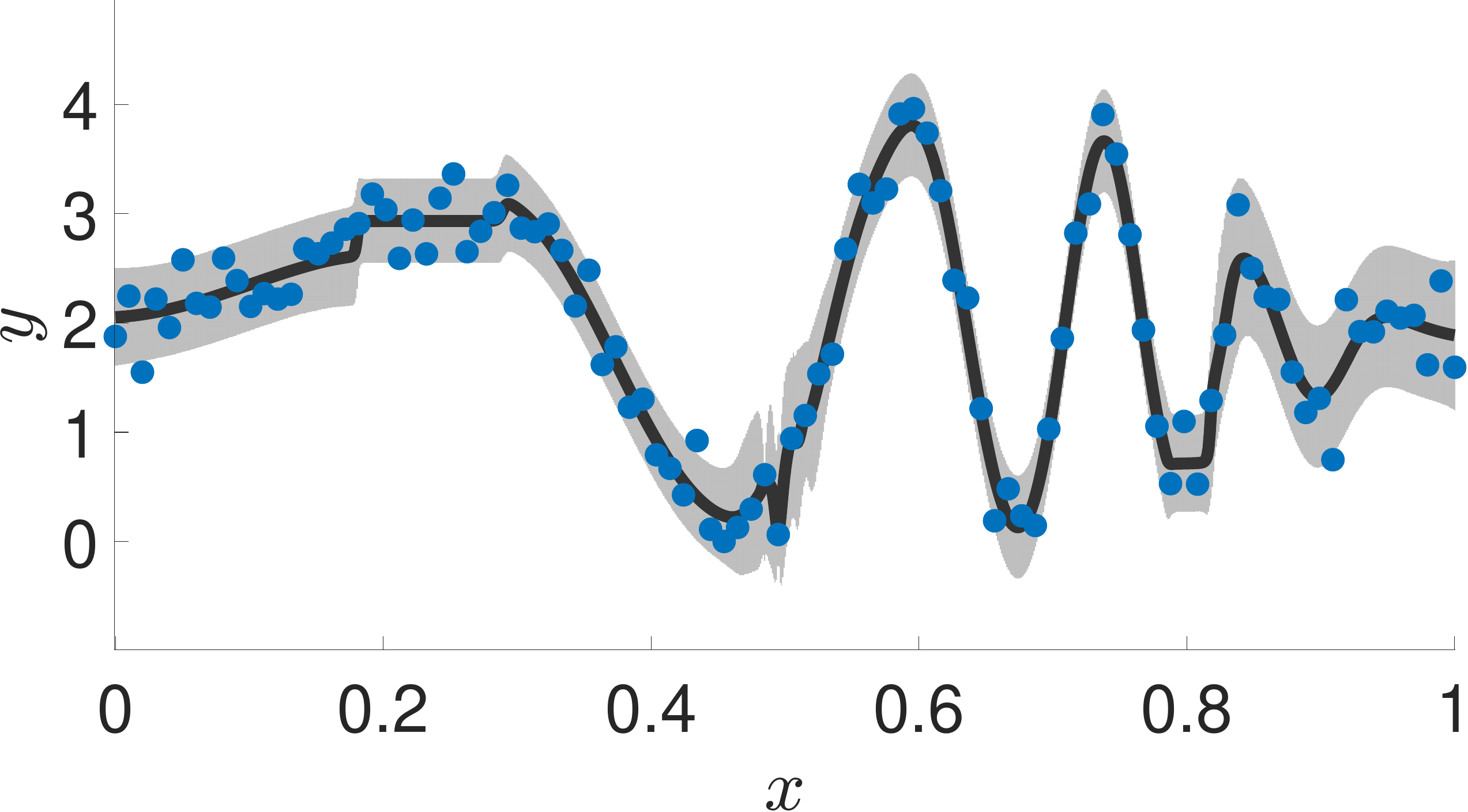}}
    \end{minipage}
    \hspace{\fill} 

    \vspace*{0.5cm} 

    \begin{minipage}{0.32\textwidth}
    \subfloat[][GP.]{\includegraphics[width=\textwidth]{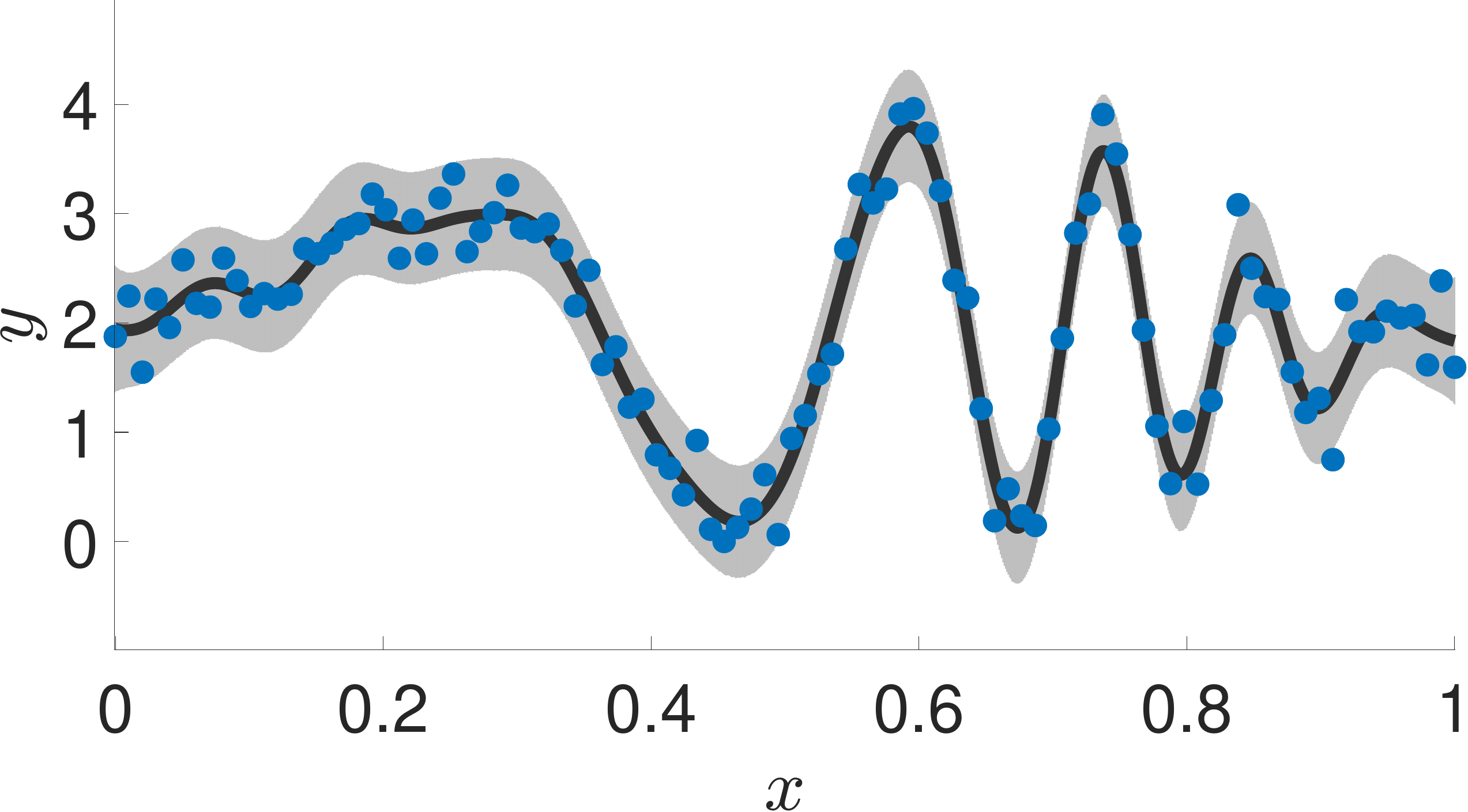}}
    \end{minipage}
    \hspace{\fill} 
    \begin{minipage}{0.32\textwidth}
    \subfloat[][RBCM.]{\includegraphics[width=\textwidth]{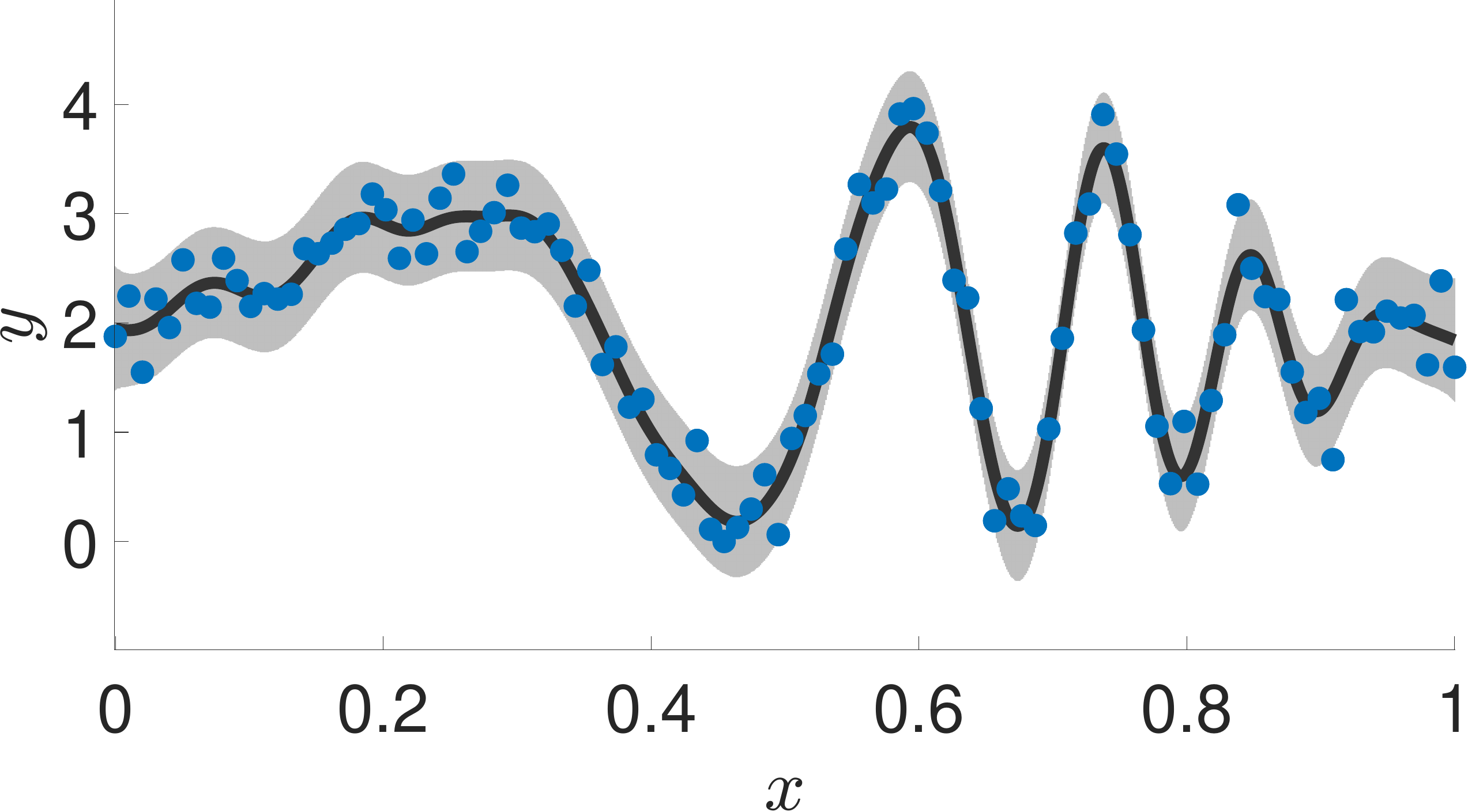}}
    \end{minipage}
    \hspace{\fill} 
    \begin{minipage}{0.32\textwidth}
    \subfloat[][treedGP.]{\includegraphics[width=\textwidth]{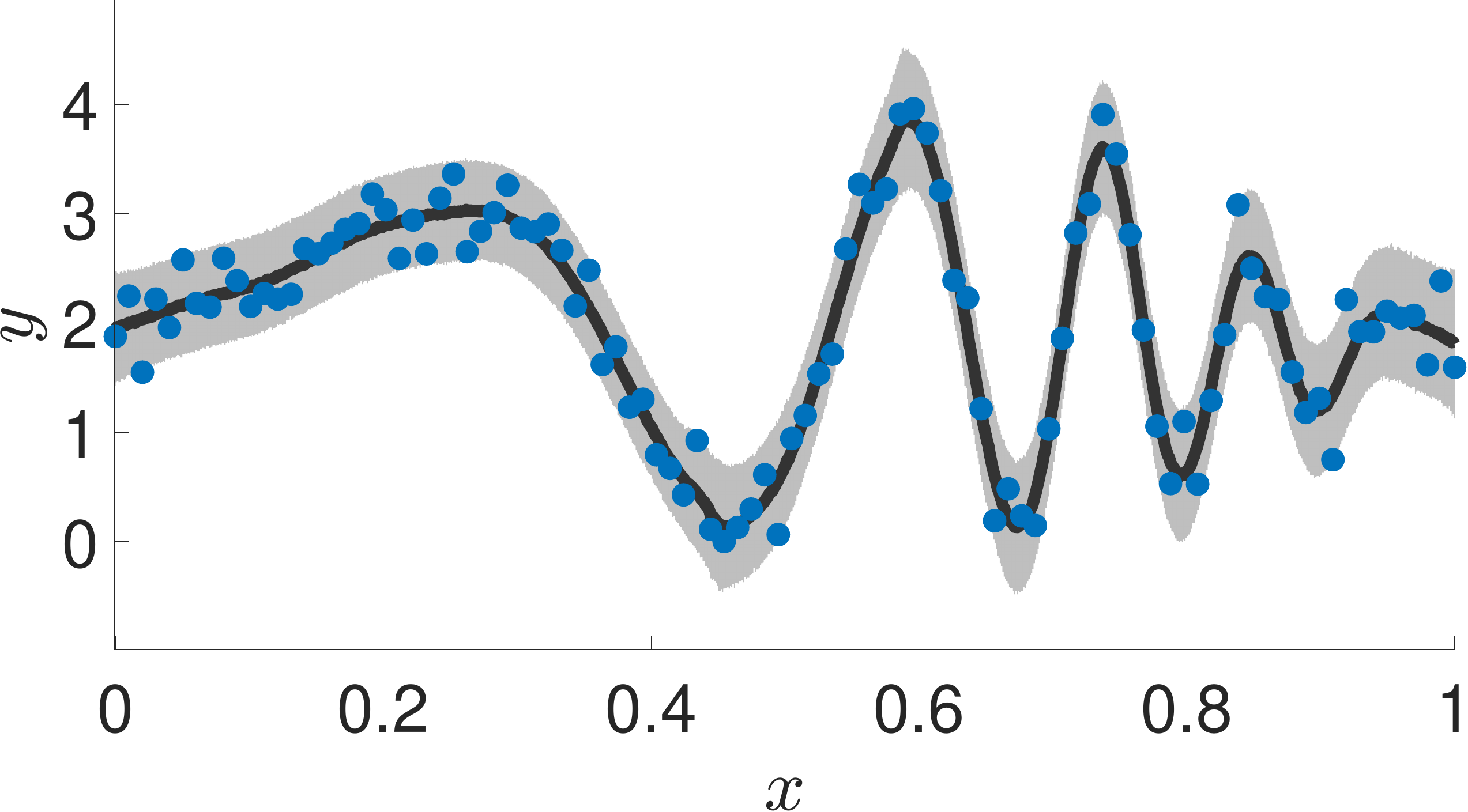}}
    \end{minipage}
    \hspace{\fill} 

    \vspace*{0.5cm} 
    
    \caption{ \textit{First synthetic data set}: non-stationary function, homoskedastic noise. In a), the generated data points are shown (in blue) with a 90\% highest density region of the underlying ground truth distribution in gray and the true function as a solid black line. In b), c), and d) the highest density region and posterior median for MoE with SMC$^2$ are shown for different values of the Dirichlet concentration parameter $\alpha = 0.1,1,K/2$. In e), f), and g) similar results are shown for IS-MoE with the same Dirichlet concentration parameters. In the bottom row, h), i), and j) show the predictive distributions for a single GP, RBCM, and treedGP, respectively.} \label{im:stationaryComparisonFigure}
\end{figure}

\begin{figure}[!t]
    \begin{minipage}{0.32\textwidth}
    ~
    \end{minipage}
    \hspace{\fill} 
    \begin{minipage}{0.32\textwidth}
    \subfloat[][Ground truth. ]{\includegraphics[width=\textwidth]{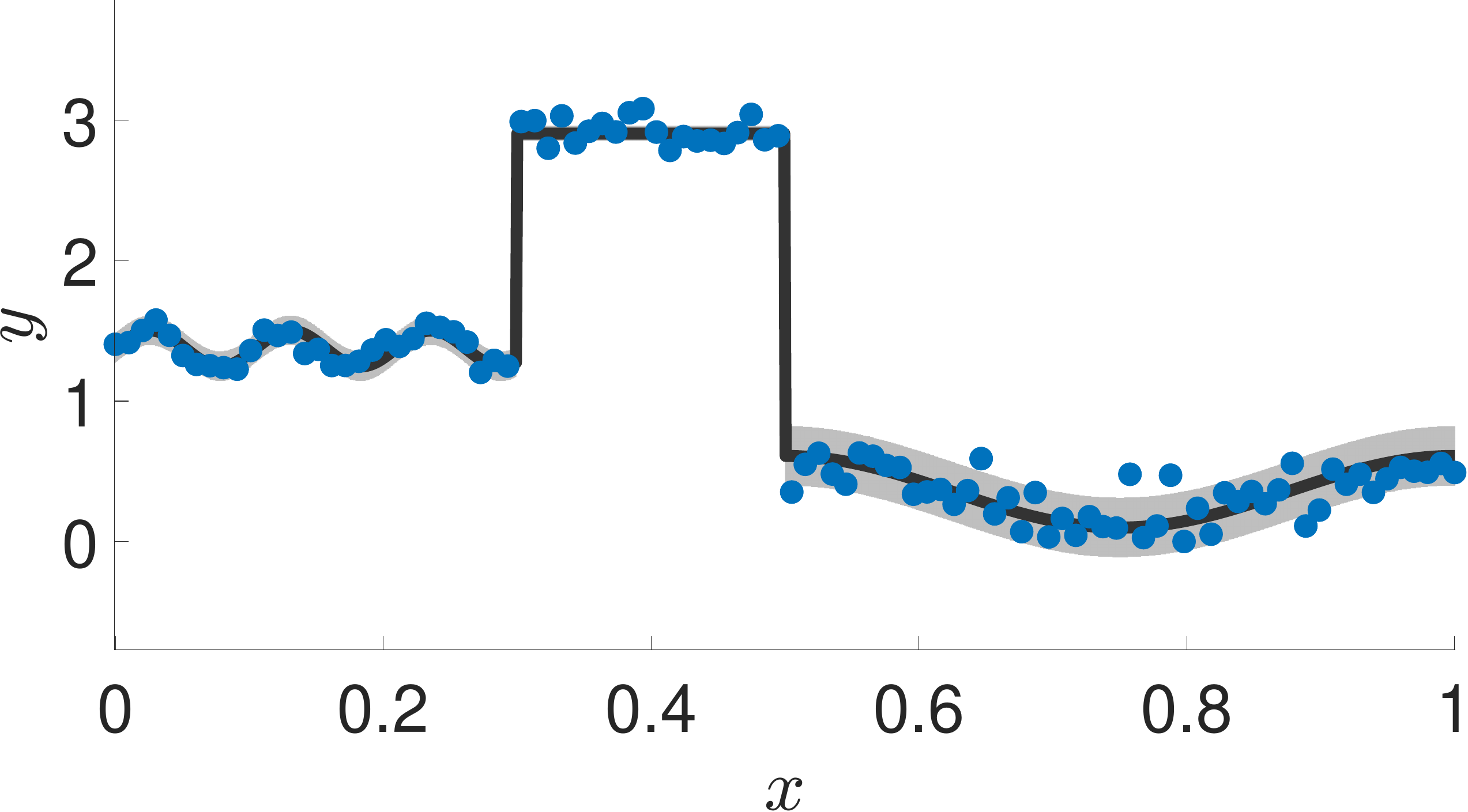}}
    \end{minipage}
    \hspace{\fill} 
    \begin{minipage}{0.32\textwidth}
    ~
    \end{minipage}
    \hspace{\fill} 

    \vspace*{0.5cm} 

    \begin{minipage}{0.32\textwidth}
    \subfloat[][SMC$^2$-MoE, $\alpha = 0.1$. ]{\includegraphics[width=\textwidth]{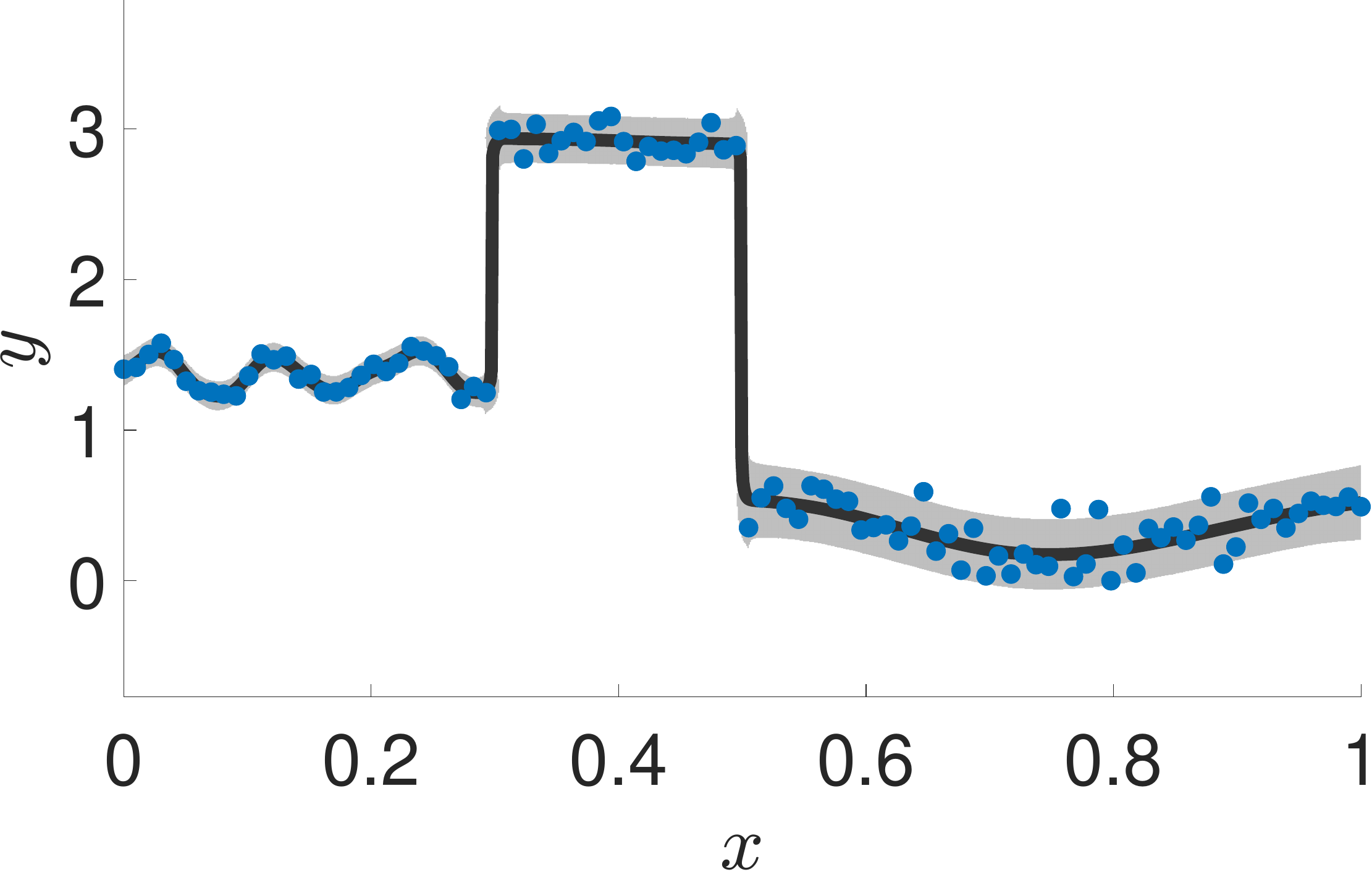}}
    \end{minipage}
    \hspace{\fill} 
    \begin{minipage}{0.32\textwidth}
    \subfloat[][SMC$^2$-MoE, $\alpha = 1$. ]{\includegraphics[width=\textwidth]{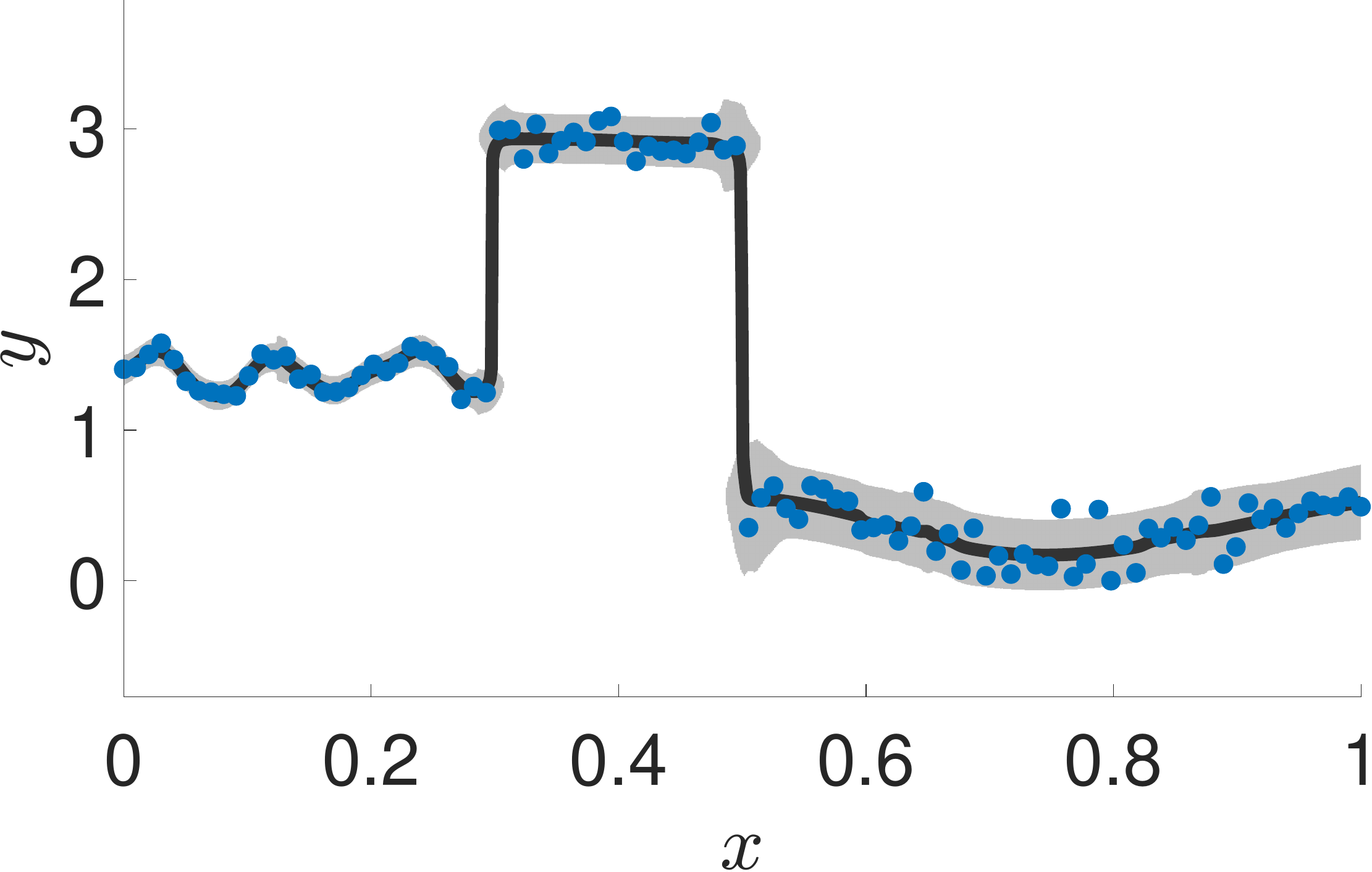}}
    \end{minipage}
    \hspace{\fill} 
    \begin{minipage}{0.32\textwidth}
    \subfloat[][SMC$^2$-MoE, $\alpha = K/2$. ]{\includegraphics[width=\textwidth]{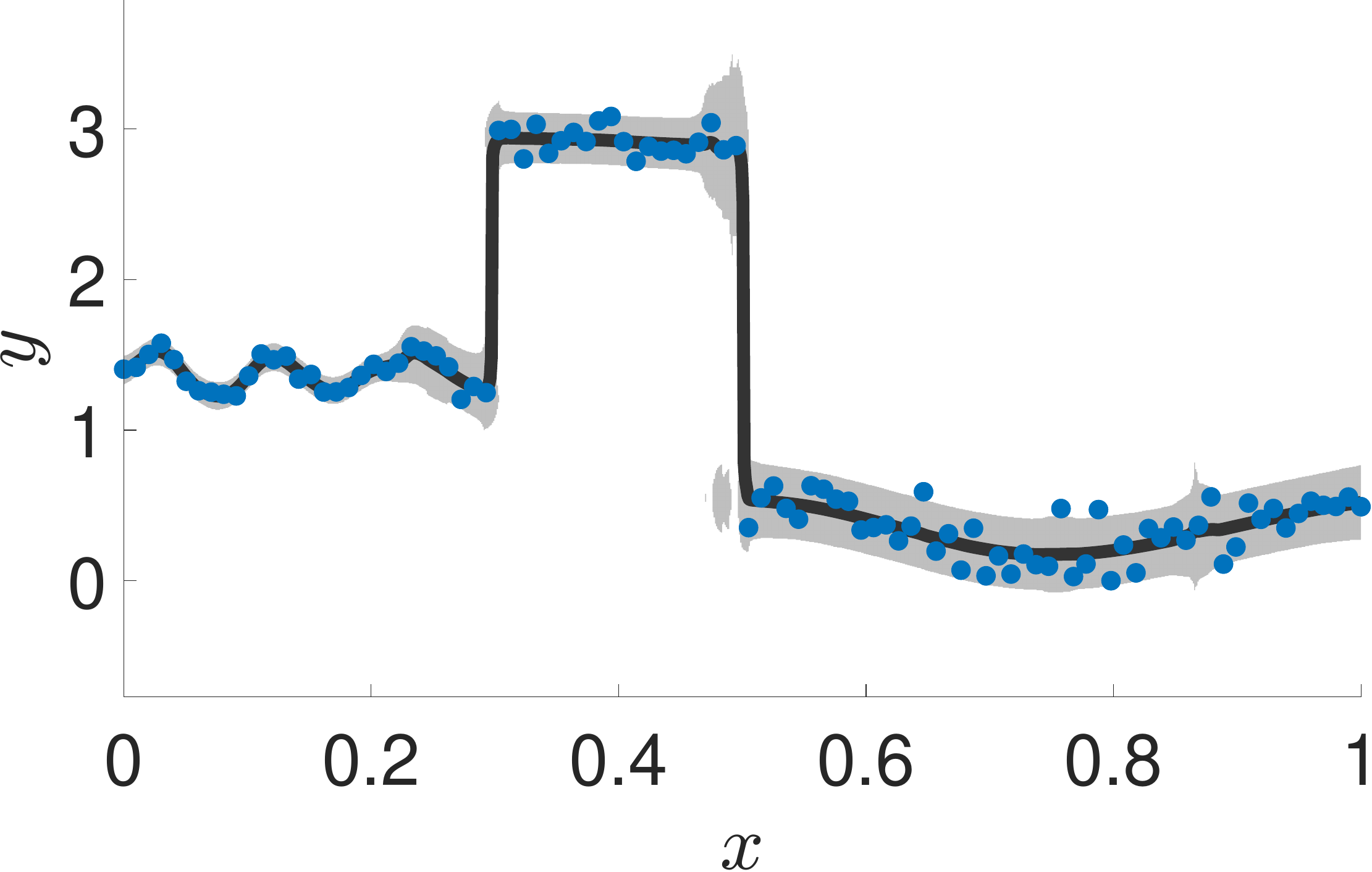}}
    \end{minipage}
    \hspace{\fill} 

    \vspace*{0.5cm} 

    \begin{minipage}{0.32\textwidth}
    \subfloat[][IS-MoE, $\alpha = 0.1$. ]{\includegraphics[width=\textwidth]{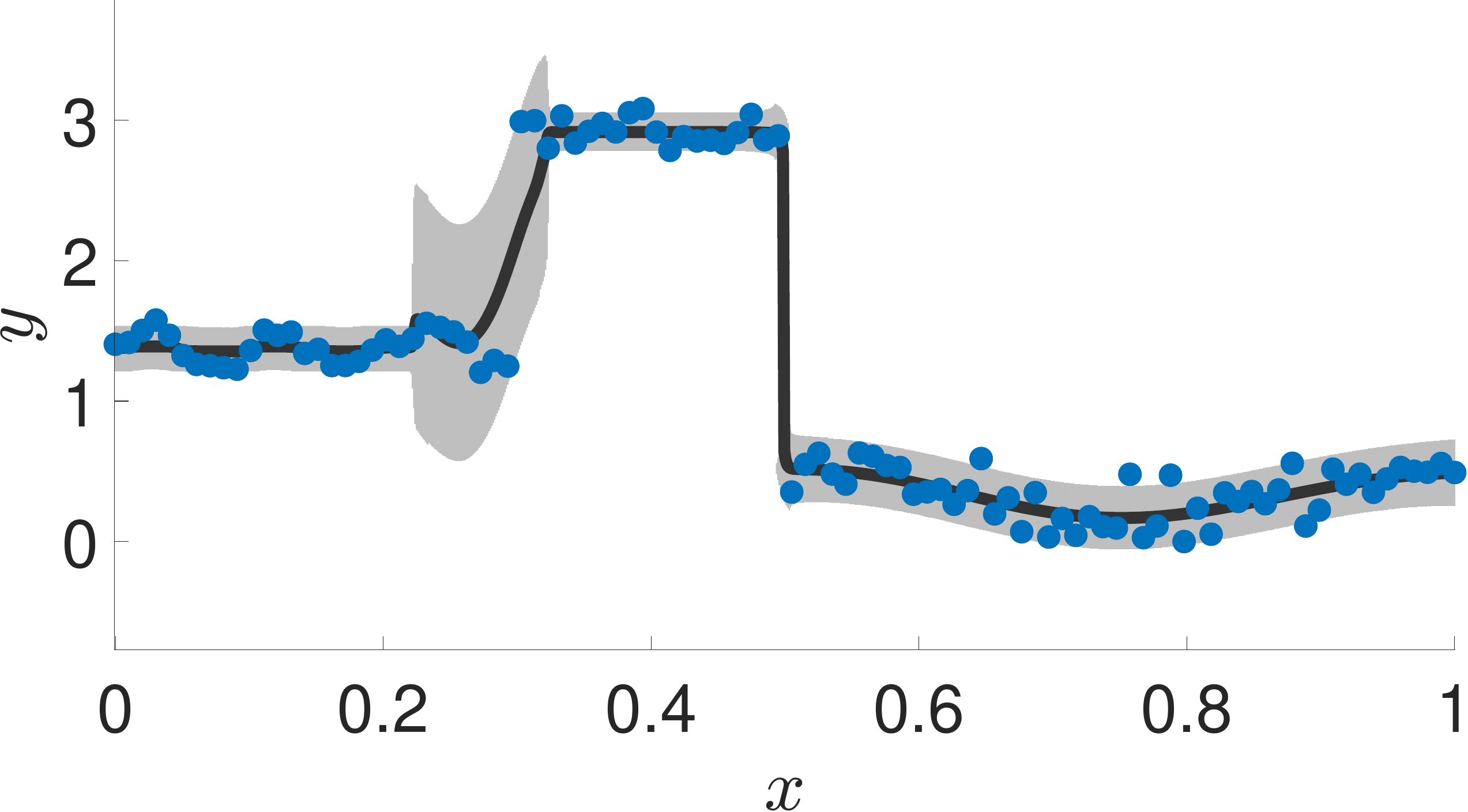}}
    \end{minipage}
    \hspace{\fill} 
    \begin{minipage}{0.32\textwidth}
    \subfloat[][IS-MoE, $\alpha = 1$. ]{\includegraphics[width=\textwidth]{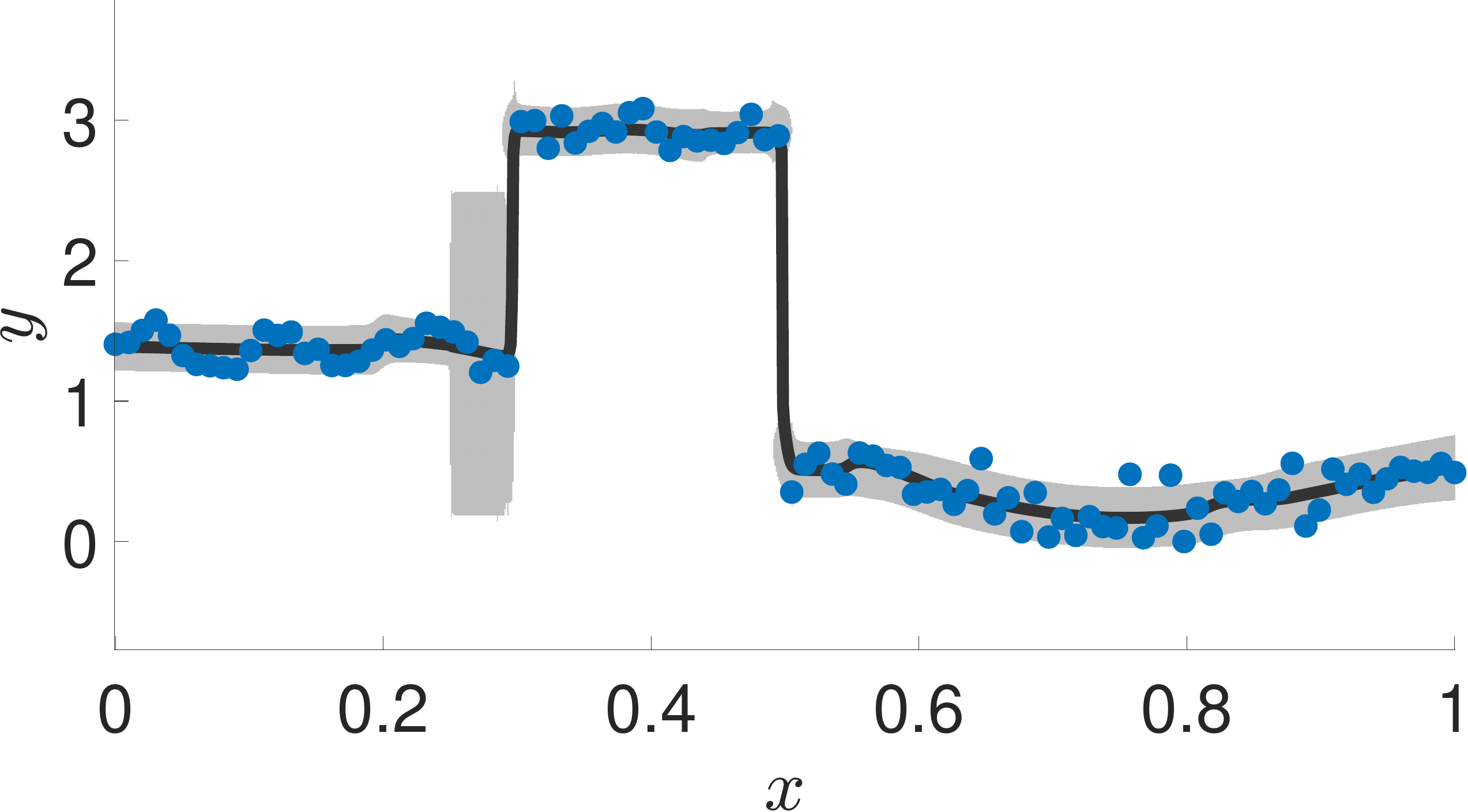}}
    \end{minipage}
    \hspace{\fill} 
    \begin{minipage}{0.32\textwidth}
    \subfloat[][IS-MoE, $\alpha = K/2$. ]{\includegraphics[width=\textwidth]{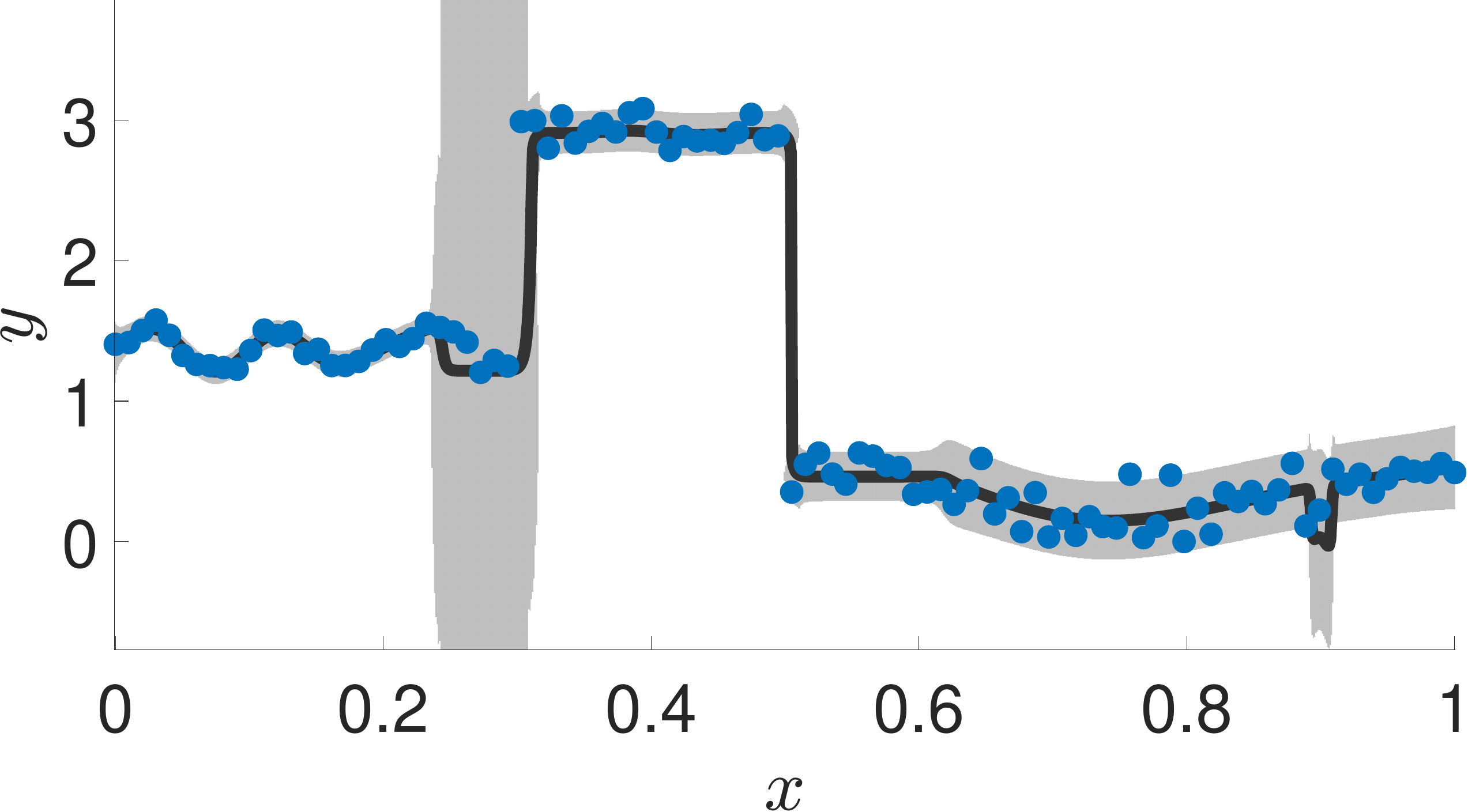}}
    \end{minipage}
    \hspace{\fill} 

    \vspace*{0.5cm} 

    \begin{minipage}{0.32\textwidth}
    \subfloat[][GP.]{\includegraphics[width=\textwidth]{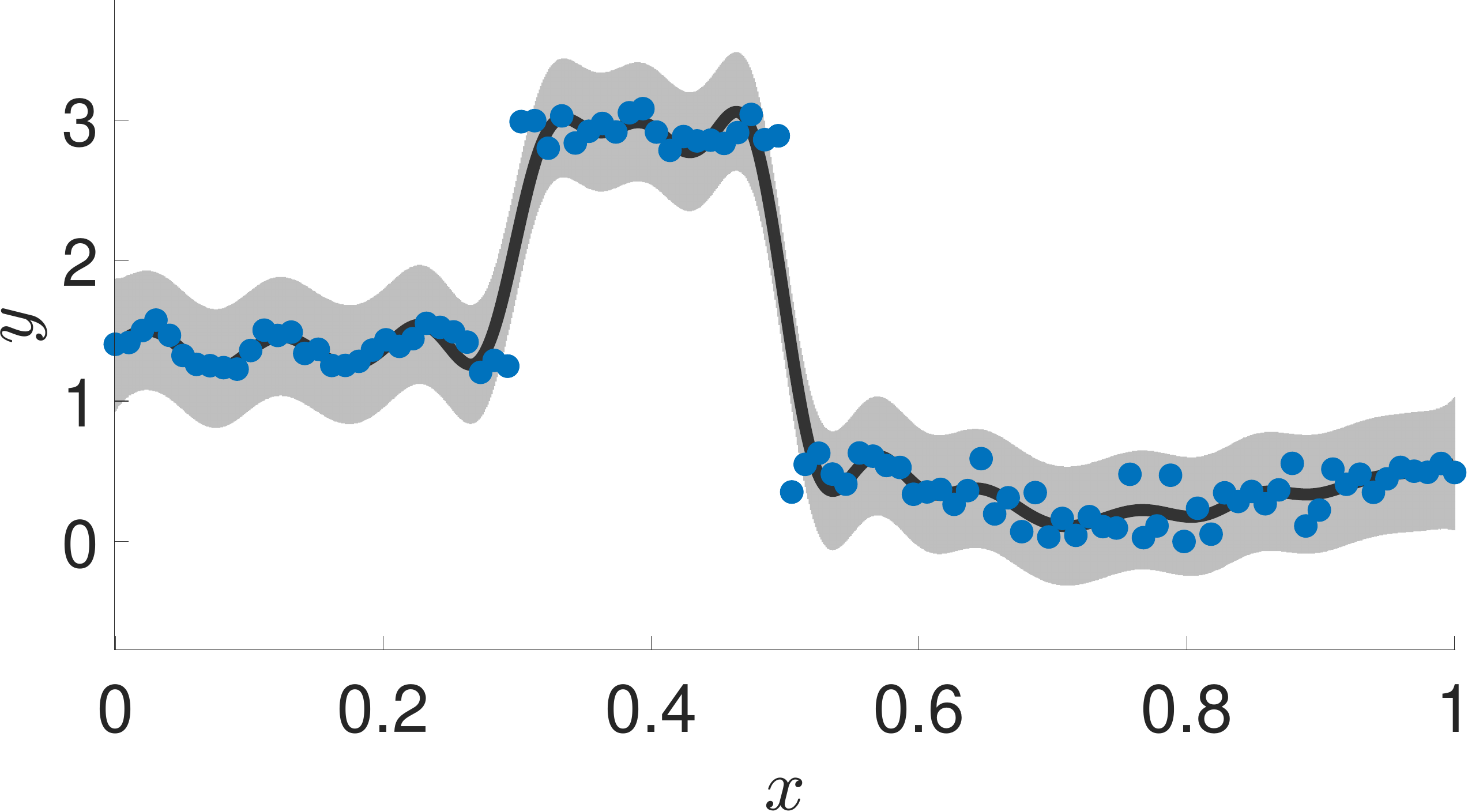}}
    \end{minipage}
    \hspace{\fill} 
    \begin{minipage}{0.32\textwidth}
    \subfloat[][RBCM.]{\includegraphics[width=\textwidth]{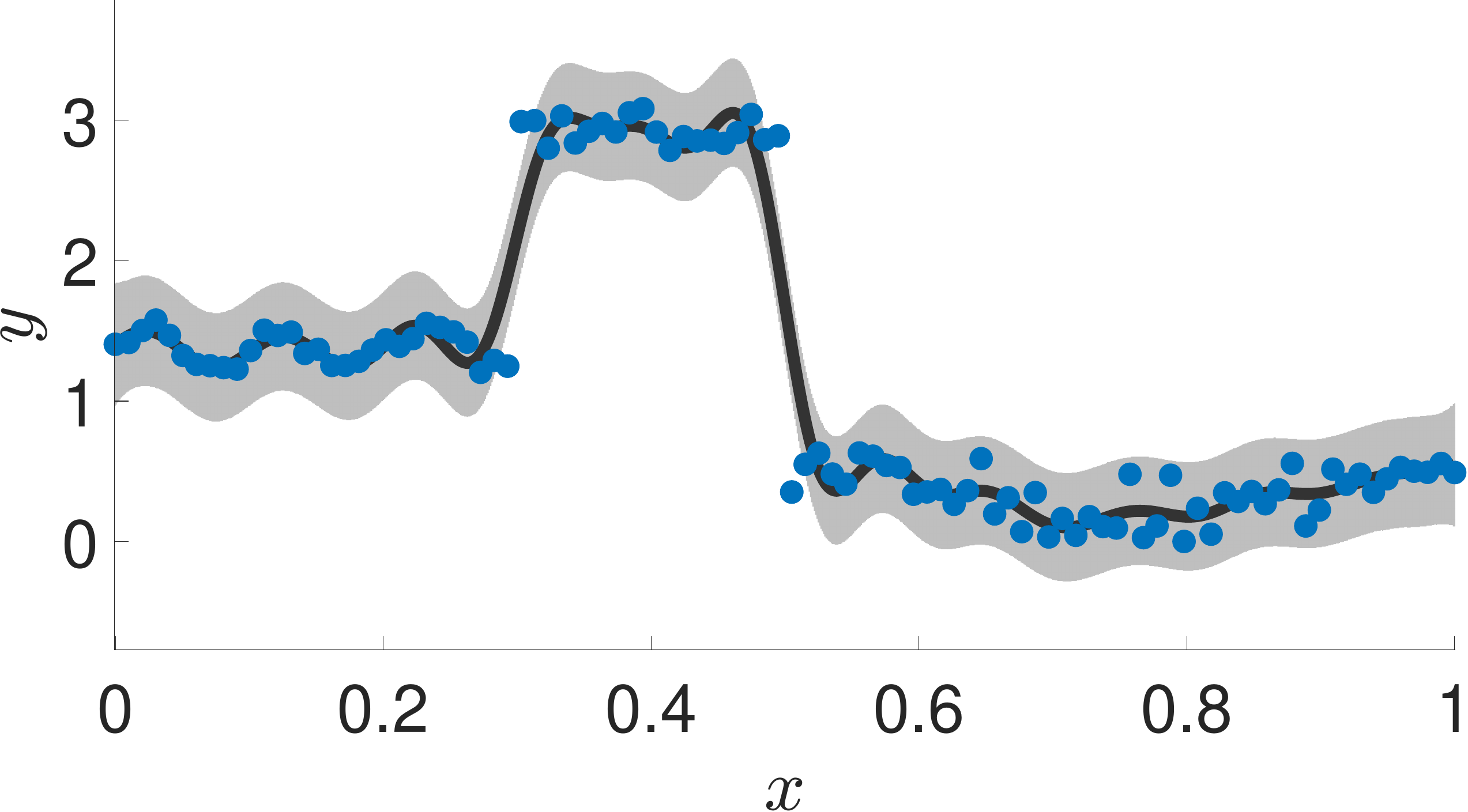}}
    \end{minipage}
    \hspace{\fill} 
    \begin{minipage}{0.32\textwidth}
    \subfloat[][treedGP.]{\includegraphics[width=\textwidth]{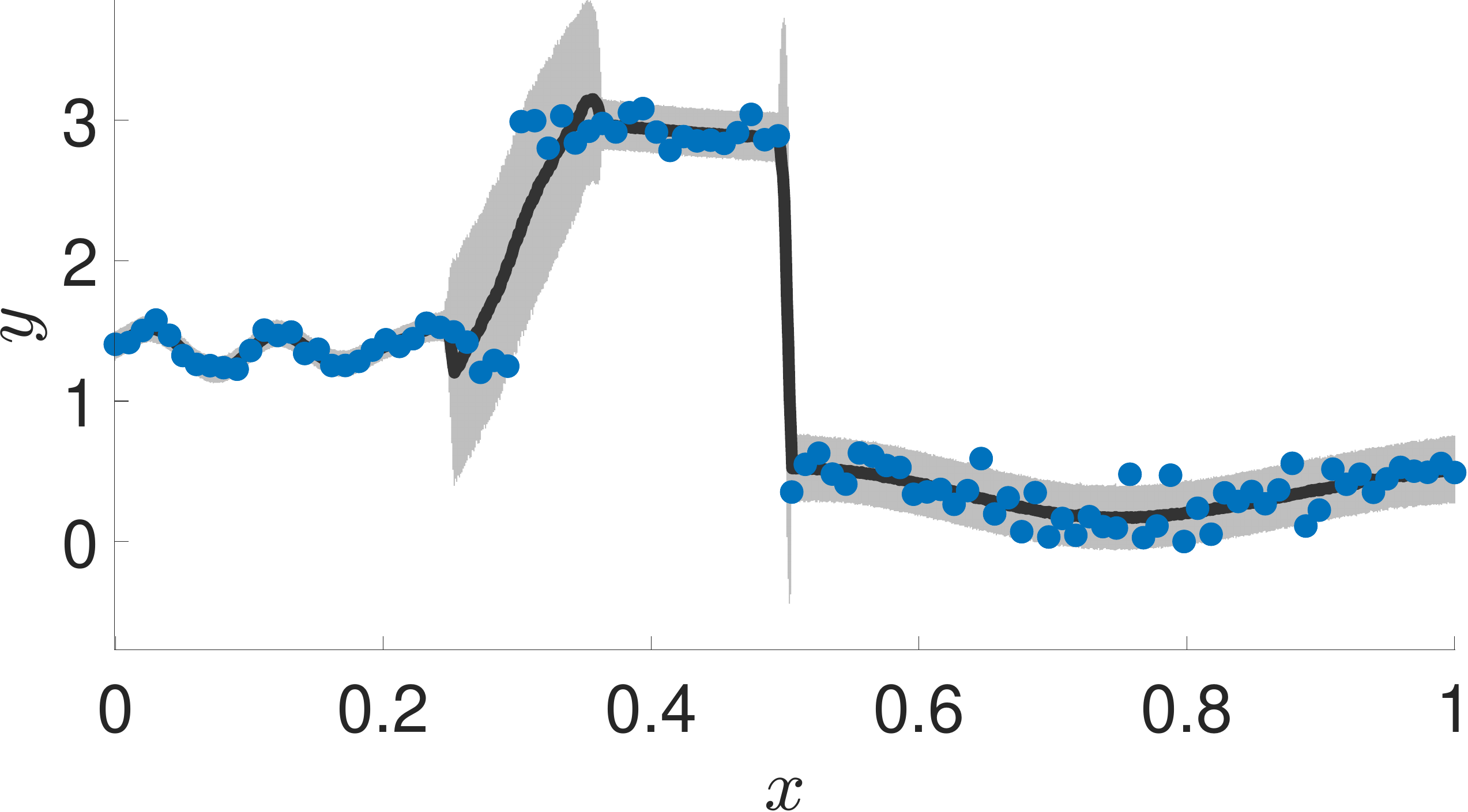}}
    \end{minipage}
    \hspace{\fill} 

    \vspace*{0.5cm} 
    
    \caption{ \textit{Second synthetic data set}: discontinuous function, heteroskedastic noise. In a) the generated data points are shown (in blue) with a highest density region of the underlying ground truth distribution in gray and the true function as a solid black line. In b), c), and d) the highest density region and posterior median for MoE with SMC$^2$ are shown for different values of the Dirichlet concentration parameter $\alpha = 0.1,1,K/2$. In e), f), and g) similar results are shown for IS-MoE with same Dirichlet concentration parameters. In the bottom row, h), i), and j) show the predictive distributions for a single GP, RBCM, and treedGP, respectively.} \label{im:nonstationaryComparisonFigure}
\end{figure}

\begin{table}[!t]
\centering
\begin{tabular}{l|r|r|r|r}
\toprule
  & $L^1$ & $L^2$ & $L^1$ median & LL \\
\midrule
\text{SMC$^2$-MoE, $\alpha = 0.1$}&375.731&0.860&135.978 & -296.381 \\
\text{SMC$^2$-MoE, $\alpha = 1$}&402.887&0.881&139.738 & -298.903 \\
\text{SMC$^2$-MoE, $\alpha = K/2$}&403.566&0.893&138.866 & -299.365 \\
\text{IS-MoE, $\alpha = 0.1$}&588.068&1.042&154.580 & -293.514 \\
\text{IS-MoE, $\alpha = 1$}&599.560&1.073&159.100 & -299.317\\
\text{IS-MoE, $\alpha = K/2$}&602.536&1.075&159.797 & -297.154 \\
\text{GP}&373.676&0.912&142.085 & -298.689 \\
\text{RBCM}&424.606&1.063&148.719 & -297.406 \\
\text{TreedGP}&363.797&0.855&134.044 & -297.720 \\
  \bottomrule
\end{tabular}
\caption{\textit{First synthetic data set}: non-stationary function, homoskedastic noise. Column $L^1$ shows the $L^1$ vector norm of the difference between the estimated distributions and the ground truth distribution used to generate the data set for a given method. Column $L^2$ shows the $L^2$ vector norm of the difference between the aforementioned distributions and column $L^1$ median shows the $L^1$ vector norm of the distance between the posterior predictive median and the ground truth median. Column LL displays predictive log-likelihood results.}
\label{tb:stationaryDifferences}
\end{table}
\begin{table}[!t]
\centering
\begin{tabular}{l|r|r|r|r}
\toprule
  & $L^1$ & $L^2$ & $L^1$ median & LL \\
\midrule
\text{SMC$^2$-MoE, $\alpha = 0.1$}&509.876&2.368&46.418 & -224.955 \\
\text{SMC$^2$-MoE, $\alpha = 1$}&530.417&2.406&46.407 & -222.160 \\
\text{SMC$^2$-MoE, $\alpha = K/2$}&558.190&2.499&47.203 & -225.534 \\
\text{IS-MoE, $\alpha = 0.1$}&993.265&3.301&84.464 & -258.943 \\
\text{IS-MoE, $\alpha = 1$}&934.844&3.167&85.765 & -248.291 \\
\text{IS-MoE, $\alpha = K/2$}&925.779&3.133&85.209 & -248.900 \\
\text{GP}&1078.485&3.496&95.274 & -316.766 \\
\text{RBCM}&1217.425&3.770&187.151 & -316.085 \\
\text{TreedGP}&655.436&2.857&87.843 & -246.613\\
  \bottomrule
\end{tabular}
\caption{\textit{Second synthetic data set}: discontinuous function, heteroskedastic noise. Column $L^1$ shows the $L^1$ vector norm of the difference between the estimated distributions and the ground truth distribution used to generate the data set for a given method. Column $L^2$ shows the $L^2$ vector norm of the difference between the aforementioned distributions and column $L^1$ median shows the $L^1$ vector norm of the distance between the posterior predictive median and the ground truth median. Column LL displays predictive log-likelihood results.}
\label{tb:nonstationaryDifferences}
\end{table}

\begin{figure}[!t]
    \begin{minipage}{0.40\textwidth}
    {\includegraphics[height = 3.65cm]
    {pics/vector/stationary/stationary_density_a01.pdf}
    }
    \end{minipage}
    \hspace{\fill} 
    \begin{minipage}{0.24\textwidth}
    \includegraphics[height = 3.75cm]{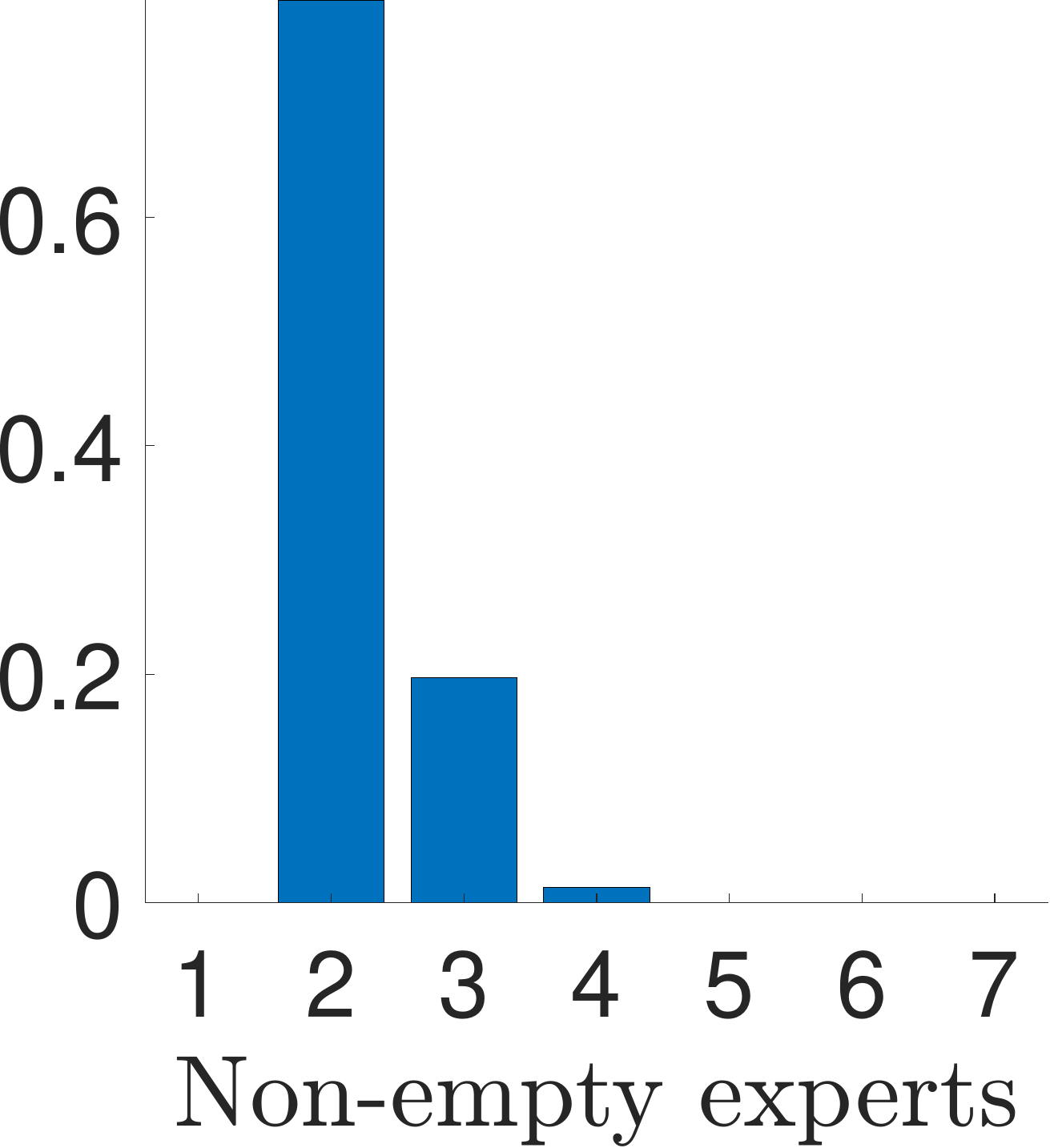}
    \end{minipage}
    \hspace{\fill} 
    \begin{minipage}{0.24\textwidth}
    {\includegraphics[height = 2.8cm]{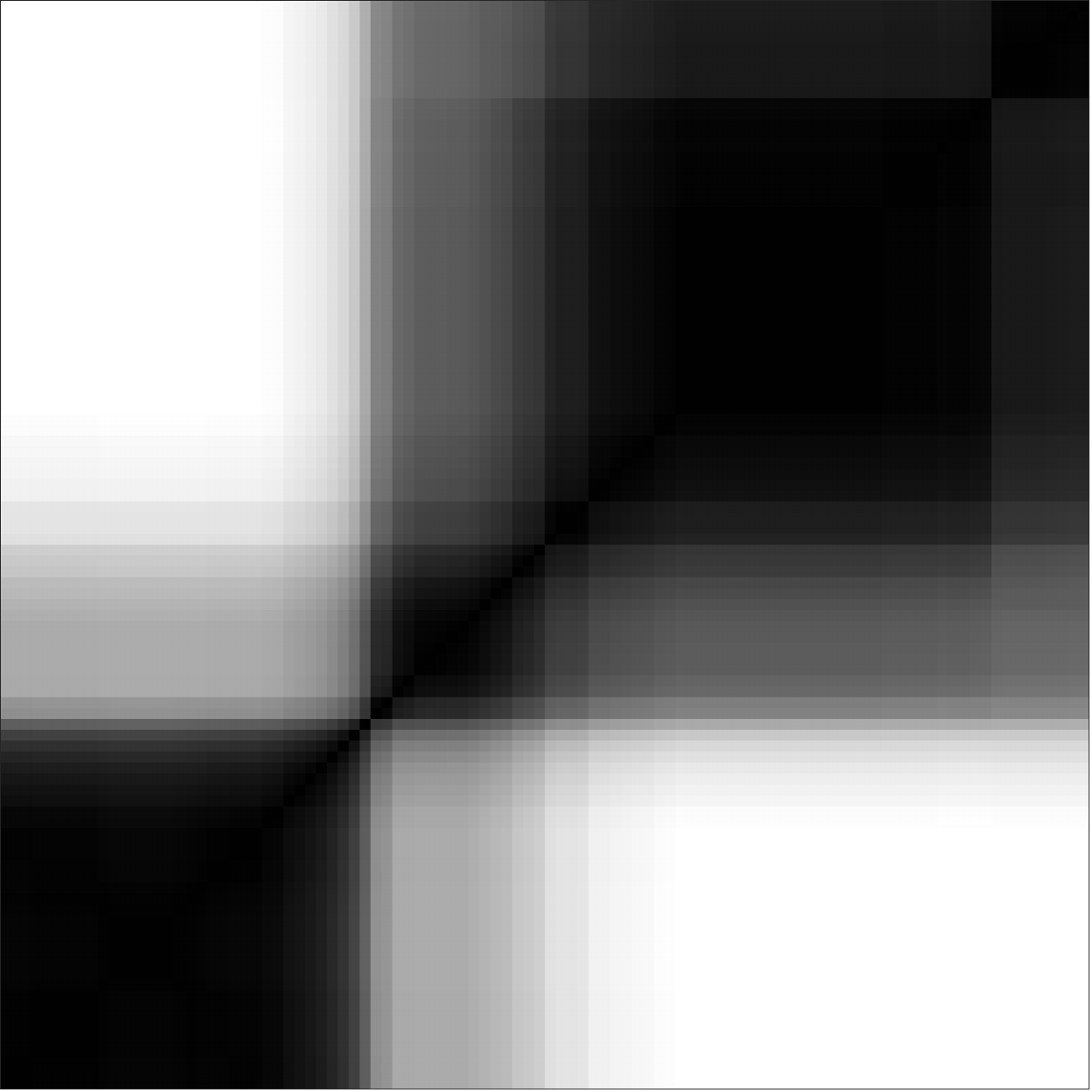}}
    \vspace{0.575cm}
    \end{minipage}

    \vspace*{1cm} 
    
    \begin{minipage}{0.40\textwidth}
    {\includegraphics[height = 3.65cm]{pics/vector/stationary/stationary_density_a1.pdf}}
    \end{minipage}
    \hspace{\fill} 
    \begin{minipage}{0.24\textwidth}
    \includegraphics[height = 3.75cm]{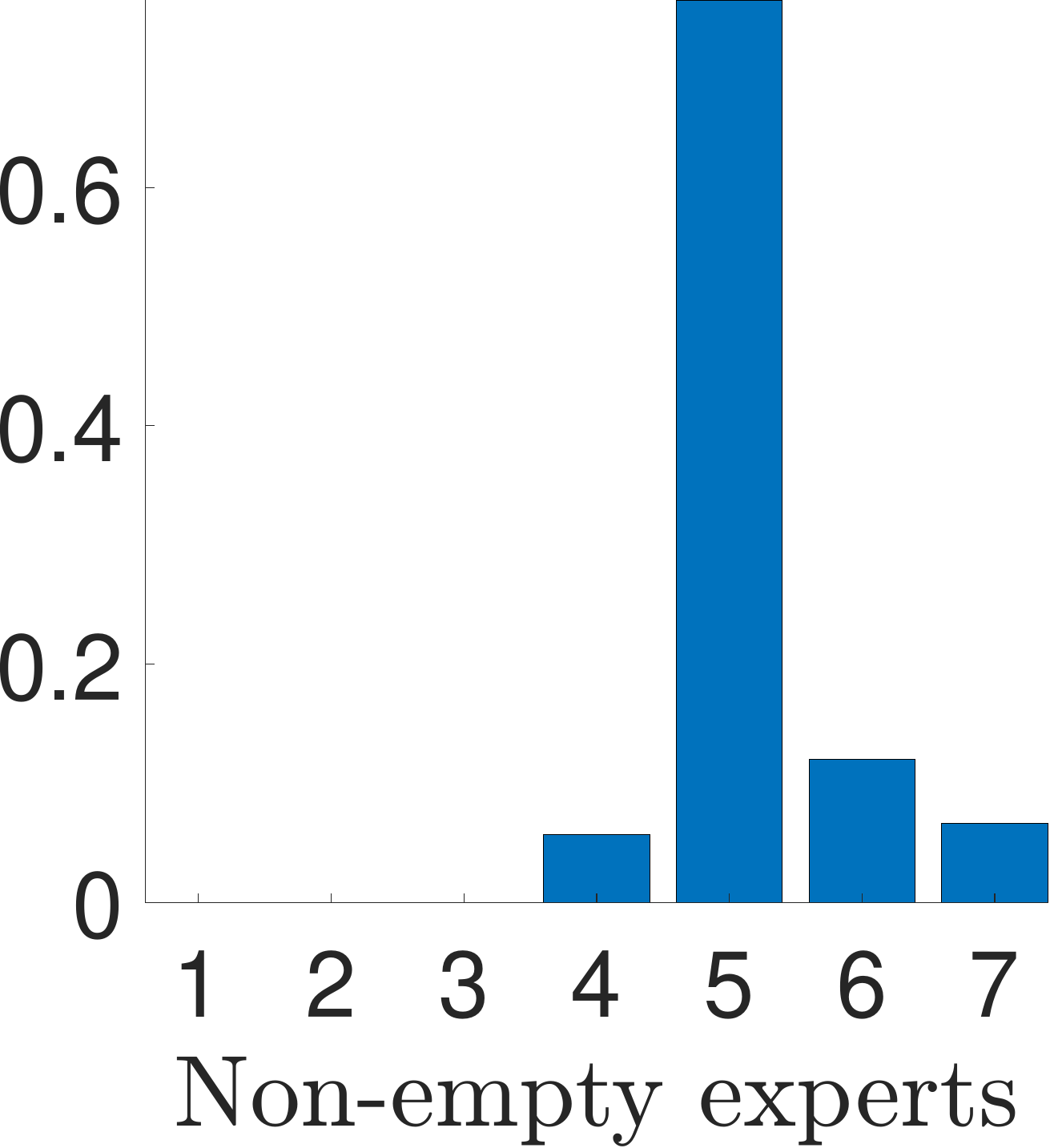}
    \end{minipage}
    \hspace{\fill} 
    \begin{minipage}{0.24\textwidth}
    {\includegraphics[height = 2.8cm]{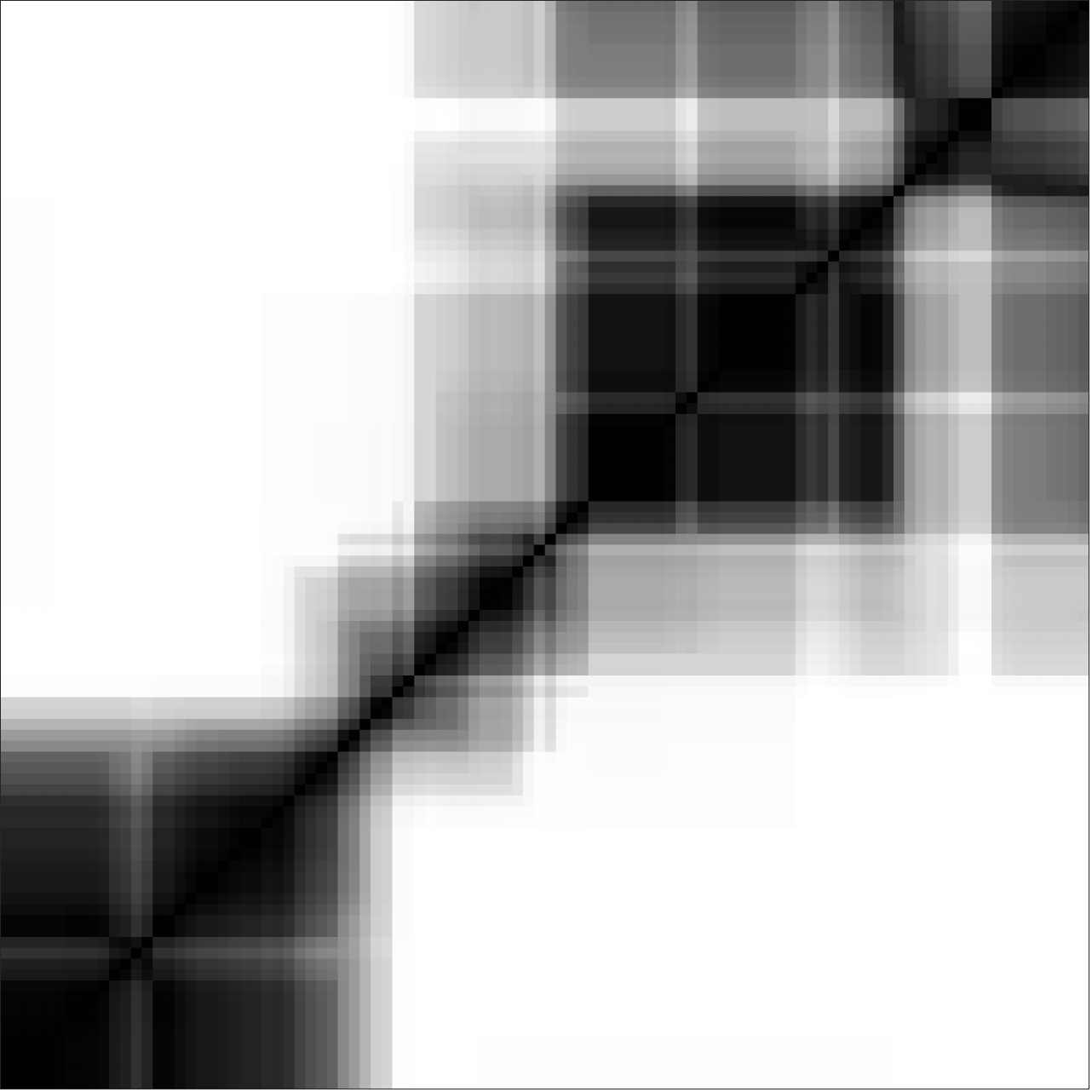}}
    \vspace{0.575cm}
    \end{minipage}

    \vspace*{1cm} 

    \begin{minipage}{0.40\textwidth}
    {\includegraphics[height = 3.65cm]{pics/vector/stationary/stationary_density_aJeffreys.pdf}}
    \end{minipage}
    \hspace{\fill} 
    \begin{minipage}{0.24\textwidth}
    \includegraphics[height = 3.75cm]{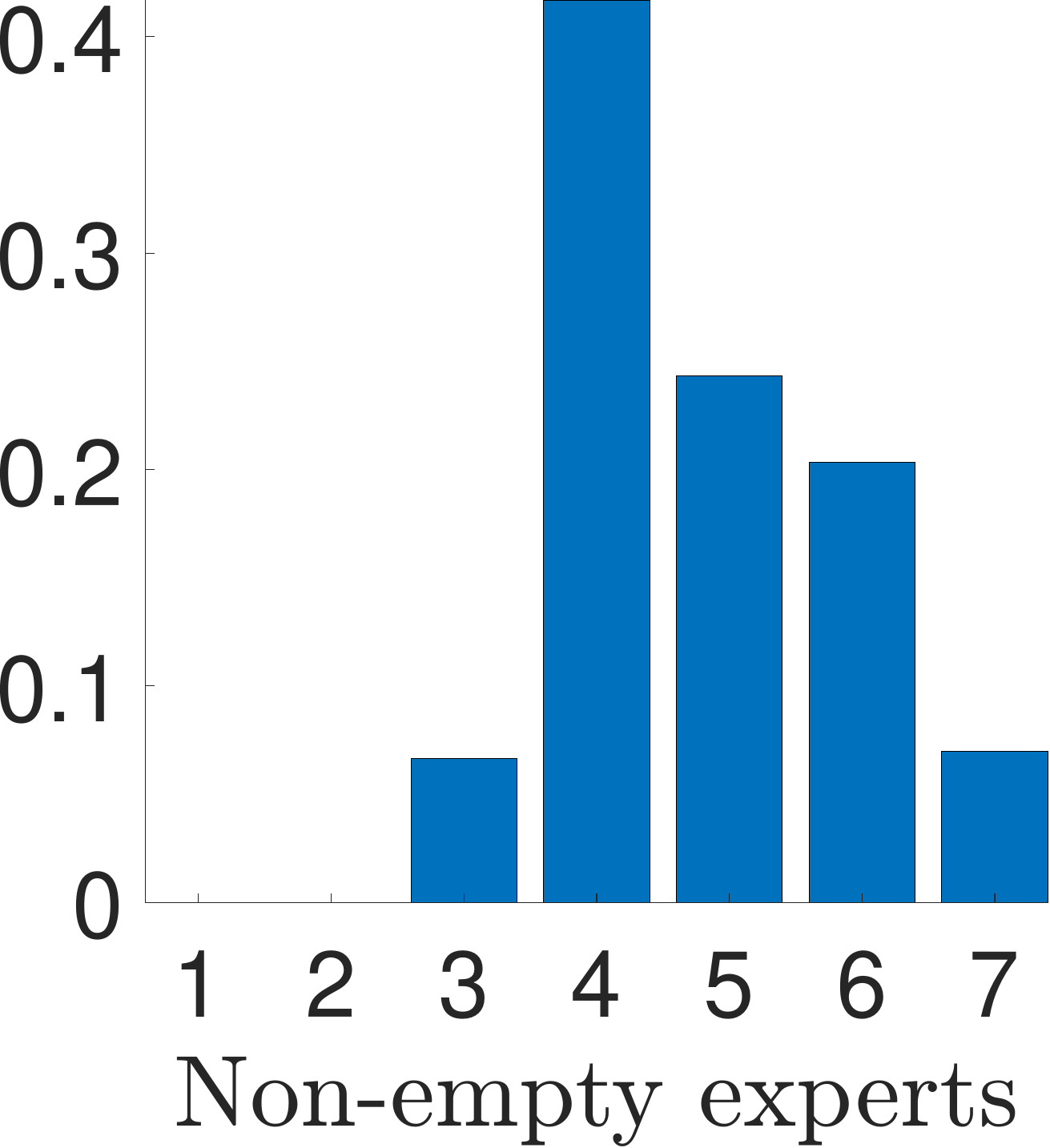}
    \end{minipage}
    \hspace{\fill} 
    \begin{minipage}{0.24\textwidth}
    {\includegraphics[height = 2.8cm]{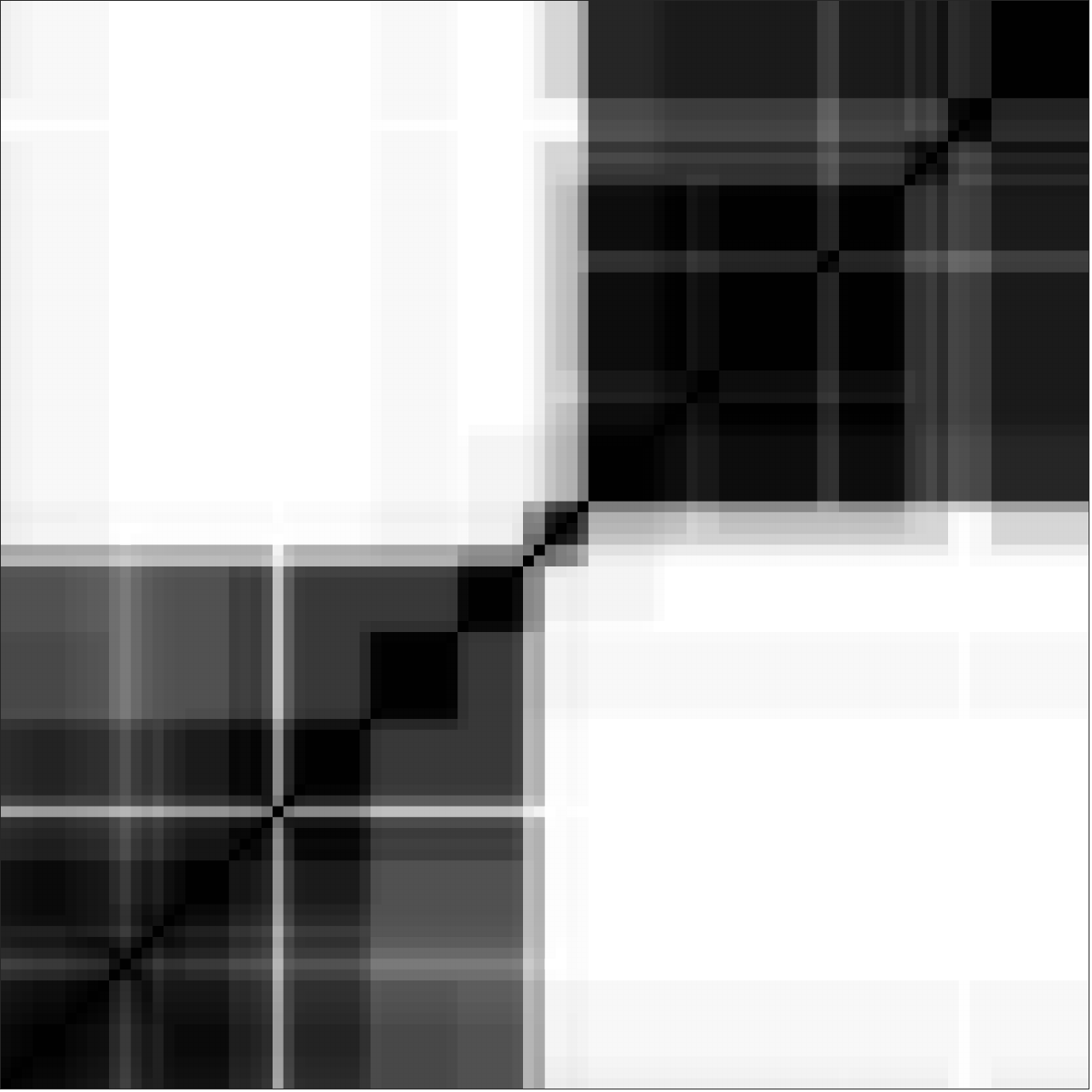}}
    \vspace{0.575cm}
    \end{minipage}

    \vspace*{0.5cm} 
    
    \caption{\textit{First synthetic data set}: non-stationary function, homoskedastic noise. On the left, the highest density regions and median for SMC$^2$-MoE are shown in gray and black, along with the data points in blue, for different values of the Dirichlet concentration parameter $\alpha$. In the middle column, the posterior distribution for the number of non-empty experts is shown.  The last column shows the posterior similarity matrices summarizing the posterior over partitions. Values of $\alpha = 0.1$, $\alpha = 1$, and $\alpha = K/2$ are used and correspond to the top, middle, and bottom rows, respectively.} \label{im:stationaryDensityPSM}
\end{figure}
%
%
\begin{figure}
    \begin{minipage}{0.40\textwidth}
    {\includegraphics[height=3.65cm]{pics/vector/nonstationary/nonstationary_density_a01.pdf}}
    \end{minipage}
    \hspace{\fill} 
    \begin{minipage}{0.24\textwidth}
    \includegraphics[height=3.75cm]{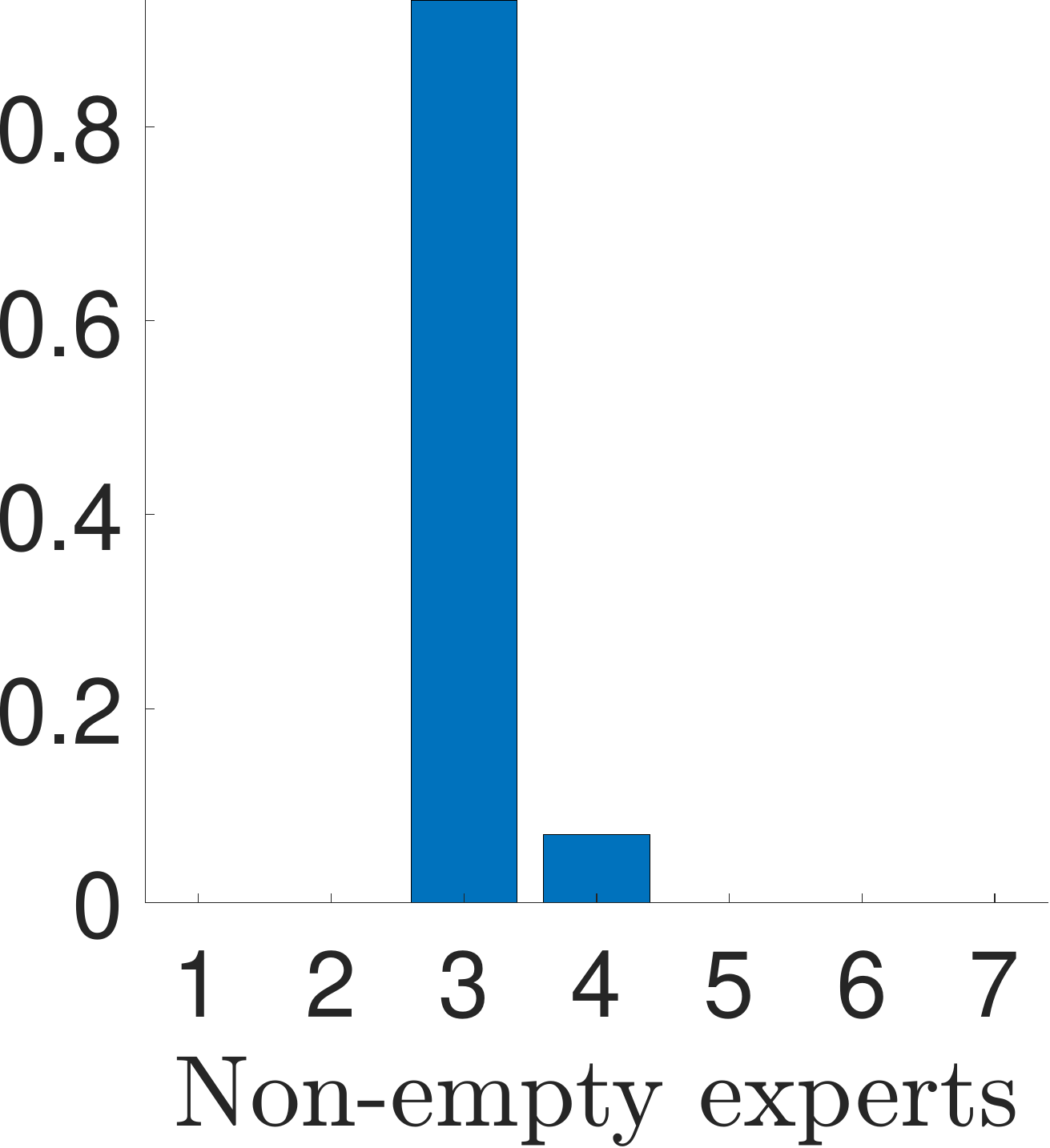}
    \end{minipage}
    \hspace{\fill} 
    \begin{minipage}{0.24\textwidth}
    {\includegraphics[height = 2.8cm]{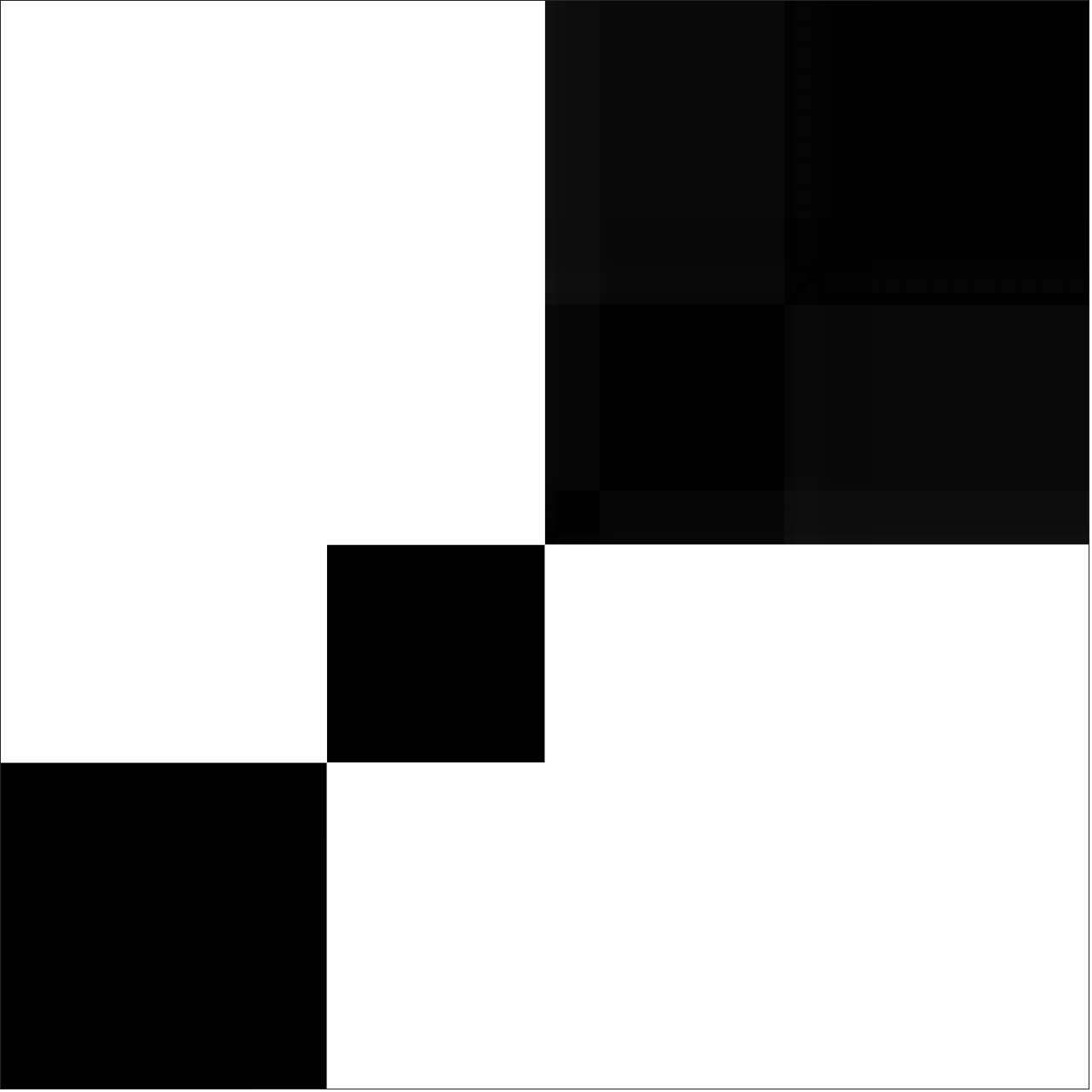}}
    \vspace{0.575cm}
    \end{minipage}

    \vspace*{1cm} 
    
    \begin{minipage}{0.40\textwidth}
    {\includegraphics[height=3.65cm]{pics/vector/nonstationary/nonstationary_density_a1.pdf}}
    \end{minipage}
    \hspace{\fill} 
    \begin{minipage}{0.24\textwidth}
    \includegraphics[height=3.75cm]{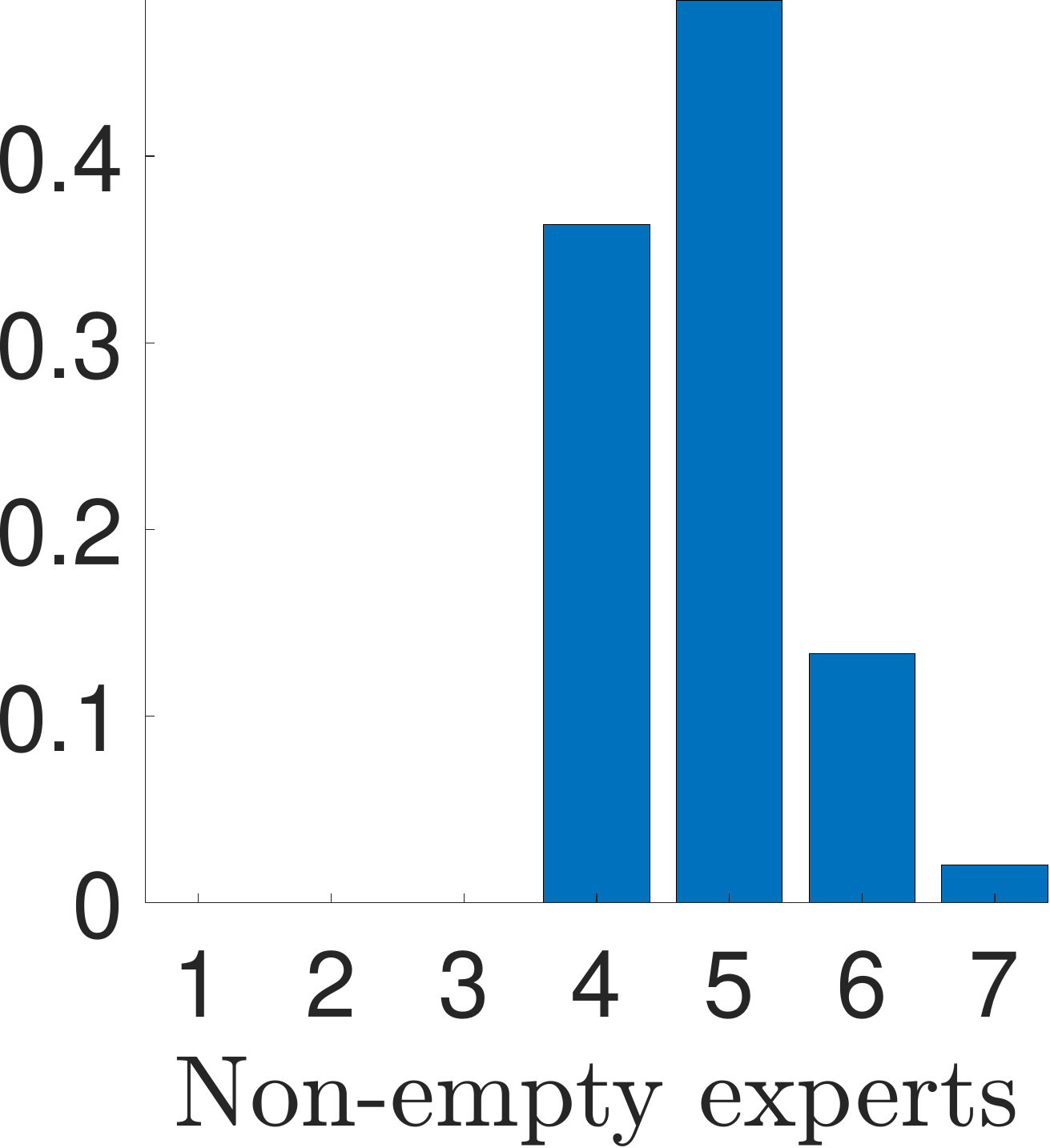}
    \end{minipage}
    \hspace{\fill} 
    \begin{minipage}{0.24\textwidth}
    {\includegraphics[height = 2.8cm]{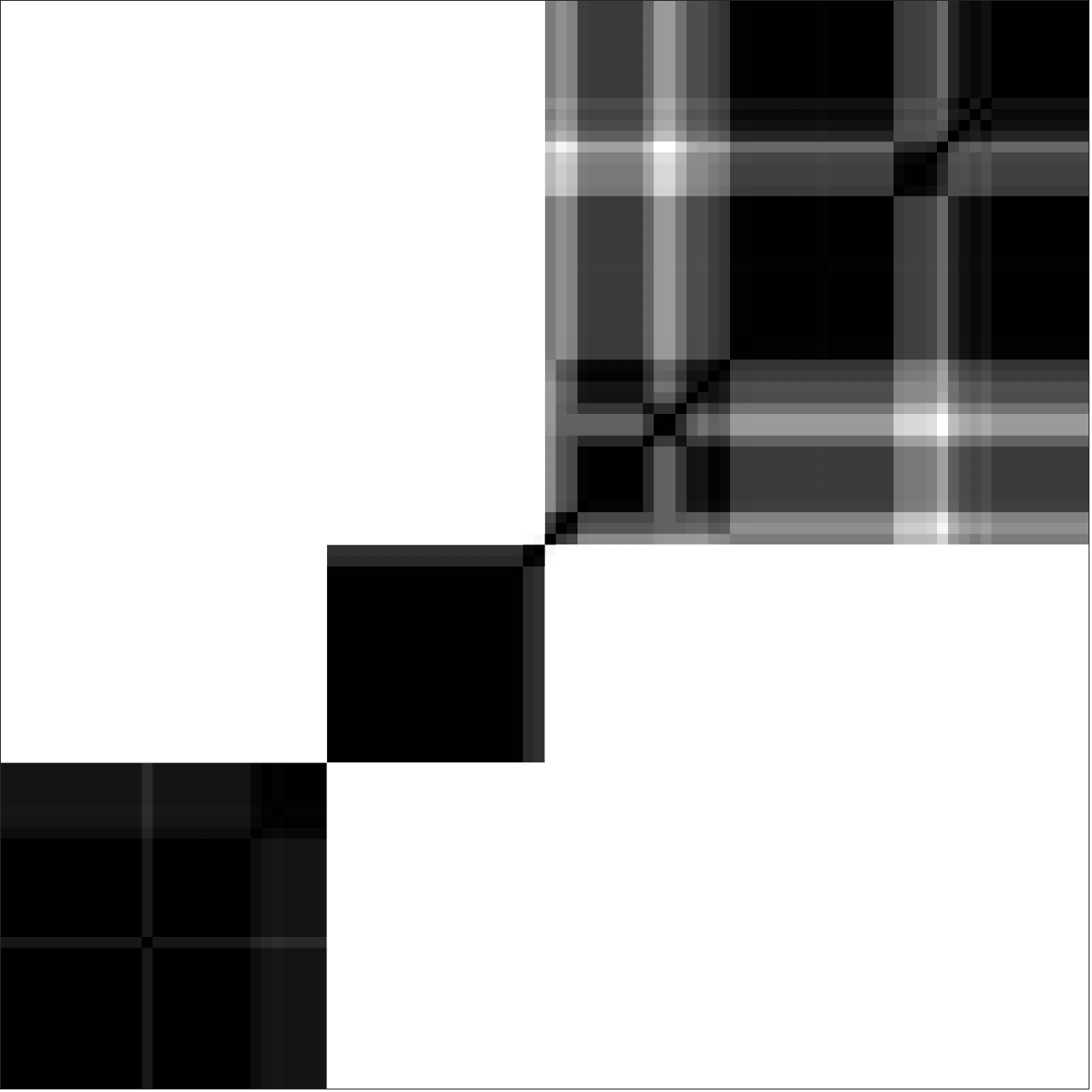}}
    \vspace{0.575cm}
    \end{minipage}

    \vspace*{1cm} 

    \begin{minipage}{0.40\textwidth}
    {\includegraphics[height=3.65cm]{pics/vector/nonstationary/nonstationary_density_aJeffreys.pdf}}
    \end{minipage}
    \hspace{\fill} 
    \begin{minipage}{0.24\textwidth}
    \includegraphics[height=3.75cm]{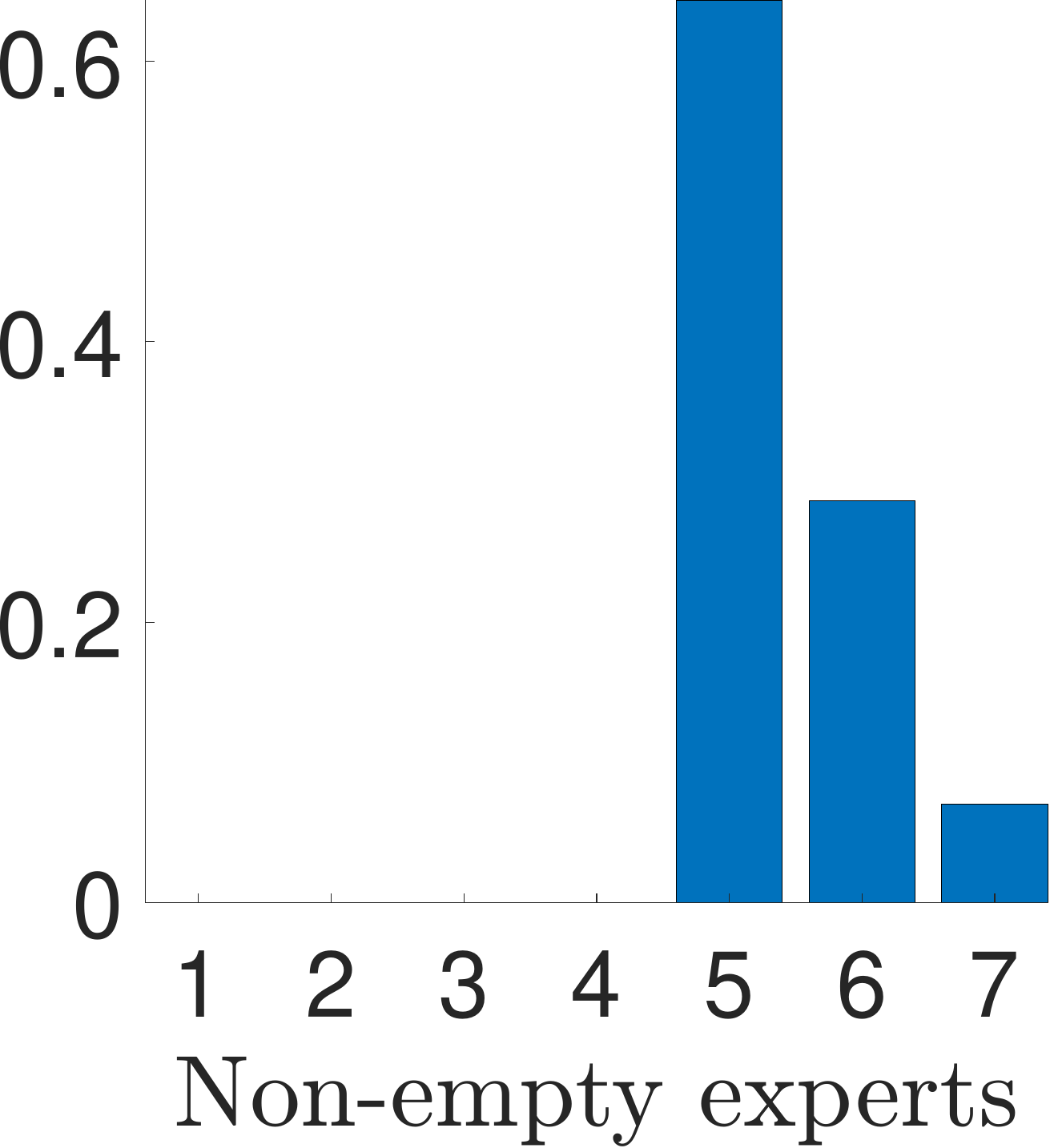}
    \end{minipage}
    \hspace{\fill} 
    \begin{minipage}{0.24\textwidth}
    {\includegraphics[height = 2.8cm]{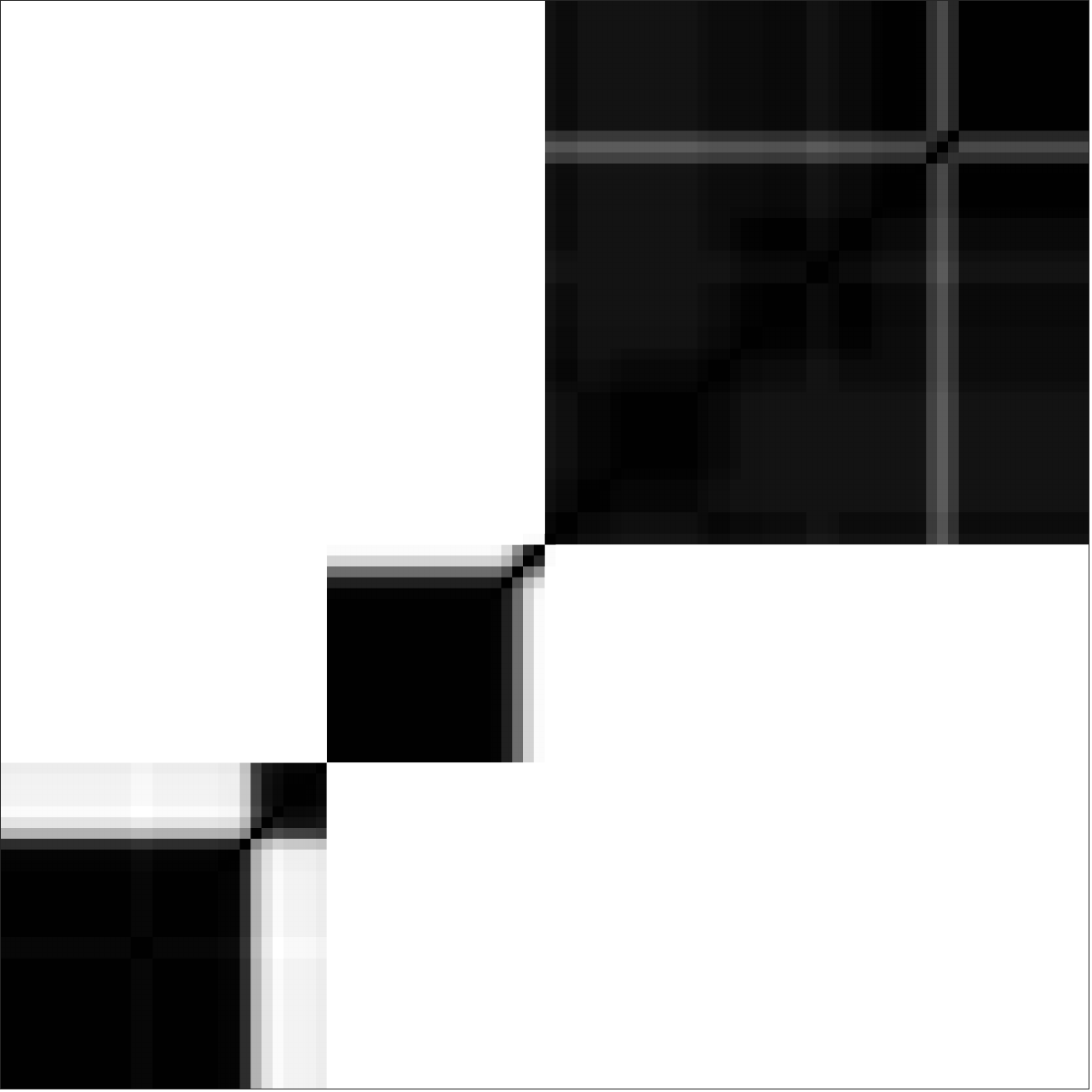}}
    \vspace{0.575cm}
    \end{minipage}

    \vspace*{0.5cm} 
    
    \caption{\textit{Second synthetic data set}: discontinuous function, heteroskedastic noise. On the left, the highest density regions and median for SMC$^2$-MoE are shown in gray and black, along with the data points in blue, for different values of the Dirichlet concentration parameter $\alpha$. In the middle column, the posterior distribution for the number of non-empty experts is shown. The last column shows the posterior similarity matrices summarizing the posterior over partitions. Values of $\alpha = 0.1$, $\alpha = 1$, and $\alpha = K/2$ are used and correspond to the top, middle, and bottom rows, respectively. } \label{im:nonstationaryDensityPSM}
\end{figure}
For the synthetic data sets, we present the estimated predictive densities as 90\% highest density regions in Figures \ref{im:stationaryComparisonFigure} -- \ref{im:nonstationaryComparisonFigure} for both the MoE with SMC$^2$ (SMC$^2$-MoE) and the MoE with IS (IS-MoE), along with ground truth and the corresponding results obtained using a single GP, RBCM \citep{Deisenroth:2015, Liu:2018}, and treedGP \citep{Gramacy:2007, Gramacy:2008, Gramacy:2010}.
For IS-MoE and SMC$^2$-MoE,  we employ an upper bound of $K=7$ experts and consider different choices of the Dirichlet concentration parameter $\alpha = 0.1, 1, K/2$.
For IS-MoE, we use the MAP approximation used in the original paper without any additional modifications such as stochastic sub-sampling.
In all cases, the number of outer particles $J$, inner particles $M$, and time steps $T$ are chosen so that $MJT^2$ is similar to the number of IS samples in order to have similar run times for IS and SMC$^2$ and fair comparisons based on a fixed computational budget.
The improvement in the estimated predictive densities with SMC$^2$ is clearly evident.
For the first data set in Figure \ref{im:stationaryComparisonFigure}, IS produces disperse densities that are too wiggly on the left side, while  SMC$^2$ produces density estimates that better match the ground truth.
Both the GP and RBCM, which is designed for distributed inference of a stationary GP model, also provide density estimates that are too wiggly on the left side, due to the non-stationary behavior of the underlying function.
Instead, the treedGP, which is a mixture of GP experts with a gating network that partitions the input space into axis-aligned regions, is better able to recover the non-stationary behavior of the true function.
However, we note that the treedGP R package \citep{Gramacy:2007} employs MCMC and thus, comes with an increased computational cost (see for example \citep{Zhang:2019} who find that treedGP increases wall-clock time by a factor of $\approx18$ compared to IS-MOE).  We note that the SMC$^2$ scheme developed here is general and could be used for other choices of gating networks, such as trees, for faster inference with treedGPs.
In the second data set, the partition plays a crucial role due to the discontinuous nature of the true function and heteroskedastic noise.
As postulated, in this setting, the results of IS-MoE in Figure \ref{im:nonstationaryComparisonFigure} are quite poor, while SMC$^2$-MoE provides substantial improvements, recovering well the ground truth.
The GP and RBCM either under-smooth or over-smooth due to the stationary assumption of the models.
The results with treedGP are better but highlight the difficulty of inferring the cut-off values determining the treed partition structure.
We also show $L^1$ and $L^2$ vector norms between the computed predictive densities, the ground truth densities and the L1 vector norm between the obtained predictive median and the ground truth median with predictive log-likelihood results in Tables  \ref{tb:stationaryDifferences} and \ref{tb:nonstationaryDifferences} for the two synthetic data sets.

Due to the important role of the partition on predictions, we study and visualize uncertainty in the clustering $C$ by constructing the posterior similarity matrix (PSM) \citep{Gadd:2020}.
The PSM is an $N$ by $N$ matrix, whose elements represent the posterior probability that two data points are clustered together, i.e. the $ii'$th element is given by $p( c_i = c_{i'} \mid X, Y)$, which is approximated by the weighted average across the particles of the indicators that $c_i$ and $c_{i'}$ belong to the same expert.
For the synthetic data sets, Figures \ref{im:stationaryDensityPSM}--\ref{im:nonstationaryDensityPSM} illustrate the posterior for the number of non-empty experts (middle column) and the posterior similarity matrices (right column) in comparison to the highest density region (left column), with rows corresponding to different values of the Dirichlet concentration parameter $\alpha=0.1,1,K/2$.
The posterior similarity matrices also show how concentrated the sampled partitions $C_j$ are with a lack of more uncertain gray areas in the PSM showing a more concentrated estimate for the partition.
A smaller value of $\alpha$ encourages sparsity and fewer clusters.
In particular, when $\alpha$ is small, we obtain larger clusters, and thus increased computational complexity, but smoother estimates.
On the other hand, when $\alpha$ is large, we obtain more clusters of small sizes and reduced computational costs, but estimates tend to be less smooth.
In general, we recommend selecting $K$ to be large enough (an upper bound on the number of clusters) and choosing $\alpha$ to balance computational cost and smoothness in the estimates. 
\begin{figure}[!t]
    \begin{minipage}{0.40\textwidth}
    {\includegraphics[height=3.65cm]{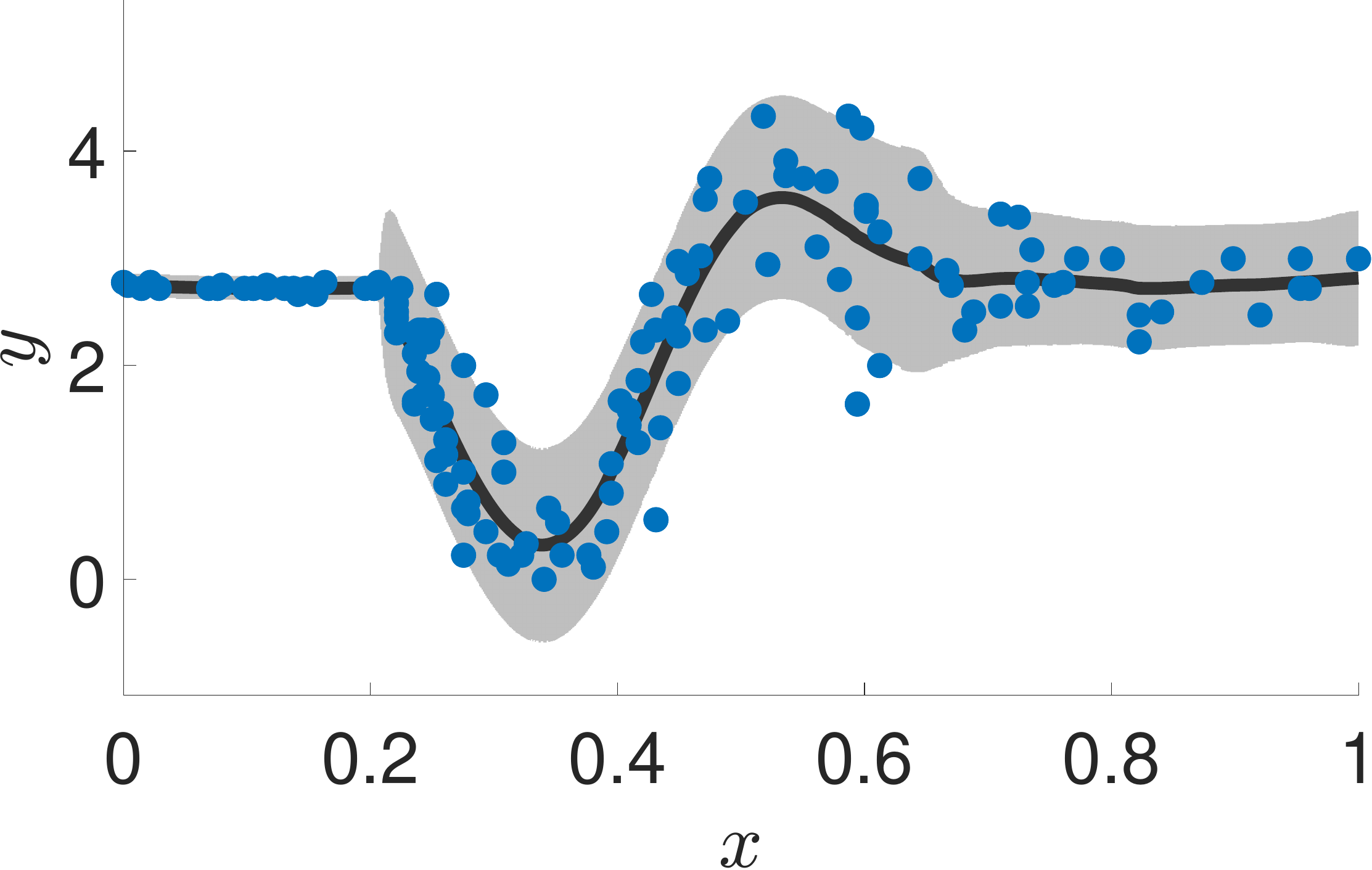}}
    \end{minipage}
    \hspace{\fill} 
    \begin{minipage}{0.24\textwidth}
    \includegraphics[height=3.75cm]{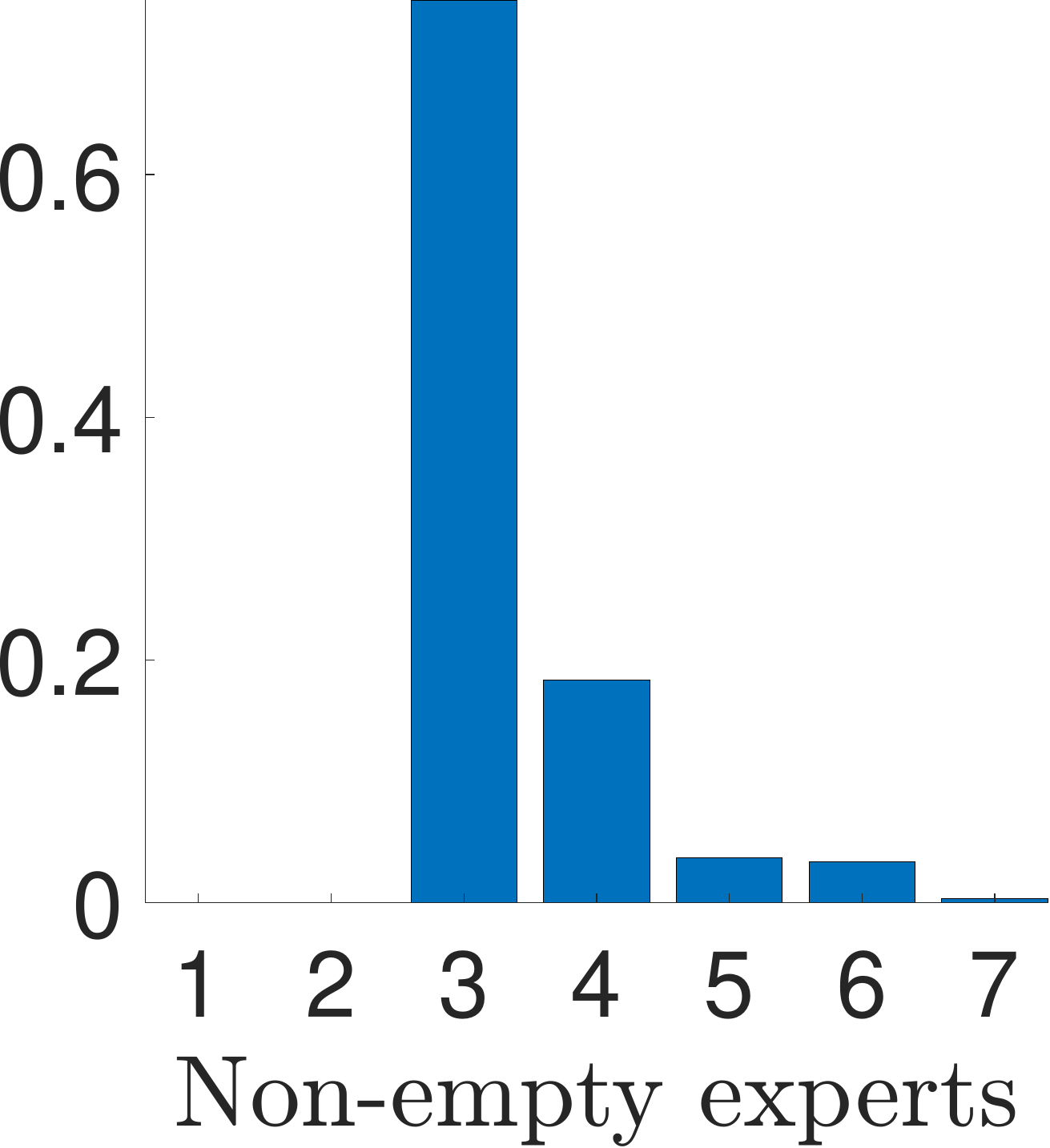}
    \end{minipage}
    \hspace{\fill} 
    \begin{minipage}{0.24\textwidth}
    {\includegraphics[height = 2.8cm]{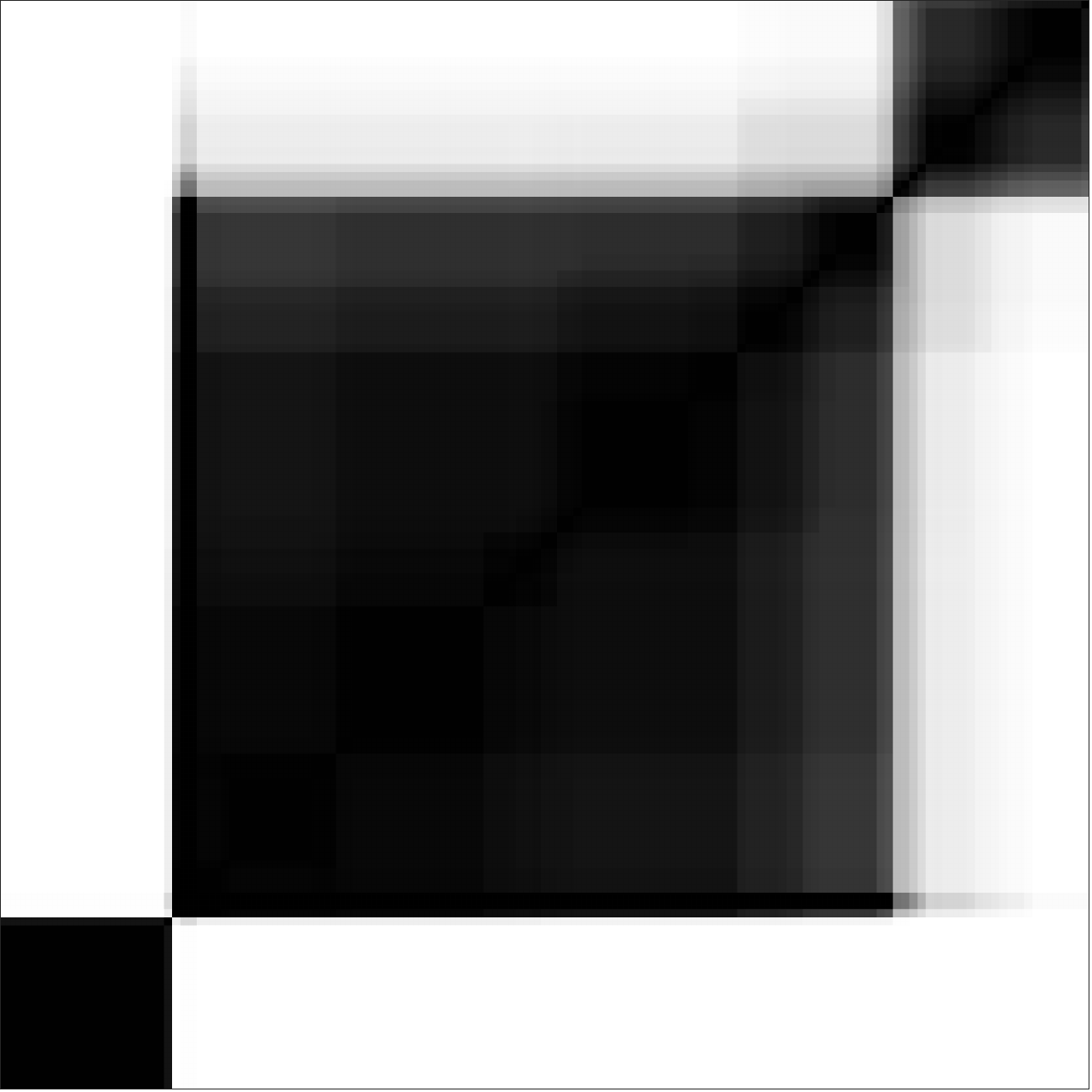}}
    \vspace{0.575cm}
    \end{minipage}

    \vspace*{1cm} 
    
    \begin{minipage}{0.40\textwidth}
    {\includegraphics[height=3.65cm]{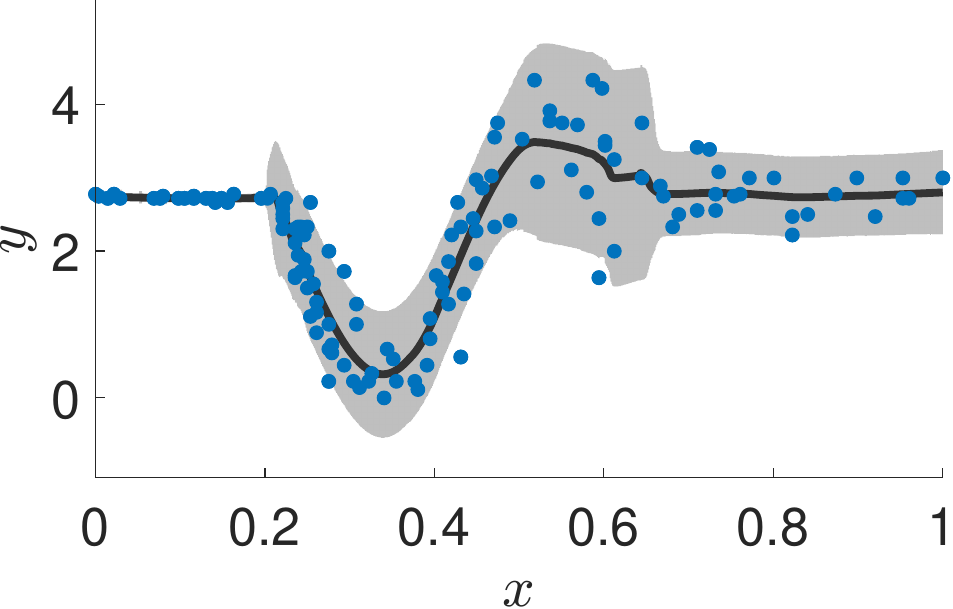}}
    \end{minipage}
    \hspace{\fill} 
    \begin{minipage}{0.24\textwidth}
    \includegraphics[height=3.75cm]{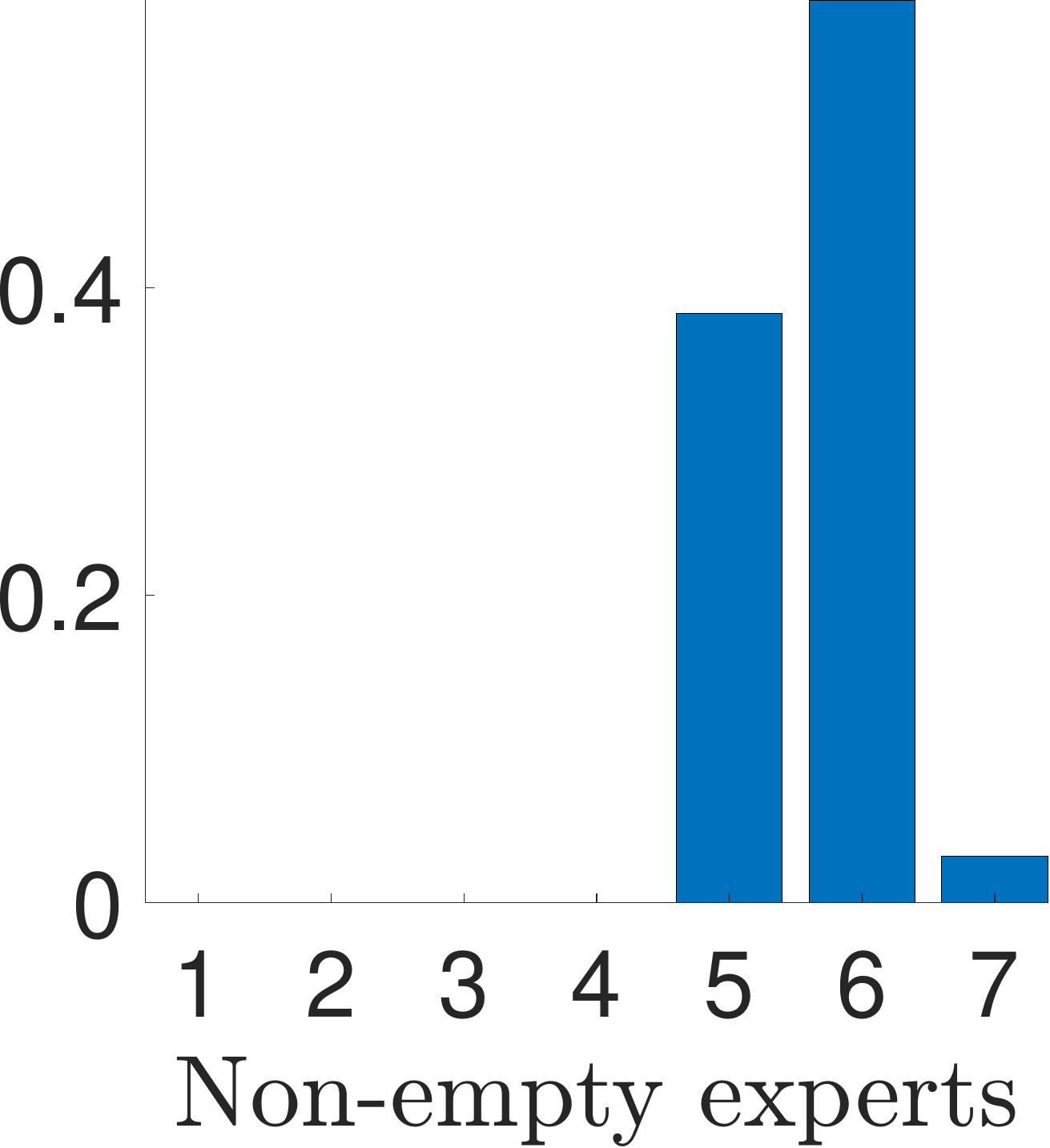}
    \end{minipage}
    \hspace{\fill} 
    \begin{minipage}{0.24\textwidth}
    {\includegraphics[height = 2.8cm]{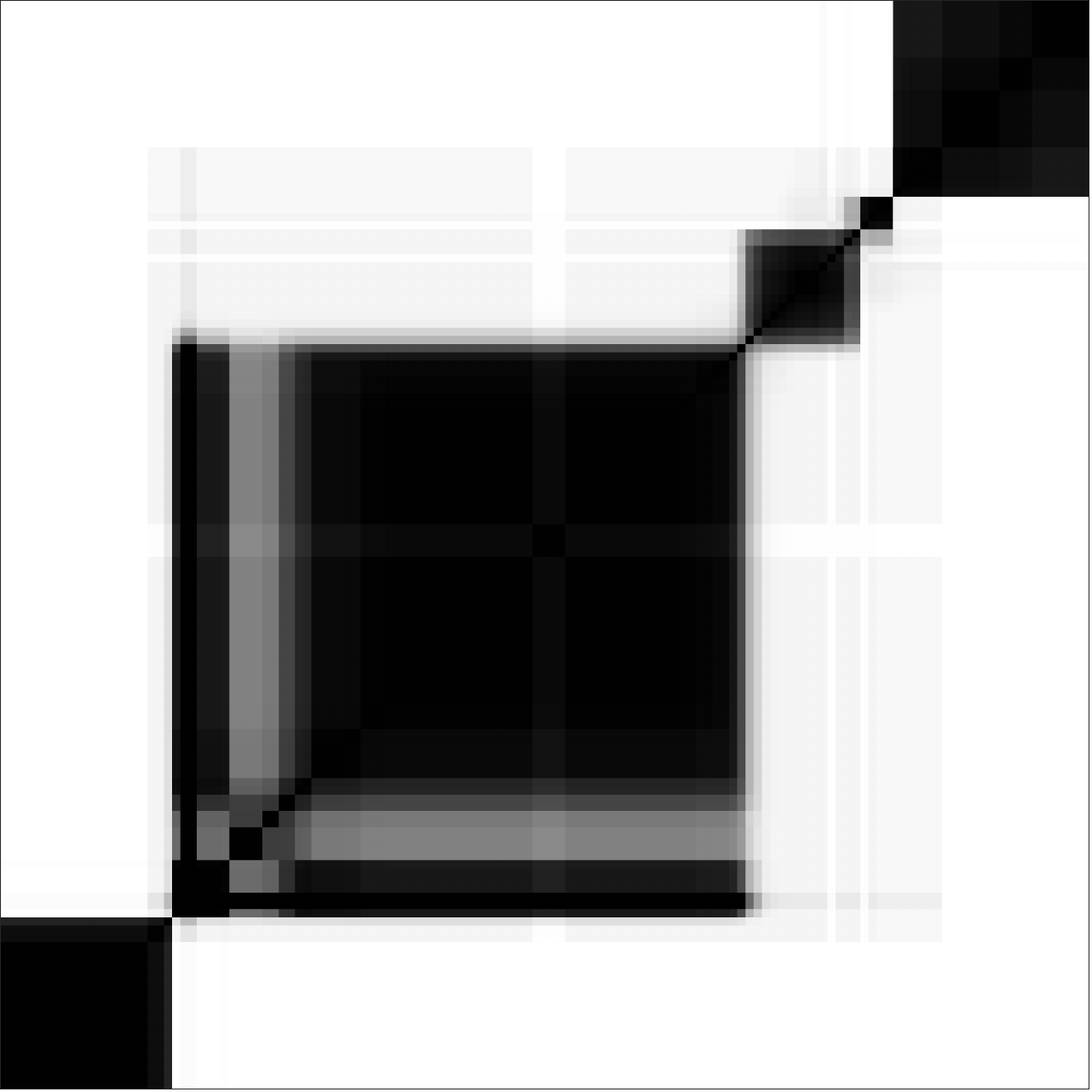}}
    \vspace{0.575cm}
    \end{minipage}

    \vspace*{1cm} 
    
    \begin{minipage}{0.40\textwidth}
    {\includegraphics[height=3.65cm]{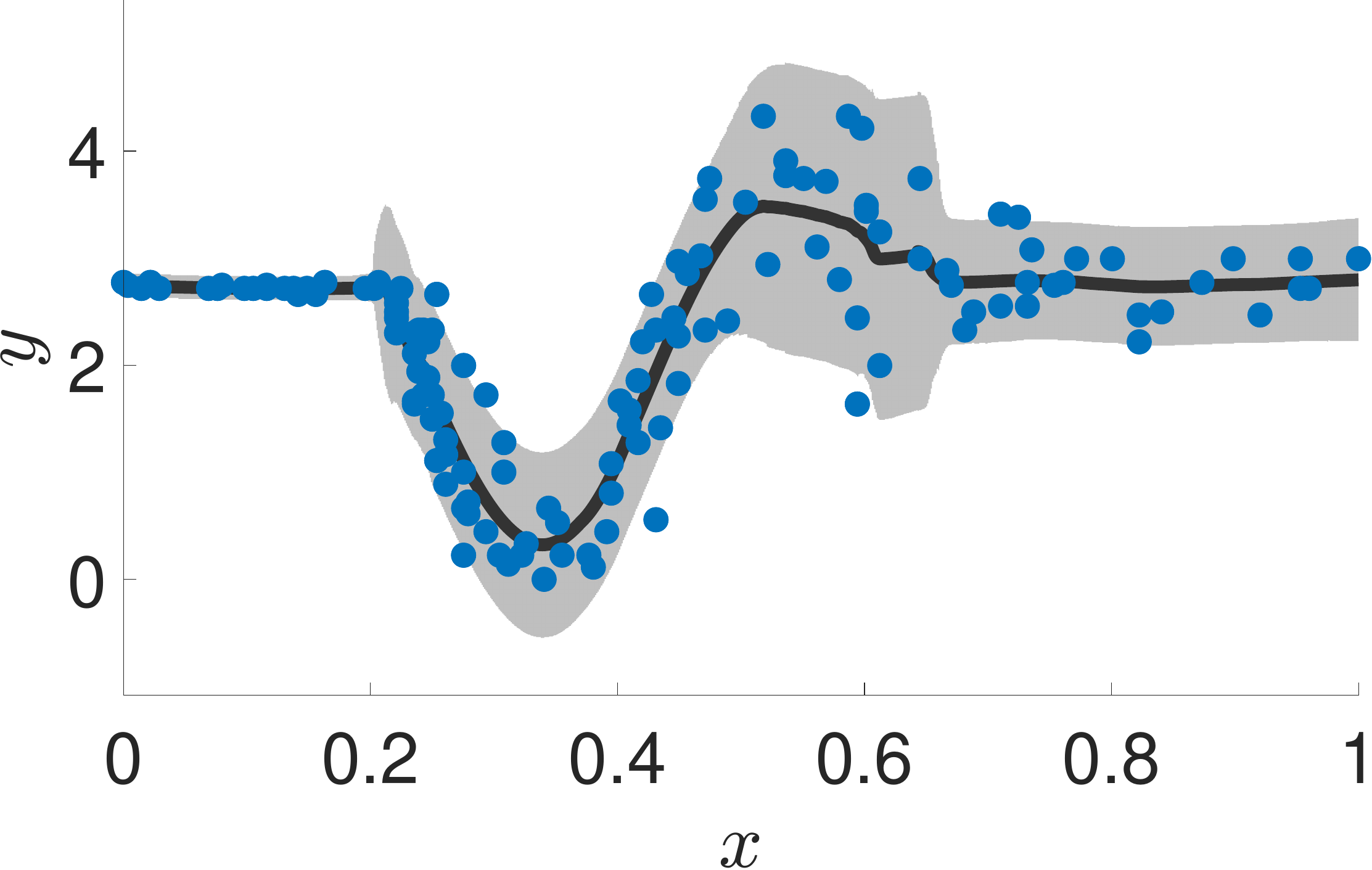}}
    \end{minipage}
    \hspace{\fill} 
    \begin{minipage}{0.24\textwidth}
    \includegraphics[height=3.75cm]{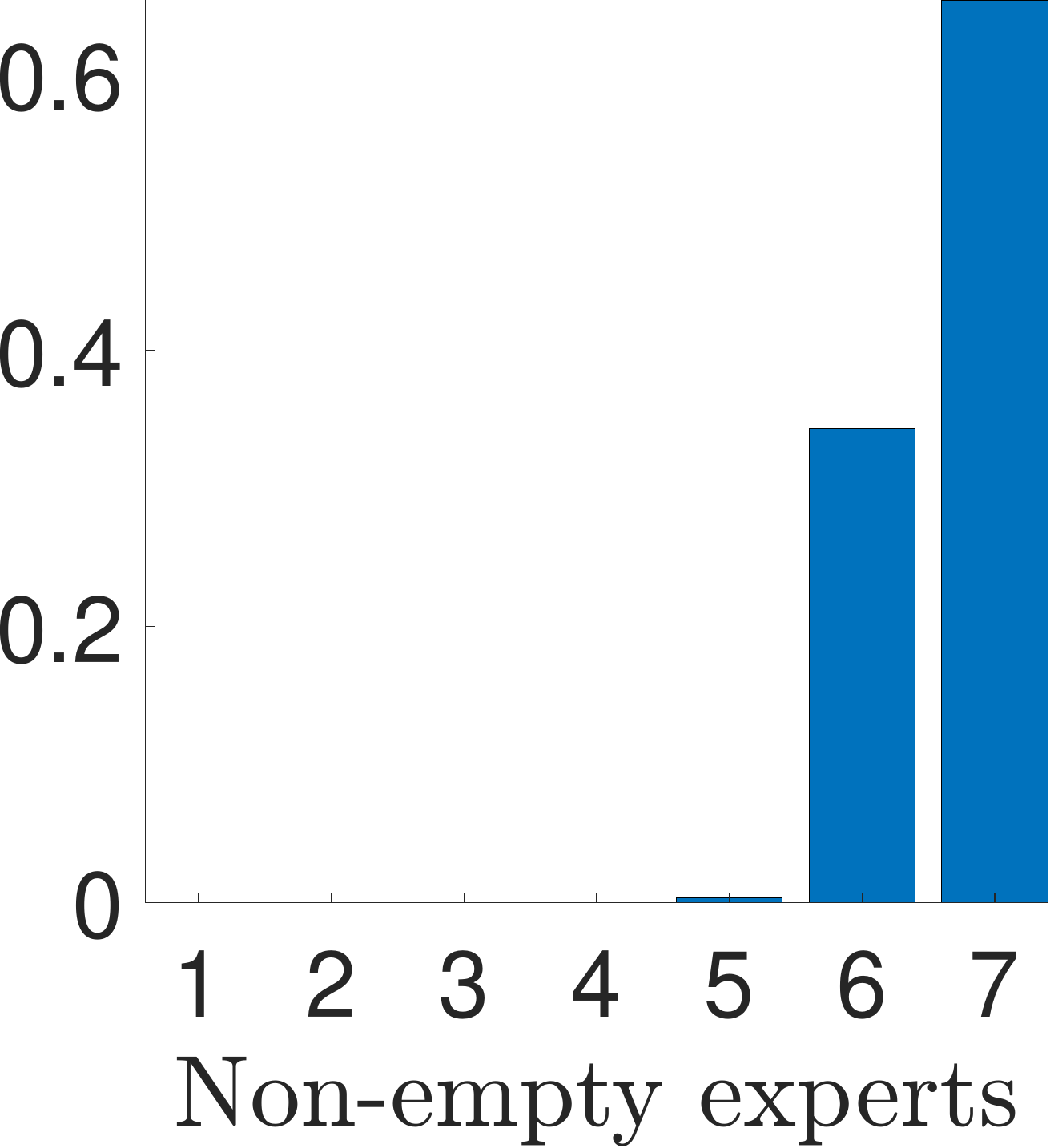}
    \end{minipage}
    \hspace{\fill} 
    \begin{minipage}{0.24\textwidth}
    {\includegraphics[height = 2.8cm]{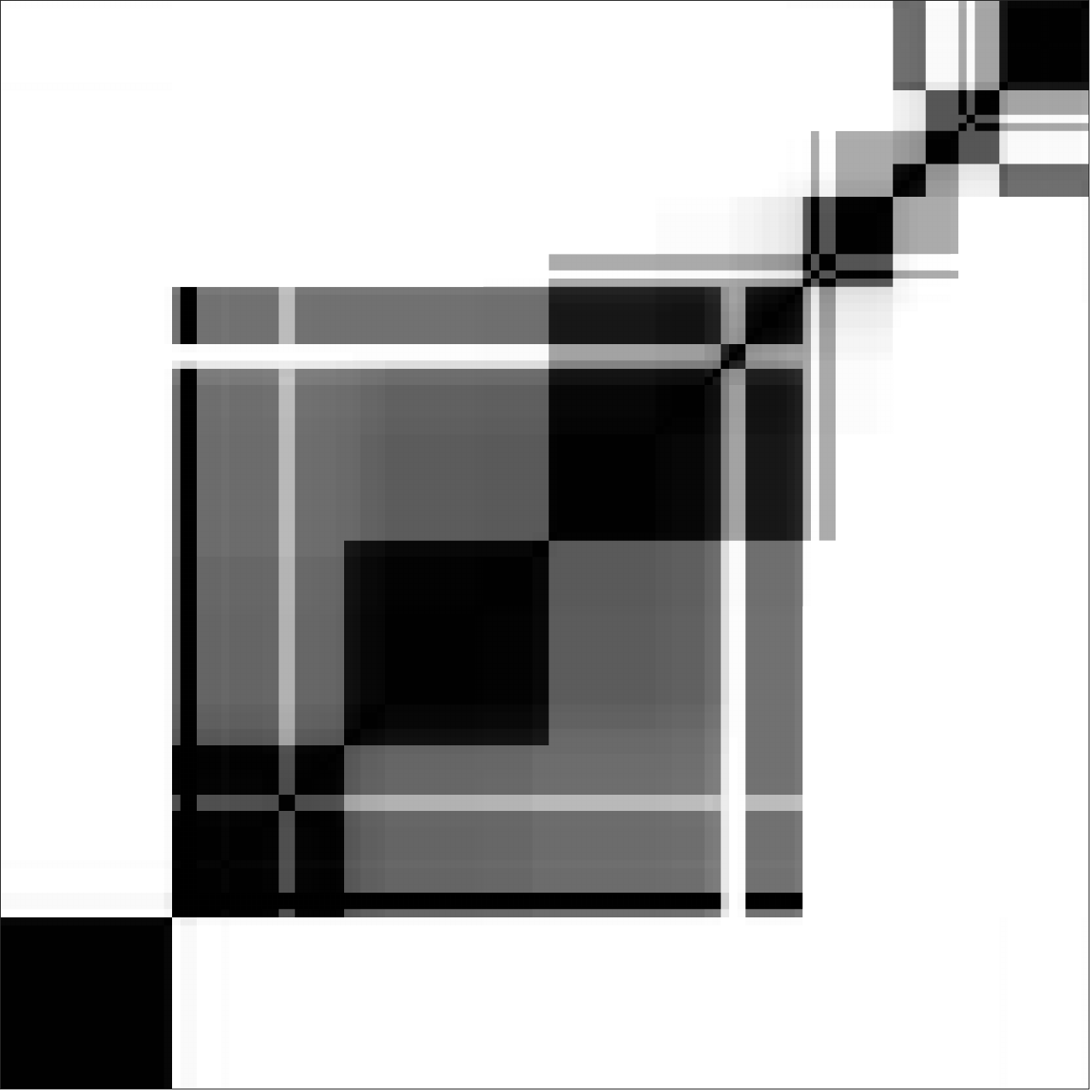}}
    \vspace{0.575cm}
    \end{minipage}

    \vspace*{0.5cm} 
    
    \caption{\textit{Motorcycle data}. On the left, the highest density regions and median are shown in gray and black along with the data points in blue for different values of the Dirichlet concentration parameter $\alpha$. In the middle column, the posterior distribution for the number of non-empty experts is shown. The last column shows the posterior similarity matrices summarizing the posterior over partitions. Values of $\alpha = 0.1$, $\alpha = 1$, and $\alpha = K/2$ are used and correspond to the top, middle, and bottom plots, respectively. } \label{im:motorcycleDensityPSM}
\end{figure}
\begin{figure}
    \begin{minipage}{0.40\textwidth}
    \includegraphics[height = 3.65cm]{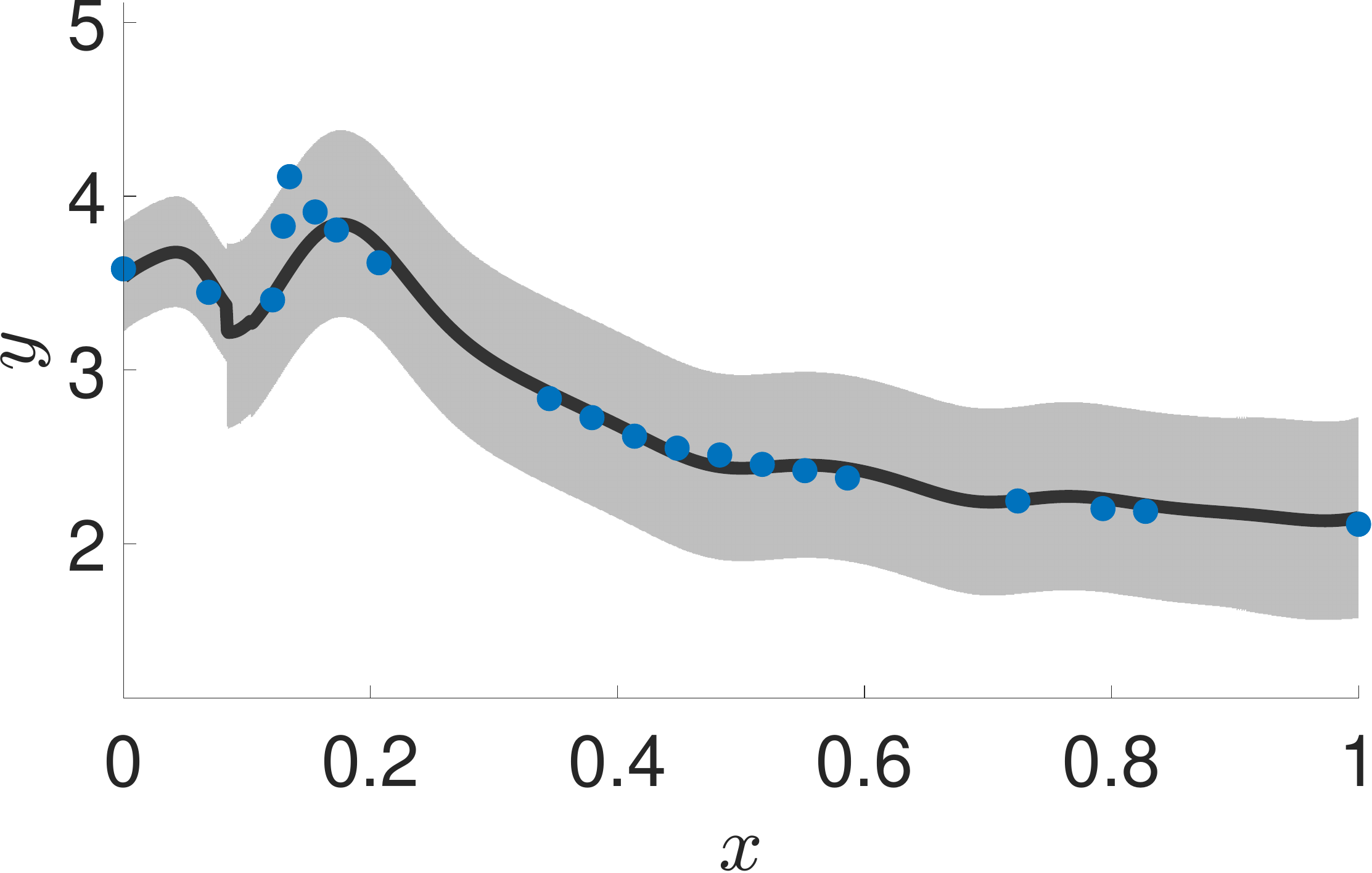}
    \end{minipage}
    \hspace{\fill} 
    \begin{minipage}{0.24\textwidth}
    \includegraphics[height = 3.75cm]{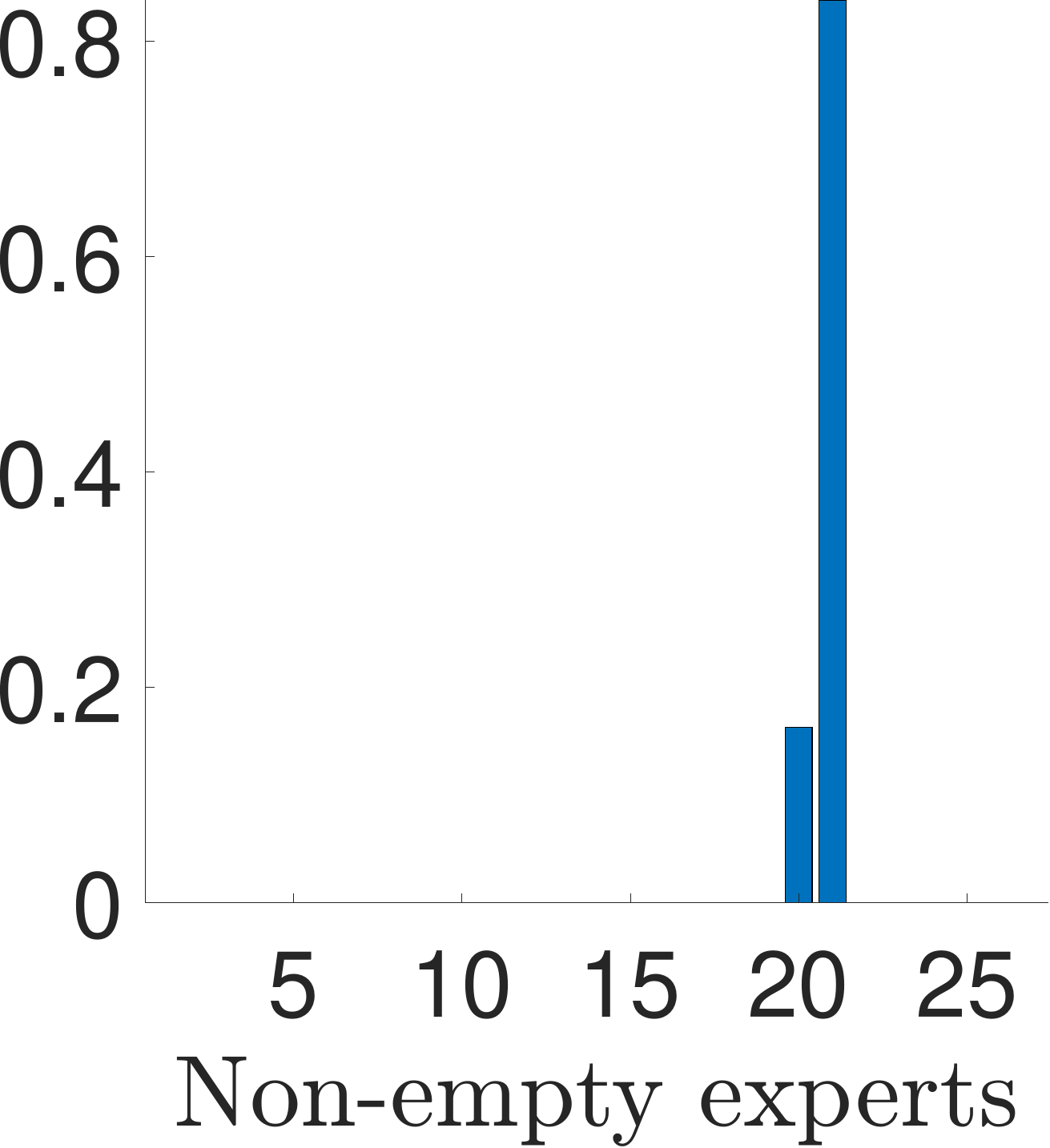}
    \end{minipage}
    \hspace{\fill} 
    \begin{minipage}{0.24\textwidth}
    \includegraphics[height = 2.8cm]{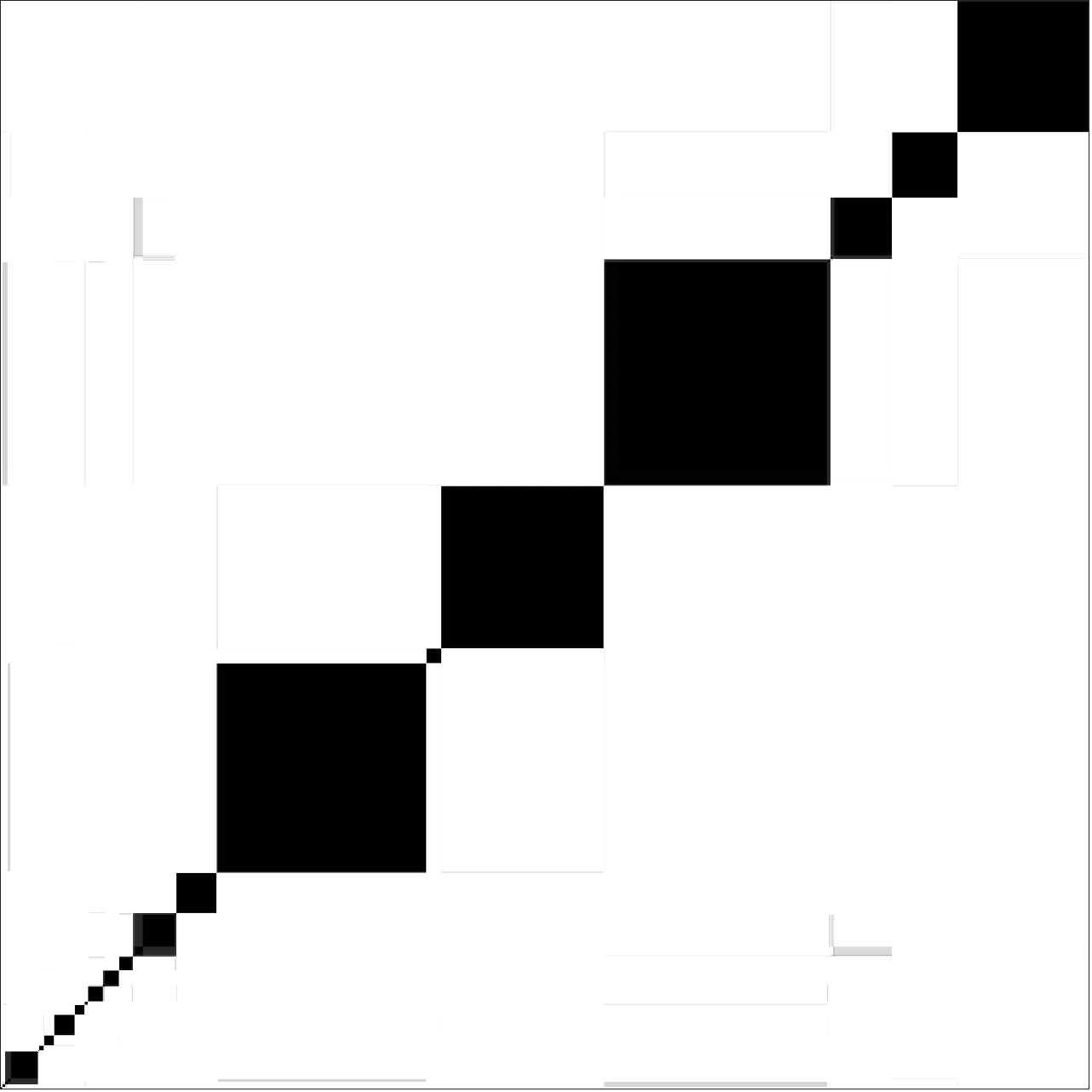}
    \vspace{0.575cm}
    \end{minipage}

    \vspace*{1cm} 
    
    \begin{minipage}{0.40\textwidth}
    \includegraphics[height = 3.65cm]{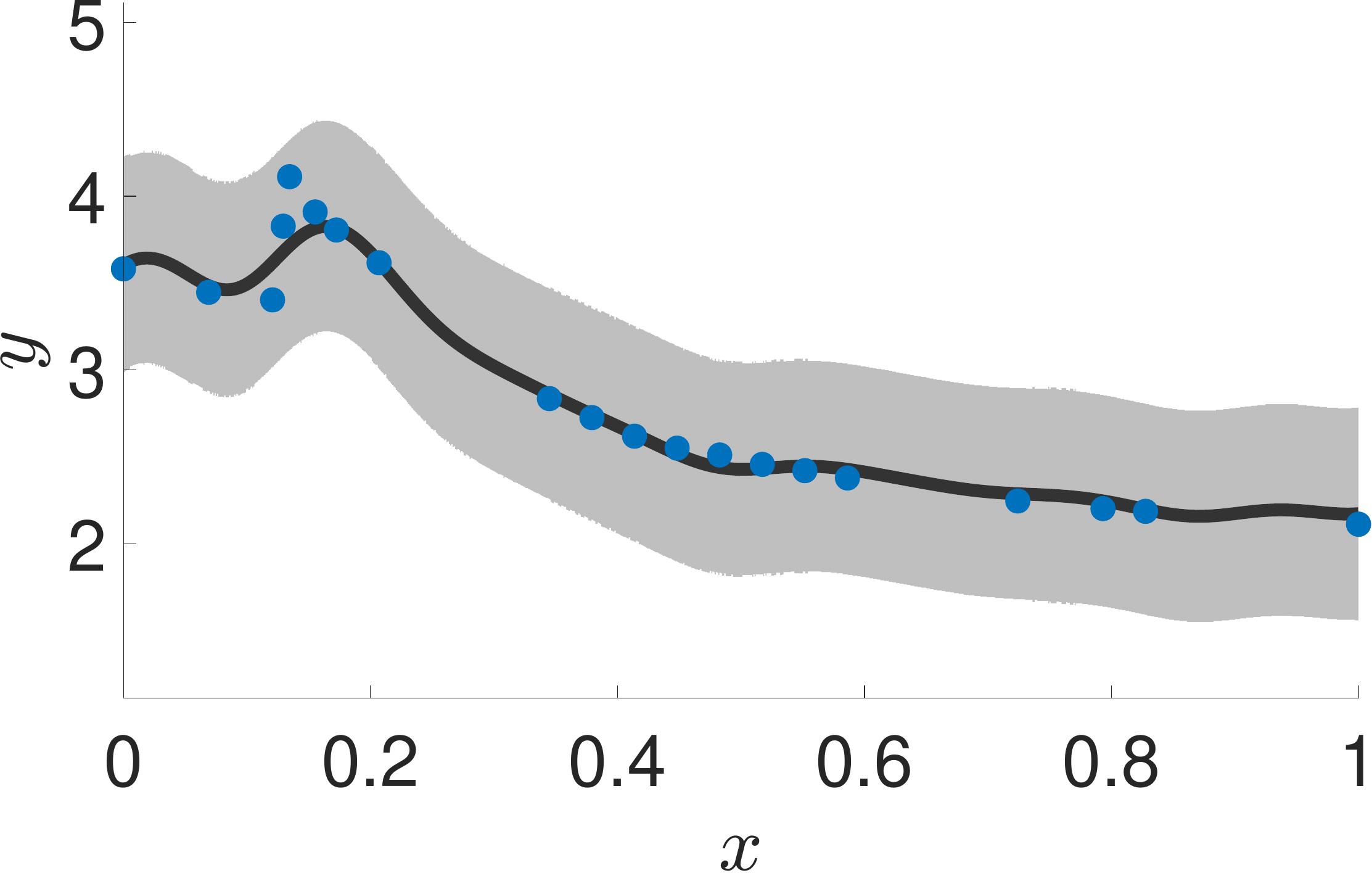}
    \end{minipage}
    \hspace{\fill} 
    \begin{minipage}{0.24\textwidth}
    \includegraphics[height = 3.75cm]{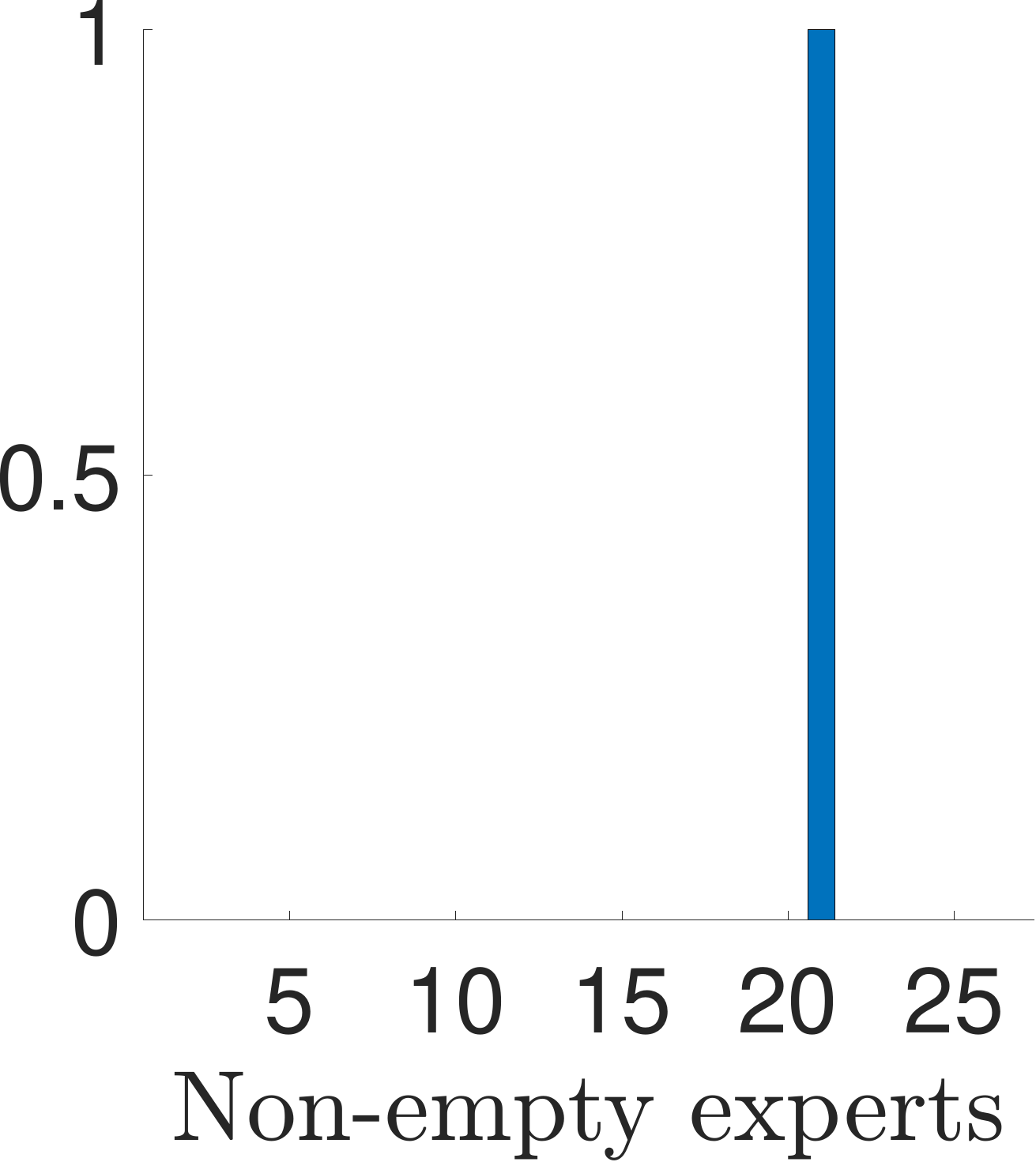}
    \end{minipage}
    \hspace{\fill} 
    \begin{minipage}{0.24\textwidth}
    \includegraphics[height = 2.8cm]{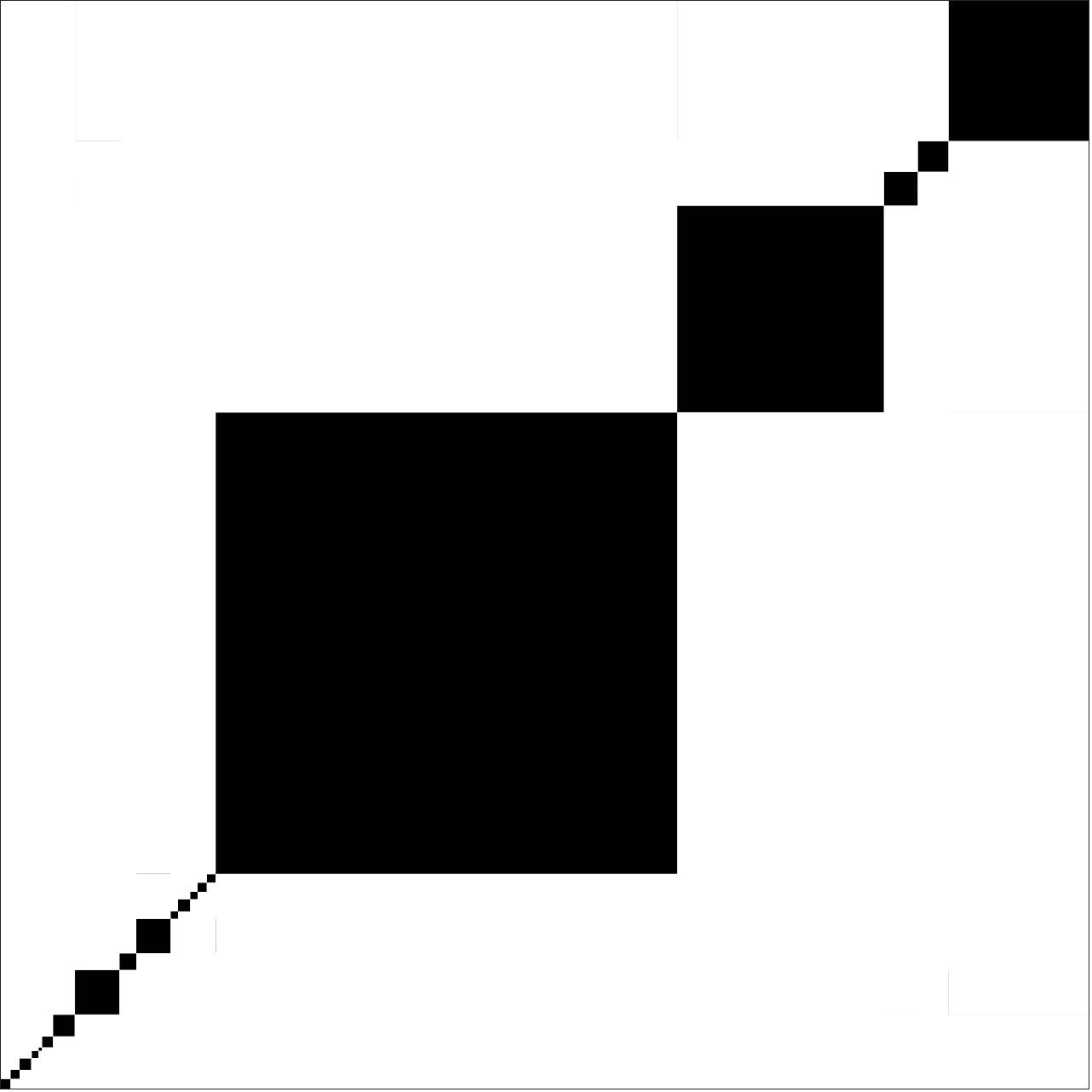}
    \vspace{0.575cm}
    \end{minipage}

    \vspace*{1cm} 

    \begin{minipage}{0.40\textwidth}
    \includegraphics[height = 3.65cm]{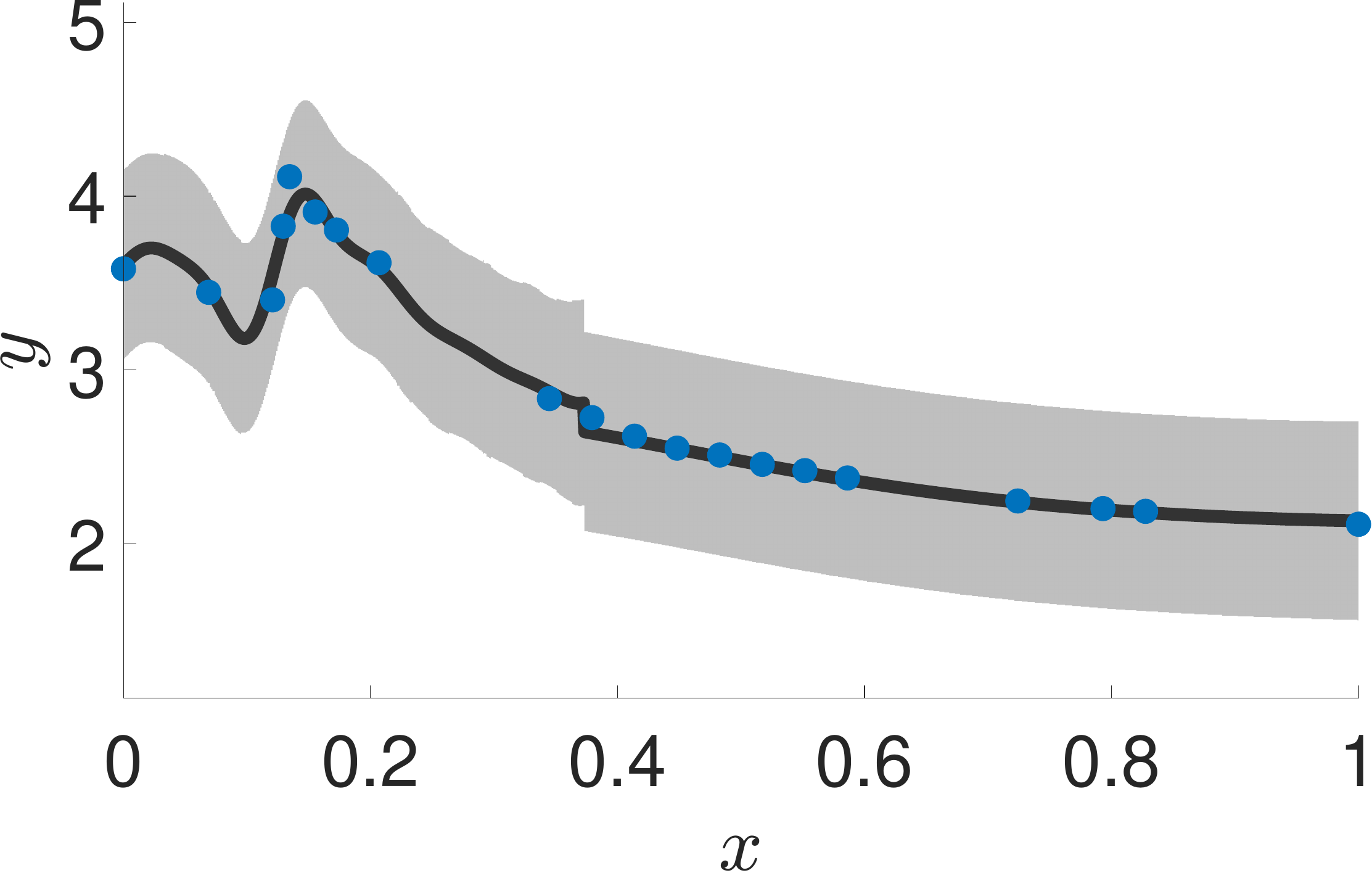}
    \end{minipage}
    \hspace{\fill} 
    \begin{minipage}{0.24\textwidth}
    \includegraphics[height = 3.75cm]{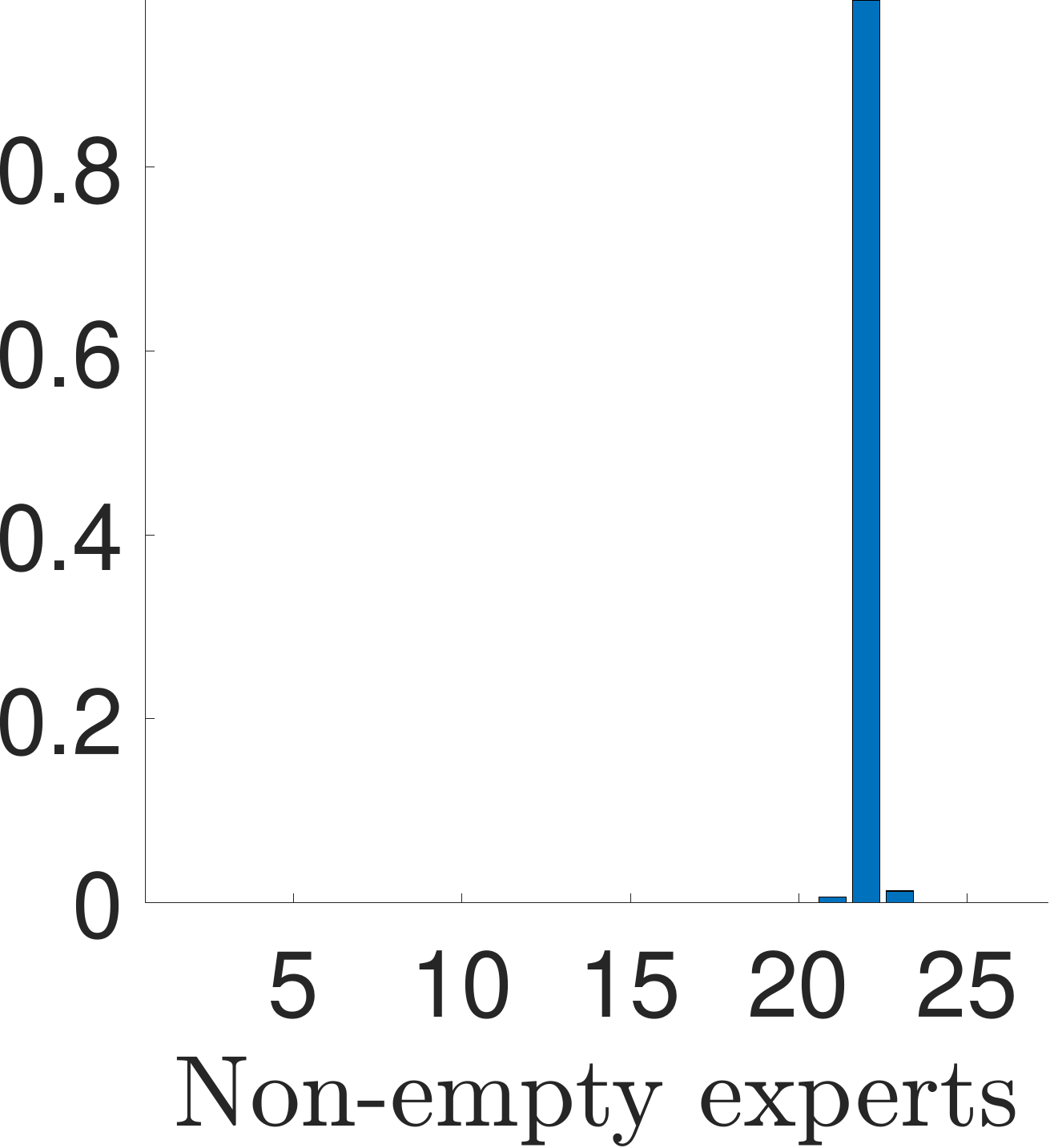}
    \end{minipage}
    \hspace{\fill} 
    \begin{minipage}{0.24\textwidth}
    \includegraphics[height = 2.8cm]{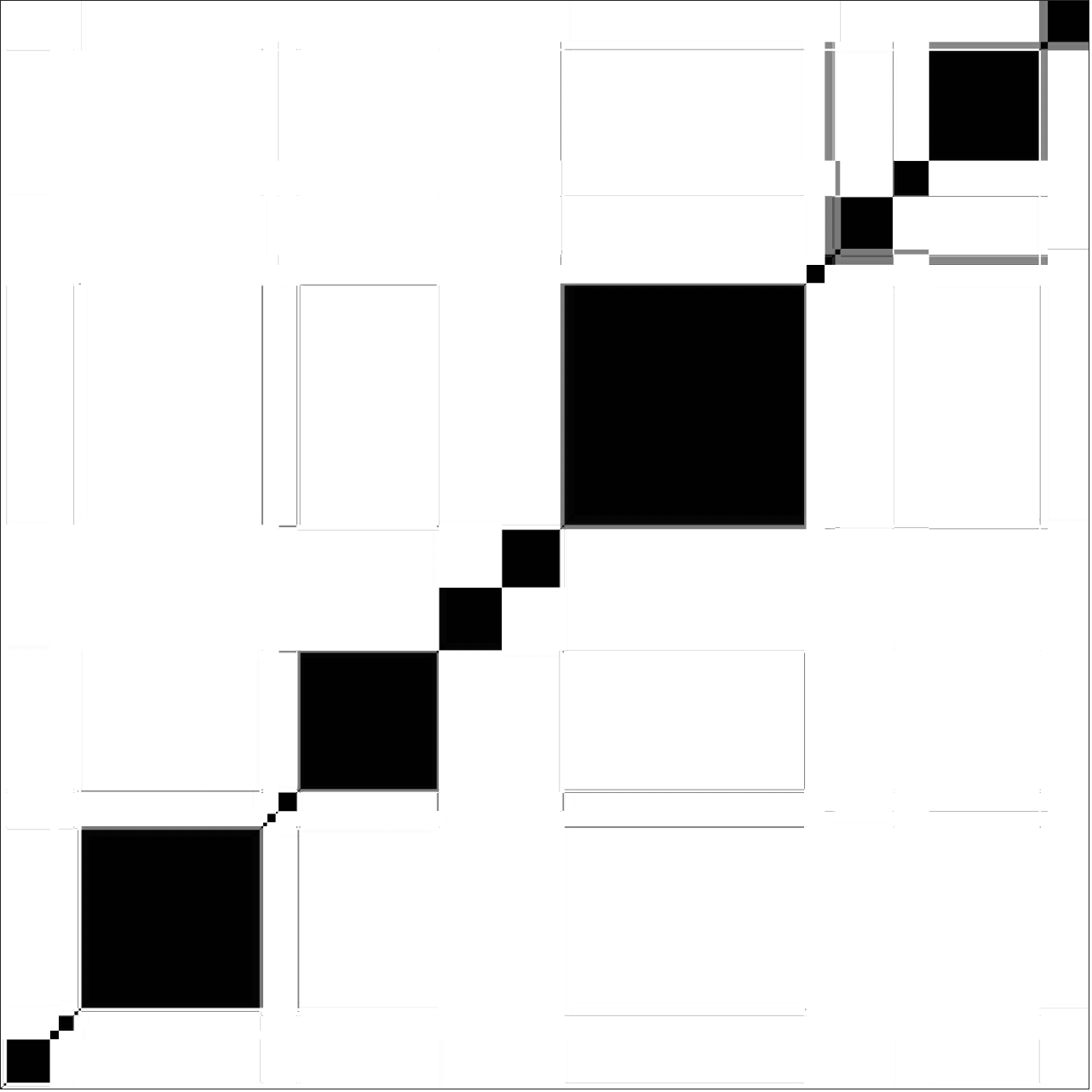}
    \vspace{0.575cm}
    \end{minipage}

    \vspace*{0.5cm} 
    
    \caption{\textit{3D NASA Langley glide-back booster simulation data}. On the left, the highest density regions and median for a one-dimensional slice of the input space are shown in gray and black along with the data points in blue for different values of the Dirichlet concentration parameter $\alpha$. In the middle column, the posterior distribution for the number of non-empty experts is shown.  The last column shows the posterior similarity matrices summarizing the posterior over partitions. Values of $\alpha = 2.7$, $\alpha = 6.75$, and $\alpha = K/2 = 13.5$ are used and correspond to the top, middle, and bottom plots, respectively.} \label{im:nasaDensityPSM}
\end{figure}

Results for the motorcycle data set with SMC$^2$-MoE and $K=7$ are provided in Figure \ref{im:motorcycleDensityPSM} and highlight the ability of the model to recover both the non-stationary behavior and heteroskedastic noise present in the data.
Rows correspond to different values of the Dirichlet concentration parameter $\alpha$, highlighting the trade-off between sparsity, computational cost, and smoothness.
For the NASA data, Figure \ref{im:nasaDensityPSM} illustrates the predictive densities as highest density regions for a one-dimensional slice of the 3D input space, along with the posterior over the number of clusters and the posterior similarity matrix for SMC$^2$-MoE and $K = 27$.
In this case, a larger number of experts and larger value of $\alpha$ are required to capture the ridge at $\text{Mach}=1$ showing the change from subsonic to supersonic flow.

\begin{figure}
    \begin{minipage}{0.40\textwidth}
    \includegraphics[height=3.65cm]{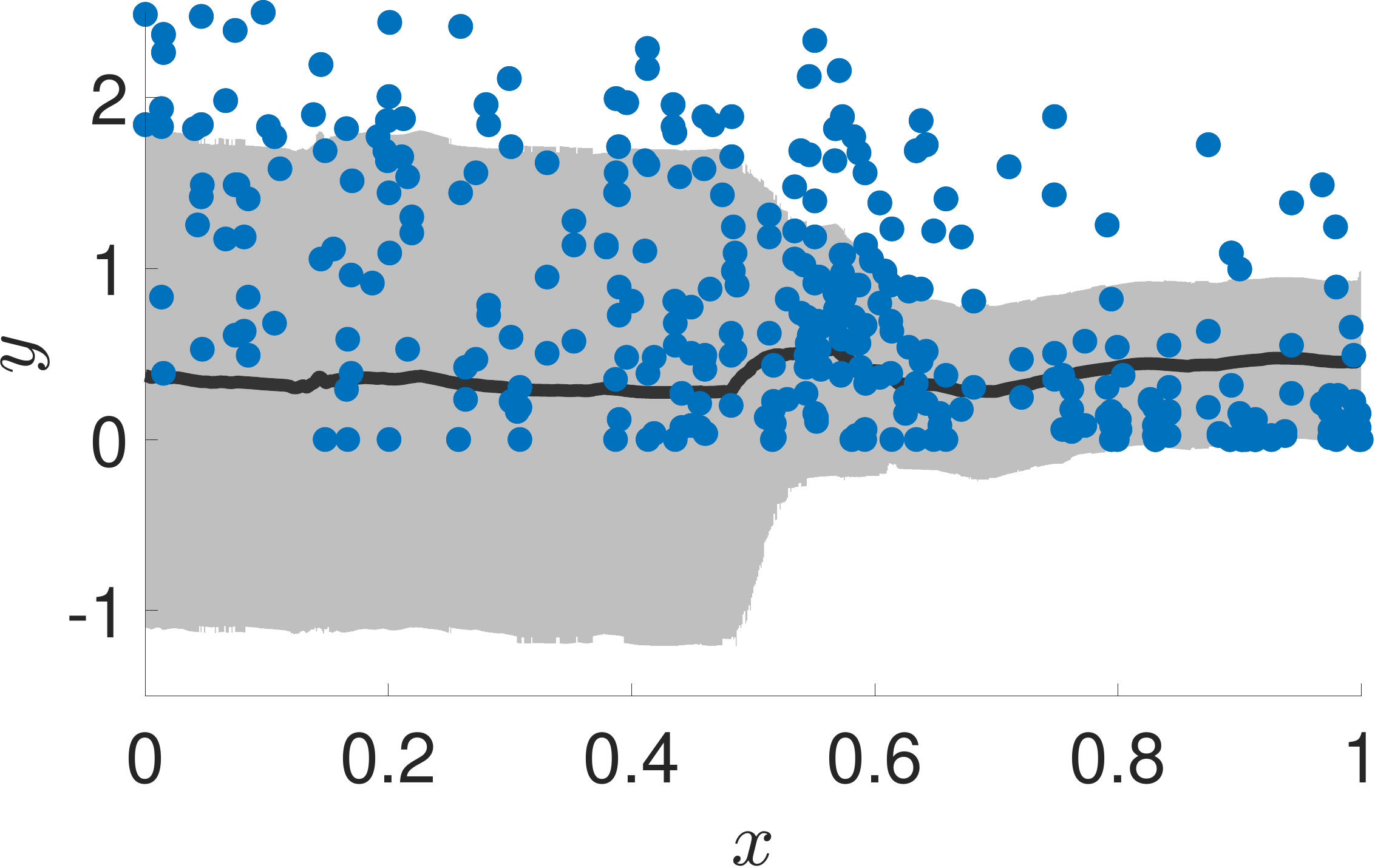}
    \end{minipage}
    \hspace{\fill} 
    \begin{minipage}{0.24\textwidth}
    \includegraphics[height=3.75cm]{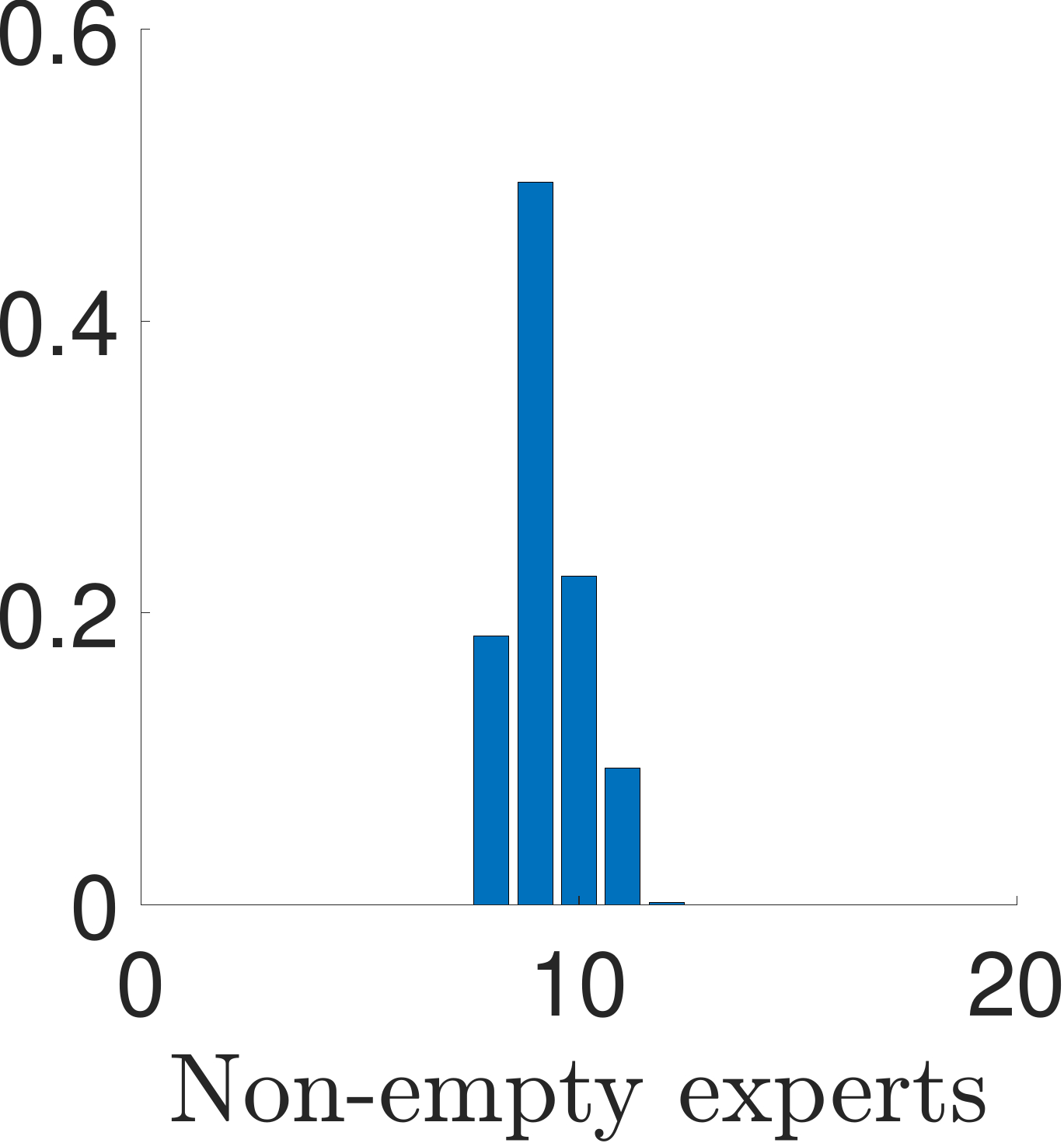}
    \end{minipage}
    \hspace{\fill} 
    \begin{minipage}{0.24\textwidth}
    \includegraphics[height=2.80cm]{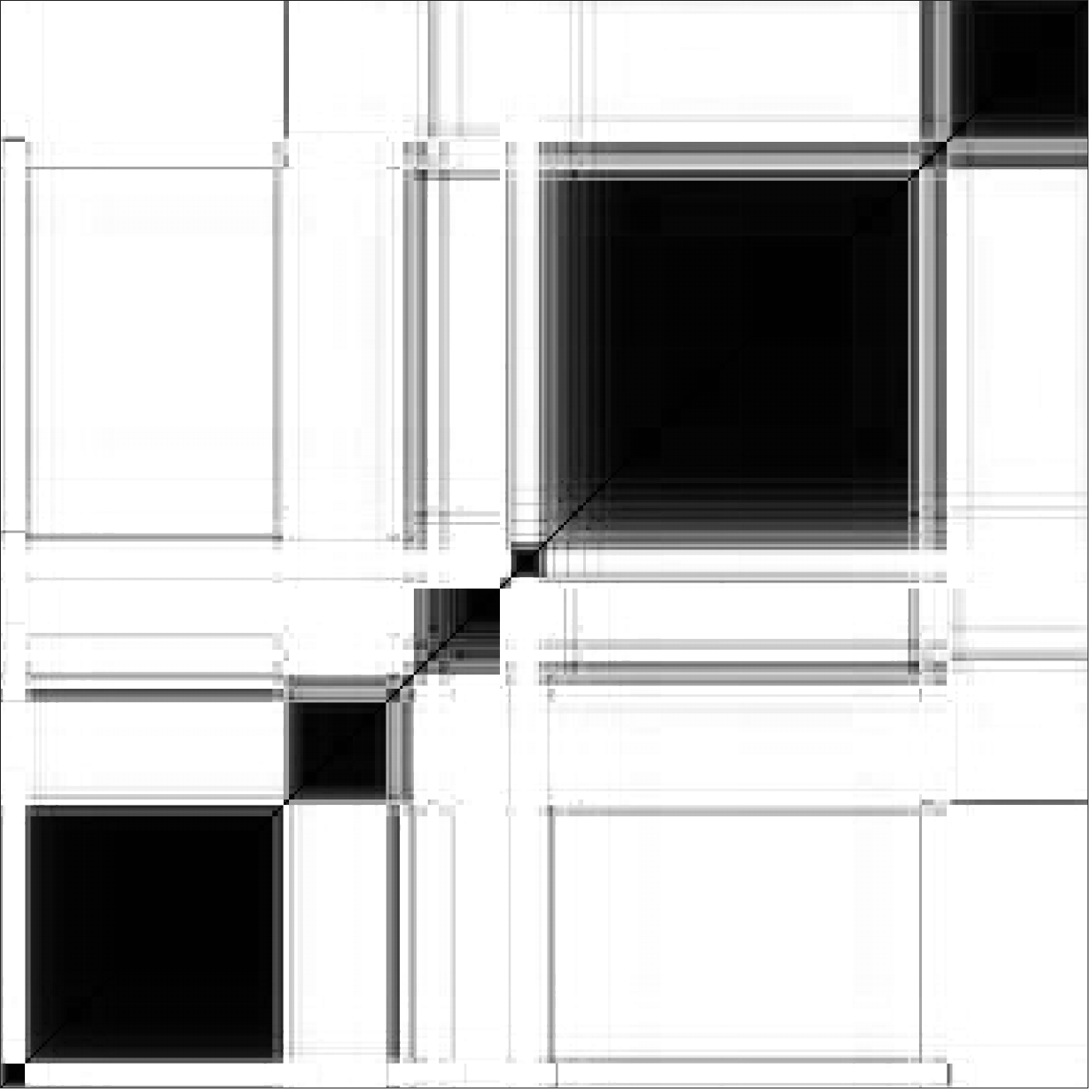}
    \vspace{0.575cm}
    \end{minipage}

    \vspace*{1cm} 
    
    \begin{minipage}{0.40\textwidth}
    \includegraphics[height=3.65cm]{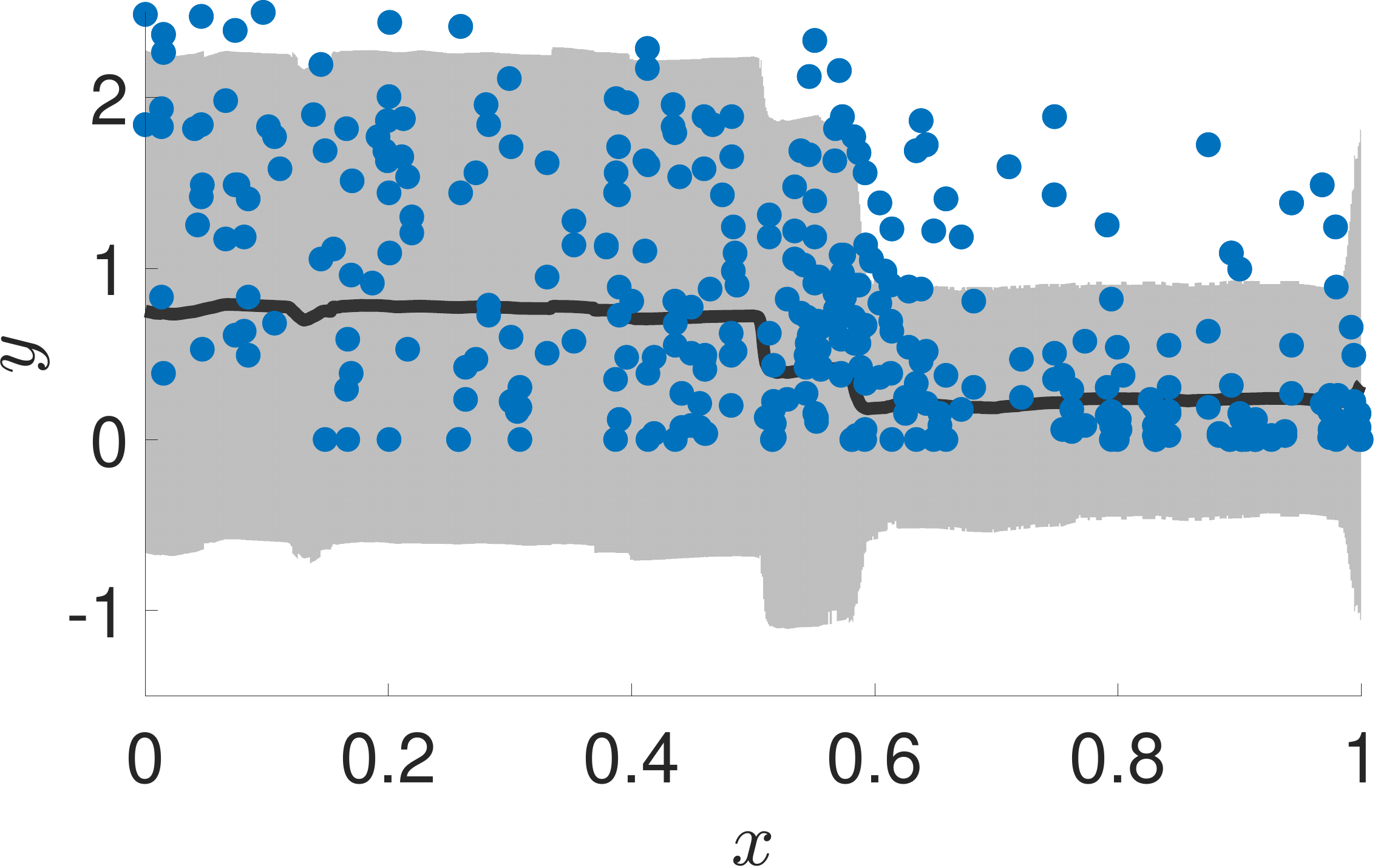}
    \end{minipage}
    \hspace{\fill} 
    \begin{minipage}{0.24\textwidth}
    \includegraphics[height=3.75cm]{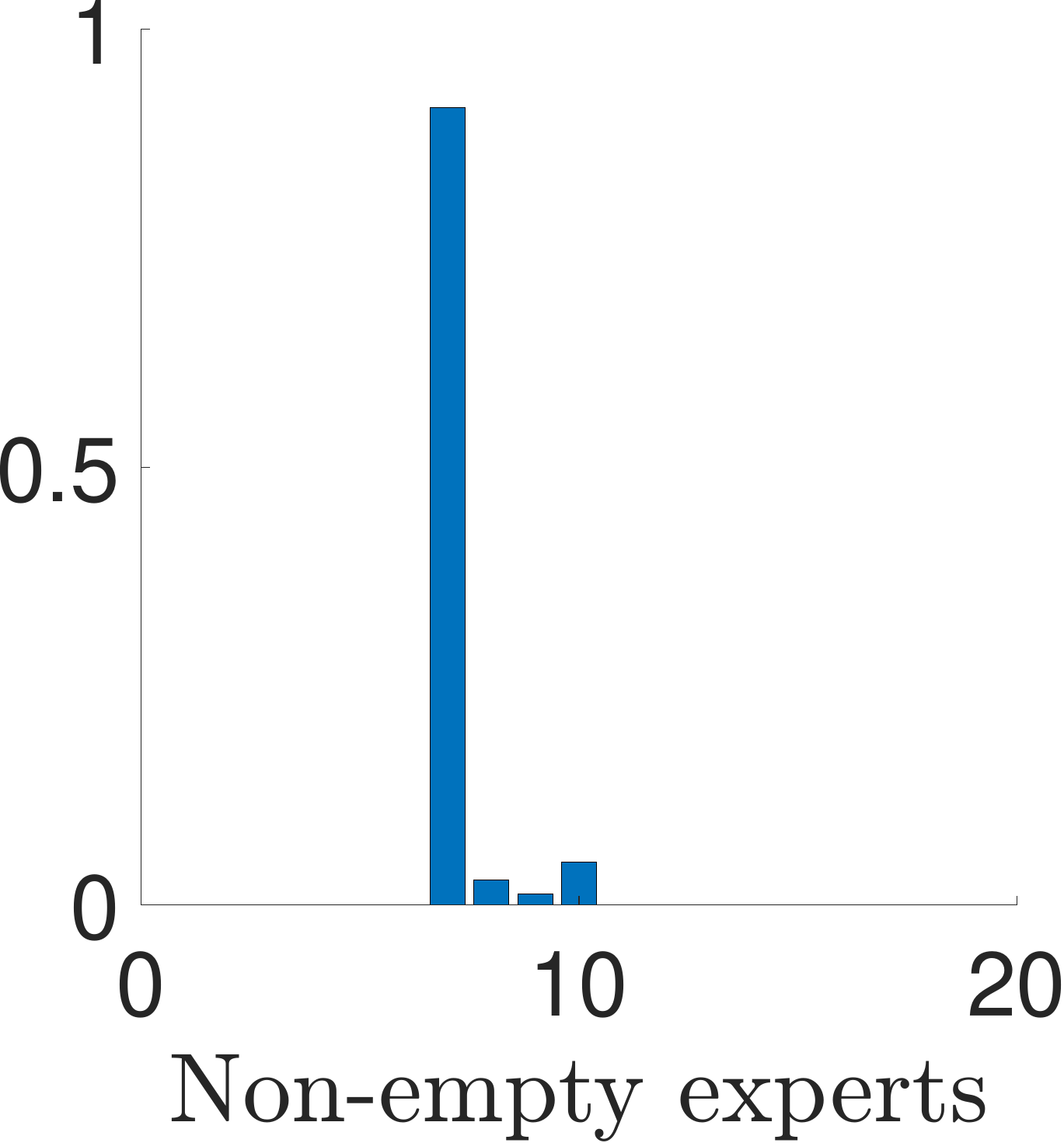}
    \end{minipage}
    \hspace{\fill} 
    \begin{minipage}{0.24\textwidth}
    \includegraphics[height=2.80cm]{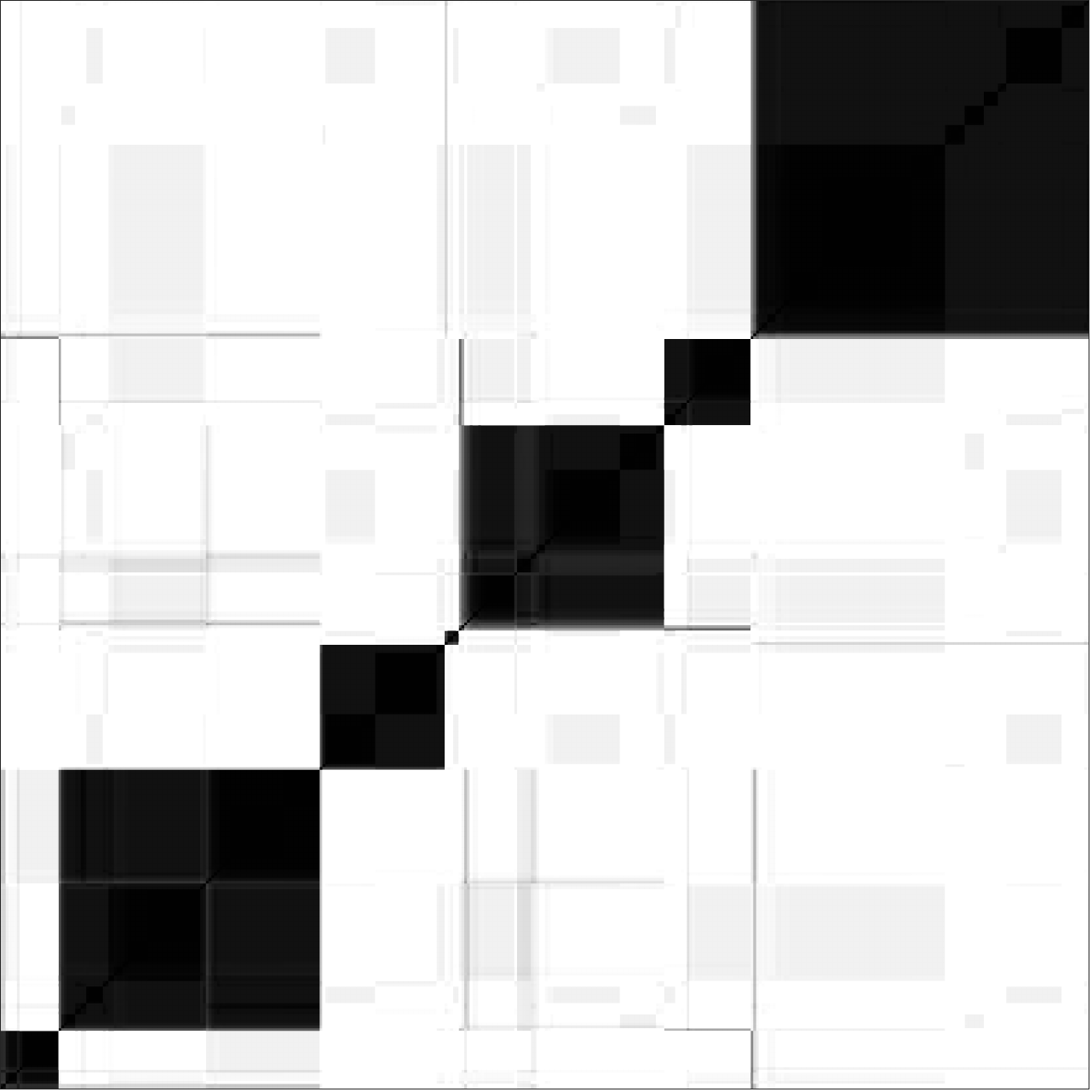}
    \vspace{0.575cm}
    \end{minipage}

    \vspace*{1cm} 
    
    \begin{minipage}{0.40\textwidth}
    \includegraphics[height=3.65cm]{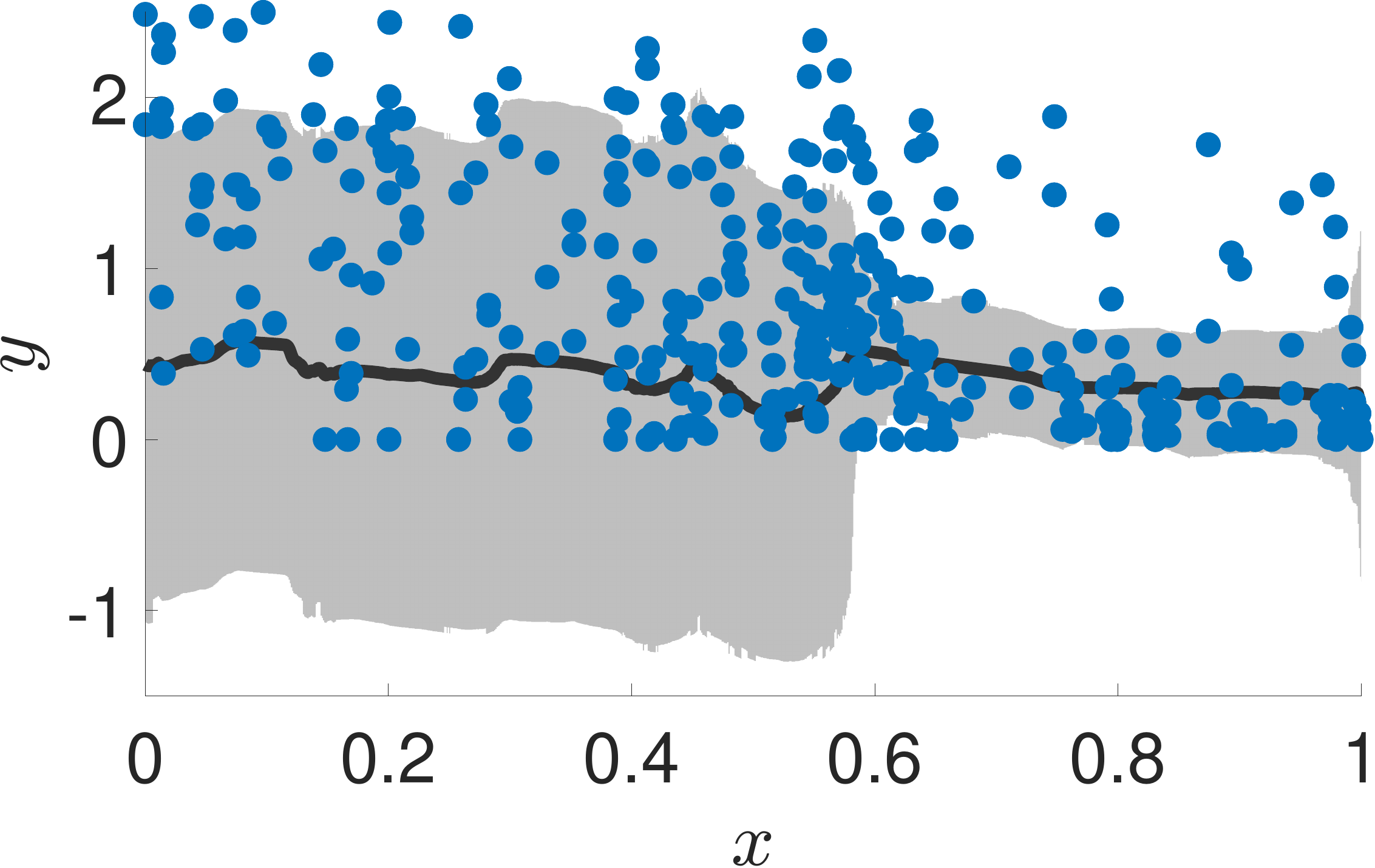}
    \end{minipage}
    \hspace{\fill} 
    \begin{minipage}{0.24\textwidth}
    \includegraphics[height=3.75cm]{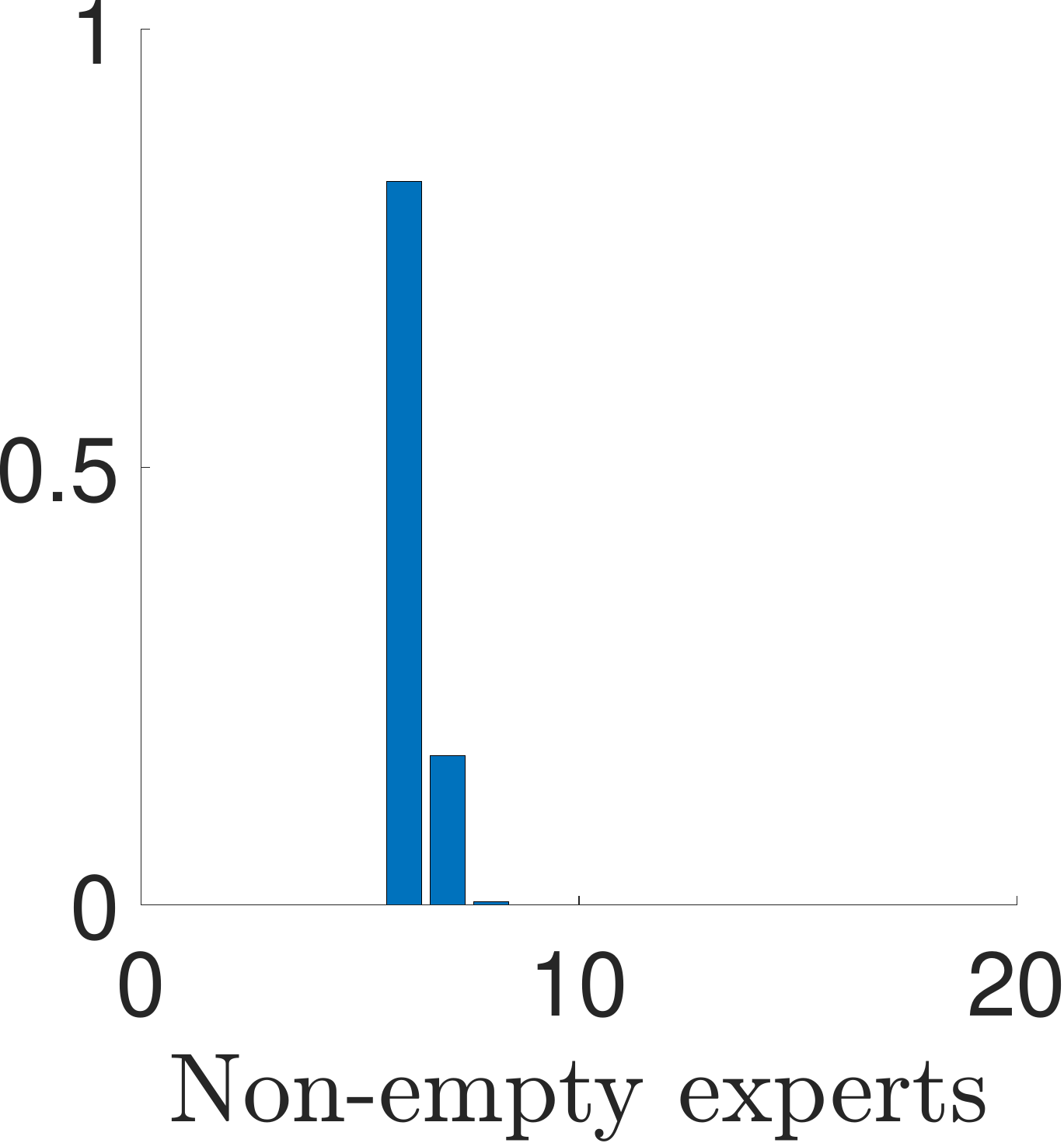}
    \end{minipage}
    \hspace{\fill} 
    \begin{minipage}{0.24\textwidth}
    \includegraphics[height=2.80cm]{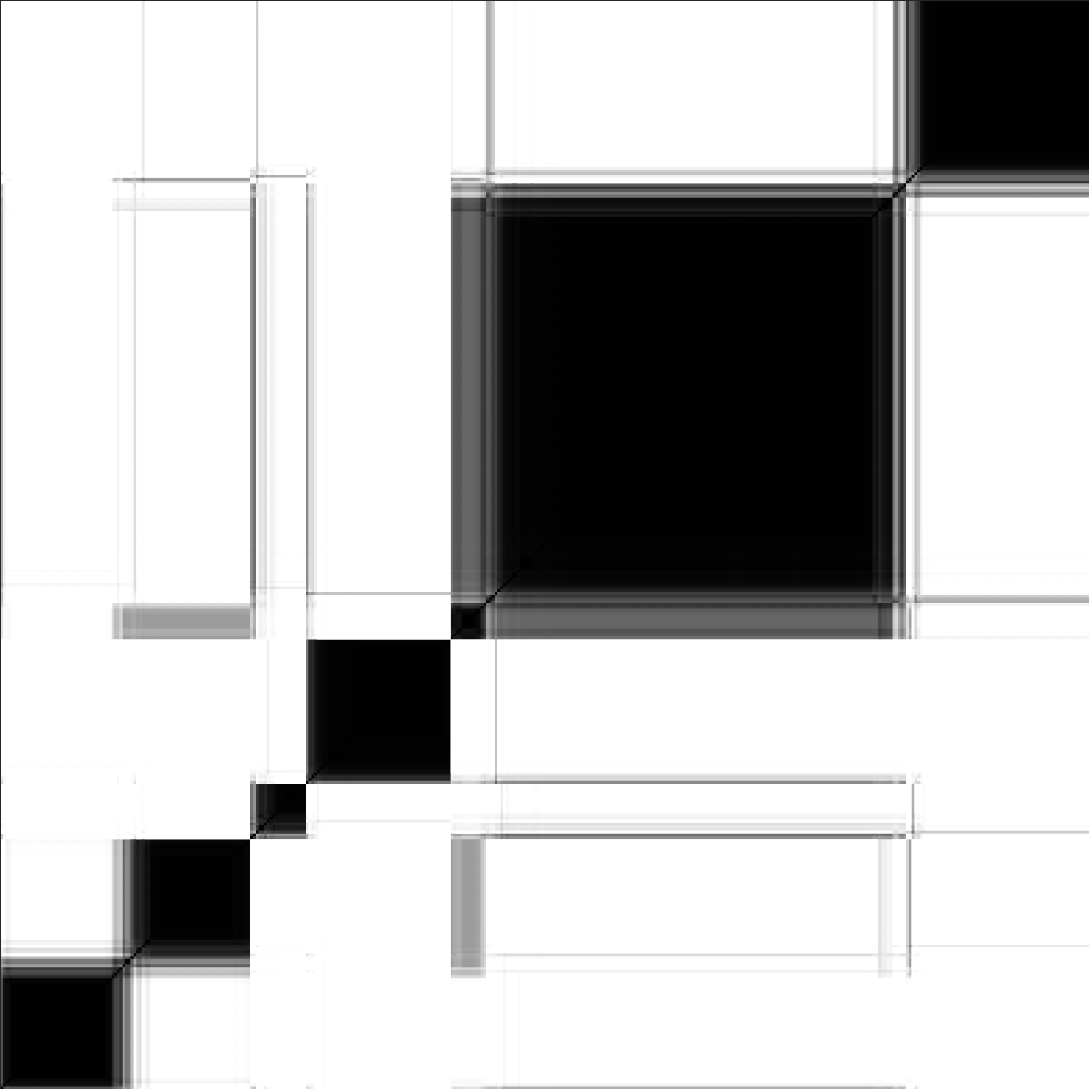}
    \vspace{0.575cm}
    \end{minipage}

    \vspace*{0.5cm} 
    
    \caption{\textit{4D Colorado precipitation data}. On the left, the highest density regions and median for a one-dimensional slice of the input space are shown in gray and black along with the data points in blue for different values of the Dirichlet concentration parameter $\alpha$. In the middle column, the posterior distribution for the number of non-empty experts is shown.  The last column shows the posterior similarity matrices summarizing the posterior over partitions. Values of $\alpha = 1.6$, $\alpha = 4$, and $\alpha = K/2 = 8$ are used and correspond to the top, middle, and bottom plots, respectively.} \label{im:coloradoDensityPSM}
\end{figure}

Lastly, the highest density regions of the predictive densities, the posterior over the number of clusters, and the posterior similarity matrix for SMC$^2$-MoE with $K = 16$ are presented in Figure \ref{im:coloradoDensityPSM} for the 4D Colorado precipitation data set with the 392 data points superimposed over the predictive densities.
The presented one-dimensional predictive distribution is computed for a one-dimensional slice of the 4D input space.
The slice corresponds to a latitude of 39.00$^{\circ}$, with a longitude range between the minimum and maximum longitude values present in the data set, at a time point corresponding to November, and with predictive elevation locations computed using linear interpolation based on the latitudes, longitudes, and elevations of the station locations in the data set.
The results show a clear transition around $x = 0.6$, which corresponds to a longitude of approximately -105.586$^{\circ}$.
The higher-variation area at $x \leq 0.6$ corresponds to areas of higher elevation with higher mean monthly precipitation \citep{Mahoney:2015}.

\section{Discussion}
\label{sec:discussion}
In this work, we introduced the use of SMC$^2$ in the context of mixtures of Gaussian process experts to conduct full Bayesian inference.
We have discussed and demonstrated how the previously proposed importance sampling based approach requires the number of importance samples to be exponential in the Kullback-Leibler divergence between the prior and posterior, which is computationally infeasible in many examples.
This is particularly true for complex targets that demonstrate departures from standard GP models, such as non-stationarity, heteroskedasticity, and discontinuity.
To overcome this, we extended the importance sampling based approach by using nested sequential Monte Carlo samplers to reduce the number of samples required while still being embarrassingly parallel.
These benefits offset the additional computational complexity introduced by the nested sequential Monte Carlo samplers, particularly for complex target distributions.
The inner SMC can also be replaced by optimization combined with a Laplace approximation to relieve some of the computational burden.
This similarly achieves the improvement from the exponential KL divergence of naive importance sampling to the polynomial KL divergence of an SMC sampler.

We have focused on gating networks defined through normalized kernels but highlight that the SMC$^2$ scheme can also be used for other choices of gating networks, such as the tree-based construction of treedGPs.
An advantage of the normalized kernel construction is that we can include priors on the kernel locations which encourage partitions associated to distinct regions.
This is in contrast to \cite{Zhang:2019}, who assume identically distributed kernel locations, resulting in overlapping partitions in the prior and IS proposal.
While our prior is constructed based on a grid of the input space (Table \ref{table:priorParametersND}), repulsive priors could be employed in higher dimensions \citep{petralia2012repulsive}. 
Another advantage is the data-driven choice of the number of clusters, through a sparsity-promoting prior on the weights.
Our experiments highlight that the concentration parameter $\alpha$ in the sparsity-promoting prior can be selected to balance sparsity, computational cost, and smoothness.
One might also consider an automatic approach for selecting or estimating $\alpha$.
In this case, the SMC framework should allow us to compute an unbiased estimator of the marginal likelihood on $\alpha$ followed by a particle marginal Metropolis-Hastings update or stochastic gradient descent for empirical Bayes optimization.

In future work, we aim to use mixtures of Gaussian process experts as priors for inverse problems such as in computed tomography or deconvolution.
In computed tomography, mixtures of Gaussian processes could model areas corresponding to differing tissues in the human body or materials in inanimate objects, that is, this would be a flexible alternative to level set methods or edge-preserving priors, such as total variation priors. 
For deconvolution, mixtures could be used to improve results by enabling modelling of discontinuous functions.
In these complex settings, efficient inference schemes for mixtures of GP experts are required.  
Additionally, it would be interesting to combine methods of online GP methods \citep{Leen:2000, Ranganathan:2011, Law:2022} with our methodology for computational speed-ups and extend the approach to alternative likelihoods over the partitions.

\acks{TH and LR were supported by Research Council of Finland (grant numbers 327734, 336787, 353094 and 359183). SW was supported by the Royal Society of Edinburgh (RSE) (grant number 69938).}


\bibliography{ref.bib}

\end{document}